\documentclass[runningheads]{llncs}

\usepackage[final,year=2026]{eccv}

\usepackage{microtype}
\usepackage{graphicx}
\usepackage{amsmath}
\usepackage{amssymb}
\usepackage{booktabs}
\usepackage{multirow}
\usepackage{tabularx}
\usepackage{array}
\usepackage{makecell}
\usepackage{stfloats}
\usepackage{adjustbox}
\usepackage{colortbl}
\usepackage{placeins}
\usepackage{pifont}
\usepackage[accsupp]{axessibility}

\usepackage[breaklinks,colorlinks,citecolor=eccvblue,allcolors=eccvblue]{hyperref}
\ifdefined\pdfminorversion
\fi
\definecolor{firstpurple}{RGB}{224, 211, 246}
\definecolor{secondpurple}{RGB}{239, 230, 250}
\definecolor{tableline}{RGB}{200, 200, 200}
\definecolor{tableheader}{RGB}{230, 230, 230}

\newcolumntype{C}{>{\centering\arraybackslash}X}
\newcommand{\cmark}{\ding{51}}
\newcommand{\xmark}{\ding{55}}

\title{\texorpdfstring{RoboStream: Weaving \underline{S}patio-\underline{T}emporal \underline{Rea}soning with \underline{M}emory in Vision-Language Models for \underline{Robo}tics}{RoboStream: Weaving Spatio-Temporal Reasoning with Memory in Vision-Language Models for Robotics}}

\titlerunning{RoboStream}
\authorrunning{Y.~Huang et al.}

\author{
    \makebox[\textwidth][c]{Yuzhi Huang$^{1,2*\ddagger\spadesuit}$ \enskip Jie Wu$^{1,2*\ddagger}$ \enskip Weijue Bu$^{3*}$ \enskip Ziyi Xiong$^{2,4\ddagger}$ \enskip Gaoyang Jiang$^{5}$ \enskip Ye Li$^{1}$}\\
    \vspace{1mm}
    \makebox[\textwidth][c]{Kangye Ji$^{1,2\ddagger}$ \enskip Shuzhao Xie$^{1}$ \enskip Yue Huang$^{6}$ \enskip Chenglei Wu$^{2}$ \enskip Jingyan Jiang$^{4\dagger}$ \enskip Zhi Wang$^{1\dagger}$}
}
\institute{
\small
$^1$ Shenzhen International Graduate School, Tsinghua University\\
$^2$ YuanxingGuangnian Robotics\\
$^3$ China University of Mining and Technology\\
$^4$ Shenzhen Technology University\\
$^5$ Huazhong University of Science and Technology\\
$^6$ Xiamen University\\
\vspace{1mm}
Website: \texttt{https://robostream123.github.io/}
\vspace{-8mm}
}

\begin{document}
\maketitle

\enlargethispage{1.5cm}

\begin{abstract}
Enabling reliable long-horizon robotic manipulation is a crucial step toward open-world embodied intelligence. However, VLM-based planners treat each step as an isolated observation-to-action mapping, forcing them to reinfer scene geometry from raw pixels at every decision step while remaining unaware of how prior actions have reshaped the environment. Despite strong short-horizon performance, these systems lack the spatio-temporal reasoning required for persistent geometric anchoring and memory of action-triggered state transitions. Without persistent state tracking, perceptual errors accumulate across the execution horizon, temporarily occluded objects are catastrophically forgotten, and compounding failures lead to precondition violations that cascade through subsequent steps. In contrast, humans maintain a persistent mental model that continuously tracks spatial relations and action consequences across interactions rather than reconstructing them at each instant. Inspired by this human capacity for causal spatio-temporal reasoning with persistent memory, we propose \textbf{RoboStream}, a training-free framework that achieves geometric anchoring through Spatio-Temporal Fusion Tokens (STF-Tokens), which bind visual evidence to 3D geometric attributes for persistent object grounding, and maintains causal continuity via a Causal Spatio-Temporal Graph (CSTG) that records action-triggered state transitions across steps. This design enables the planner to trace causal chains and preserve object permanence under occlusion without additional training or fine-tuning. RoboStream achieves a 90.5\% success rate on long-horizon RLBench tasks and a 44.4\% success rate on challenging real-world block-building tasks, where both SoFar and VoxPoser score 11.1\%, demonstrating that spatio-temporal reasoning and causal memory are critical missing components for reliable long-horizon manipulation.
\end{abstract}

\vspace{-0.2em}
{\footnotesize
\noindent\rule{0.18\textwidth}{0.4pt}\\[-0.2em]
$^*$ Equal contribution. $^\dagger$ Corresponding author. $\spadesuit$ Project lead.\\
$^\ddagger$ Work done during an internship at Yuanxing Robotics.}
\vspace{0.3em}

\section{Introduction}
\label{sec:intro}

Vision-language models (VLMs) have emerged as powerful planners for robotic manipulation, leveraging internet-scale semantic knowledge and visual perception to translate high-level instructions into executable action sequences~\cite{zitkovich2023rt, kim2025openvla, driess2023palm, jiang2023vima}. VLM-based systems have demonstrated strong performance on short-horizon manipulation tasks, including precise grasping, pose-aware placement, and decomposition of language instructions into subgoal sequences~\cite{brohan2023saycan, qisofar, huang2023voxposer, shridhar2023perceiver}.

\begin{figure}[t]
        \centering  
        \includegraphics[width=1\textwidth]{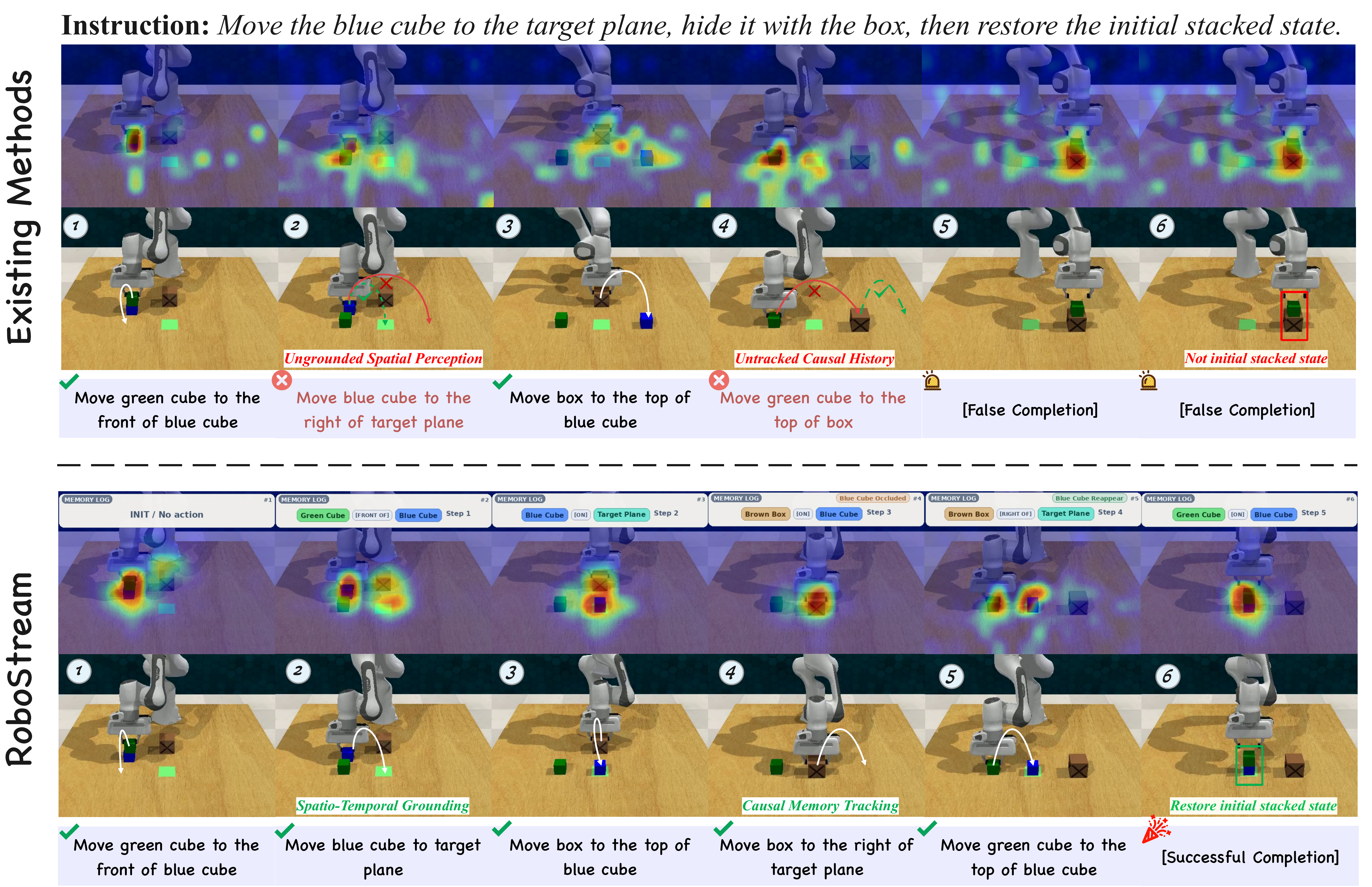}
        \caption{\textbf{Qualitative comparison between existing methods~\cite{qisofar} and RoboStream.} This long-horizon task involves object occlusion and state restoration. \textbf{Top (Existing Methods):} Diffuse attention heatmaps and absent memory logs reveal two cascading failures: \textit{Ungrounded Spatial Perception} misplaces the blue cube at Step~2, while \textit{Untracked Causal History} leaves the planner unaware of the occluded cube at Step~4, resulting in an incorrect final state. \textbf{Bottom (RoboStream):} Concentrated attention heatmaps and persistent memory logs show that \textit{Spatio-Temporal Grounding} enables accurate object placement at Step~2, while \textit{Causal Memory Tracking} maintains awareness of the occluded cube at Step~4, enabling successful task completion.}
\label{fig:teaser}
\end{figure}

However, this success does not readily extend to long-horizon tasks, which demand sustained spatial and causal awareness across multi-step action sequences that irreversibly alter the environment~\cite{shi2025hi, zhao2025cot, wu2023tidybot}. Current VLM-based planners lack the persistent state tracking required to bridge observations across steps, forcing them to reinfer world state from partial observations at each decision step~\cite{hanrobocerebra, huang2023inner, zhou2025code}. Without accumulated context, perceptual errors and action effects compound across steps, ultimately resulting in cascading failures, as illustrated in Fig.~\ref{fig:teaser}. In contrast, humans maintain a coherent mental model~\cite{lake2017building} that continuously integrates spatial relations and causal consequences as actions unfold, tracking not only where objects are but also how each action has cumulatively reshaped the world.

Replicating this human capacity for spatio-temporal reasoning requires persistent tracking of two interdependent quantities, namely the evolving geometric state of each object across steps and the causal record of how actions modify those states. Tracking alone is insufficient, as geometric states must be anchored in 3D space for reliable spatial reasoning while causal records must be structured to support inference over displaced or occluded objects. 
Current VLM-based planners fail at both, as illustrated in~Fig.~\ref{fig:teaser}, exposing two tightly coupled representation deficits.
\ding{172}~\textit{Ungrounded Spatial Perception}: spatial relations in manipulation are latent state components continually rewritten by contact and motion, yet current systems infer them implicitly via image-centric perception without geometric anchoring or cross-step identity binding~\cite{chen2024spatialvlm, huang2024copa, yuan2025robopoint}. This forces geometry reconstruction from raw pixels at every step, where small inaccuracies accumulate into spatial hallucinations and precondition violations.
\ding{173}~\textit{Untracked Causal History}: although actions irreversibly alter the environment, existing methods retain no record linking actions to induced state transitions. Without causal tracking, planners remain unaware of displaced objects and lose track of occluded ones entirely, leading to precondition violations when subsequent actions depend on altered states~\cite{wen2025dexvla, zhen20243d}.

To address these two deficits, we propose \textbf{RoboStream}, a training-free framework that weaves spatio-temporal reasoning and persistent memory directly into the VLM planning loop through \textit{Spatio-Temporal Grounding} and \textit{Causal Memory Tracking}. For grounding, Spatio-Temporal Fusion Tokens (STF-Tokens) distill visual appearance and 3D geometry (centroid and Gaussian shape) of each object into identity-persistent primitives, shifting VLM reasoning from pixel-level reconstruction to structured object-instance references that suppress cascading spatial errors. For tracking, a Causal Spatio-Temporal Graph (CSTG) records action-triggered state transitions alongside persistent object identities, preserving object permanence under occlusion without direct visual evidence.
We evaluate RoboStream on five benchmarks spanning simulated and real-world settings: SIMPLER~\cite{li2025evaluating}, long-horizon RLBench tasks~\cite{james2020rlbench}, 6-DoF SpatialBench~\cite{qisofar}, Open6DOR V2~\cite{ding2024open6dor}, and long-horizon real-world Franka experiments including block building, disassembly, and hide-and-restore under occlusion.
Across all settings, RoboStream outperforms existing VLM-based planners, demonstrating that spatio-temporal reasoning and persistent memory are essential for robust long-horizon manipulation.
Our contributions are as follows:

1) We introduce \textbf{RoboStream}, a training-free framework weaving spatio-te\-m\-p\-o\-r\-a\-l reasoning with persistent memory to overcome spatial hallucination and causal blindness in VLM-based planners. By anchoring geometric evidence across steps and tracking causal state transitions, RoboStream builds persistent world representations enabling robust long-horizon reasoning without retraining.

2) We design two synergistic mechanisms for Spatio-Temporal Grounding and Causal Memory Tracking. \textit{Spatio-Temporal Fusion Tokens (STF-Tokens)} bind visual evidence to 3D geometry, converting pixel-level inference into deterministic spatial references. A \textit{Causal Spatio-Temporal Graph (CSTG)} maintains object identities and records action-triggered state transitions, enabling causal inference of occluded states and preventing error accumulation over long horizons.

3) RoboStream achieves state-of-the-art results across five benchmarks spanning long-horizon manipulation (RLBench, real-world Franka), zero-shot generalization (SIMPLER), and spatial reasoning (6-DoF SpatialBench, Open6DOR V2), validating that weaving spatio-temporal reasoning with persistent memory is essential for robust manipulation across diverse settings.

\section{Related Work}
\label{sec:related_work}

\subsection{Spatial Understanding with VLMs}
Spatial understanding underpins robotic manipulation by recovering geometric relationships~\cite{huang2026thinking,wen2025dynamicverse}. Existing approaches address VLM limitations through two main strategies.
Tool-augmented methods~\cite{ding2024open6dor, qin2025robofactory, zhudexflywheel, tan2025roboos} delegate VLMs to orchestrate external visual foundation models or geometric engines. SpaceTools~\cite{chen2025spacetools} combines dual-interaction reinforcement learning with geometric reasoning, while TIGER~\cite{han2025tiger} synthesizes code-based geometric routines. These externalize computation but add overhead and require module coupling.
3D-data-driven fine-tuning~\cite{cheng2024spatialrgpt, song2025robospatial, daxberger2025mm, raysat, wang2025spatial457} strengthens spatial representations through pseudo-3D or real-world 3D data. RoboRefer~\cite{zhouroborefer} integrates depth encoders, SpatialBot~\cite{cai2025spatialbot} applies progressive spatial curricula, and SpatialVLM~\cite{chen2024spatialvlm} leverages large-scale 3D spatial QA datasets. Fine-tuning incurs substantial data and training costs.
RoboStream contrasts by anchoring visual evidence directly to explicit 3D geometry via STF-Tokens, requiring no fine-tuning and maintaining persistent spatial representations across sequential steps without external modules or retraining.

\subsection{VLMs for Robotic Manipulation}
Robotic manipulation bridges semantic reasoning with execution along three complementary axes.
Logic-grounded task planning~\cite{huang2022language, singh2023progprompt, zengsocratic, huang2023grounded} decomposes instructions into structured sequences. SayCan~\cite{brohan2023saycan} scores action feasibility via affordance models, while Code as Policies~\cite{liang2023code} synthesizes executable programs for long-horizon scheduling. These methods prioritize logical structure but treat execution steps independently.
Geometry-aware constraint generation~\cite{yu2023language, sharma2022correcting, yuan2025robopoint, huang2025rekep, huang2025a3vlm, fang2024moka} grounds planning in spatial constraints. VoxPoser~\cite{huang2023voxposer} synthesizes 3D value maps for zero-shot planning, CoPa~\cite{huang2024copa} defines constraints over geometric keypoints, and SoFar~\cite{qisofar} maps language to 6-DoF poses. However, these approaches reconstruct geometric constraints at each step without persistent tracking.
Closed-loop reasoning~\cite{driess2023palm, du2023vision, zhou2025code, liuinteractive} enhances adaptability through perceptual feedback. Inner Monologue~\cite{huang2023inner} implements self-correcting loops, and VILA~\cite{hu2024look} reasons over image sequences for physical intuition. Yet these methods remain reactive, recovering from failures without reasoning about action-induced state changes or preconditions.
RoboStream unifies geometry and causality through a CSTG that records action-state transitions, enabling VLMs to maintain causal consistency while reasoning about occluded states and external disturbances.

\subsection{Long-Horizon Task Planning}
Long-horizon manipulation poses a distinct challenge, requiring systems to translate high-level instructions into coherent action sequences while maintaining consistency across extended temporal spans. Dynamic visual conditions and partial observability amplify the challenge of maintaining causal consistency across sequential decisions~\cite{hanrobocerebra}.
Hierarchical decomposition~\cite{shi2024yell, belkhale2024rt, lihamster} organizes multi-step execution by encoding intermediate representations. HiRobot~\cite{shi2025hi} segments instructions into atomic commands, DexVLA~\cite{wen2025dexvla} couples VLM planning with diffusion-based policies, and RoboMatrix~\cite{mao2024robomatrix} implements three-tier hierarchical control. LoHoVLA~\cite{yang2025lohovla} aligns semantic intent via natural-language sub-goals. These strategies organize execution without tracking how past actions reshape the environment.
Chain-of-thought methods~\cite{wei2022chain, mu2023embodiedgpt, zawalski2025robotic, zhang2025learning} strengthen planning via explicit future-state reasoning. CoT-VLA~\cite{zhao2025cot} reasons over latent visual subgoals, ManualVLA~\cite{gu2025manualvla} generates executable visual guidance, and GR-3~\cite{cheang2025gr} learns flow-matching trajectories. While improving short-horizon decisions, these methods lack persistent memory to track action consequences.
RoboStream addresses this gap through a CSTG recording actions, environmental consequences, and external disturbances, enabling VLMs to trace state evolution and reason causally about how past actions constrain future possibilities under partial observability.

\section{Method}
\label{sec:method}
\subsection{Overview}
\label{sec:overview}

\begin{figure}[t]
    \centering
    \includegraphics[width=1\textwidth]{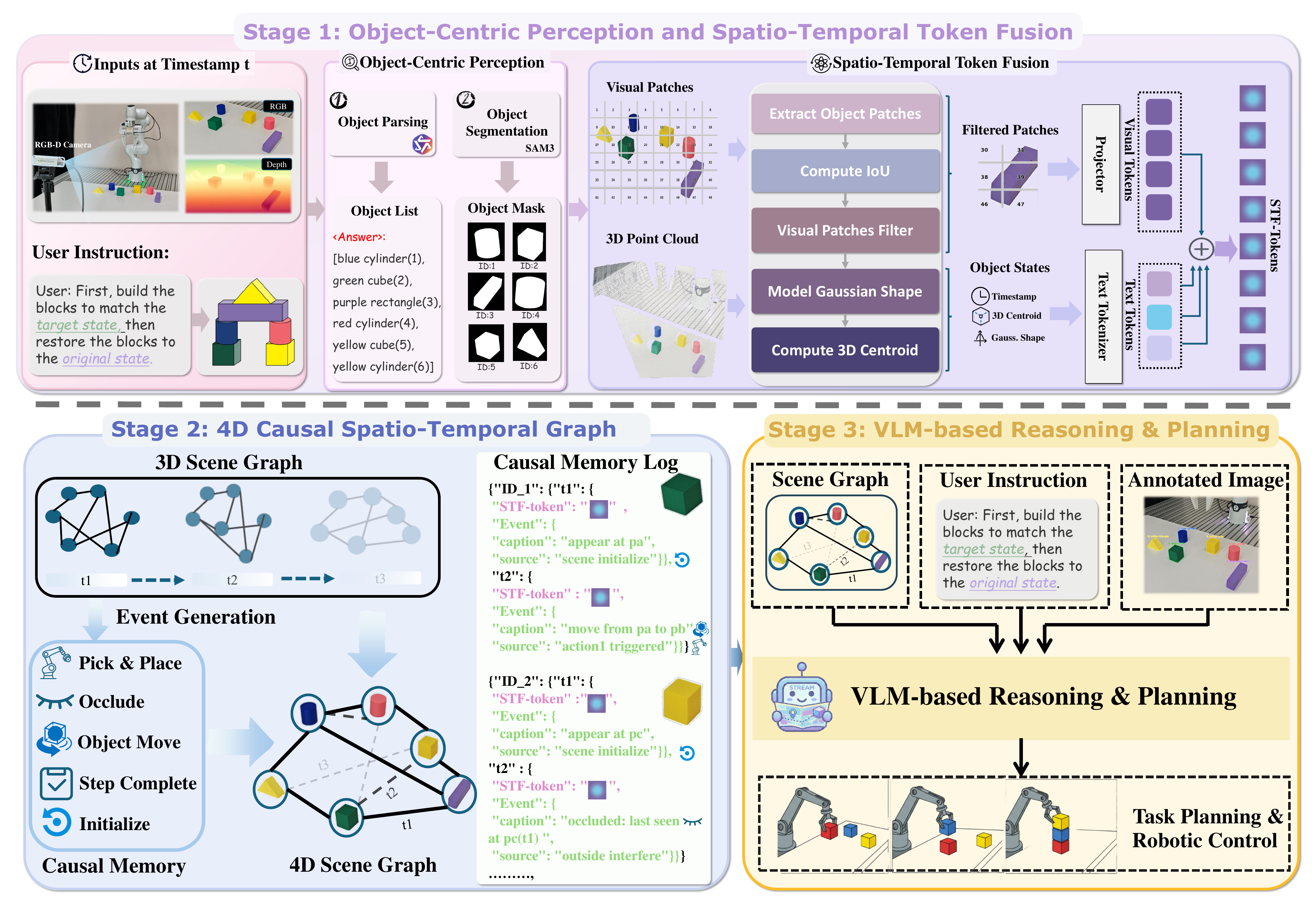}
    \caption{\textbf{Overview of the RoboStream Framework.} The system processes multi-modal RGB-D inputs through three hierarchical stages: (1) Object-Centric Perception and Spatio-Temporal Token Fusion, where raw visual data is distilled into STF-Tokens by grounding visual evidence within 3D geometric primitives (centroids and Gaussian shapes); (2) 4D Causal Spatio-Temporal Graph Construction, which maintains persistent memory encoding object identities and state transitions across time steps; and (3) VLM-based Reasoning and Planning, where the VLM leverages both the CSTG and visual inputs to perform high-level task planning and precise robotic manipulation.}
    \label{fig:overview}
\end{figure}

VLMs exhibit strong generalization in high-level manipulation planning~\cite{brohan2023saycan, zitkovich2023rt, kim2025openvla}, yet existing planners treat each step independently, lacking the persistent spatio-temporal memory required for long-horizon tasks. Without structured world-state tracking, perception errors compound over time and recovery from unexpected disturbances becomes substantially more difficult.
We propose RoboStream, which constructs geometry-grounded STF-Tokens by binding visual evidence to 3D geometric attributes from RGB-D observations (Sec.~\ref{sec:stf}), organizes them into a Causal Spatio-Temporal Graph (CSTG) that maintains persistent object identities and logs action-triggered state transitions in a sliding-window memory (Sec.~\ref{sec:cstg}), and feeds this structured memory into the VLM for chain-of-thought verification to generate 6-DoF action directives (Sec.~\ref{sec:vlm_planning}), as illustrated in Fig.~\ref{fig:overview}.

Formally, given RGB-D observation $(I_t, D_t)$, goal specification $I_{\text{goal}}$, and memory $\mathcal{M}_{<t}$, RoboStream constructs STF-Tokens encoding per-object 3D geometric properties and visual appearance:
\begin{equation}
    \mathcal{T}_t^{\text{STF}} = \Phi_{\text{enc}}(I_t, D_t \mid \mathcal{M}_{<t}),
\end{equation}
which are incorporated as nodes into the CSTG $\mathcal{G}_t^{\text{CST}}$. By recursively merging incoming STF-Tokens with historical graph state, the CSTG maintains persistent object identities and inter-object spatial relations across time without requiring model retraining:
\begin{equation}
    \mathcal{G}_t^{\text{CST}} = \Psi_{\text{graph}}(\mathcal{T}_t^{\text{STF}},\; \mathcal{G}_{t-1}^{\text{CST}}).
\end{equation}
Action generation is then decoupled to balance VLM semantic generalization with robot execution precision. Given a context-augmented prompt $\mathcal{P}_t$ assembled from $\mathcal{G}_t^{\text{CST}}$ and the current observation, the VLM produces a semantic action directive $\hat{a}_t$ conditioned on the goal $I_{\text{goal}}$ and the accumulated causal history:
\begin{equation}
    \hat{a}_t = \Pi_{\theta}(I_{\text{goal}},\; \mathcal{P}_t \mid \mathcal{G}_t^{\text{CST}}),
    \label{eq:action_directive}
\end{equation}
which is deterministically instantiated into a 6-DoF pose by resolving object identities against the geometry of STF-Tokens stored in $\mathcal{G}_t^{\text{CST}}$:
\begin{equation}
    a_t = \Phi_{\text{inst}}(\hat{a}_t,\; \mathcal{G}_t^{\text{CST}}).
\end{equation}
This decoupled design preserves VLM semantic reasoning while delegating precise coordinate extraction to deterministic geometry resolution, enabling training-free deployment across various scenarios. We elaborate on each stage in Sec.~\ref{sec:stf}--\ref{sec:vlm_planning}.

\subsection{Object-Centric Perception}
\label{sec:perception}
To construct scene representations prioritizing task-relevant entities, we employ a two-stage perception pipeline. Given the RGB-D observation $(I_t, D_t)$, goal specification $I_{\text{goal}}$, and accumulated memory $\mathcal{M}_{<t}$, we prompt a VLM to identify task-relevant objects and output open-vocabulary descriptors $\mathcal{D}=\{d_1,\ldots,d_N\}$~\cite{qisofar, zhouroborefer}. The goal specification $I_{\text{goal}}$ accepts either a reference goal image or a natural language instruction, both applicable in real-world and simulation settings. Conditioning on $\mathcal{M}_{<t}$ allows the descriptor set to maintain consistent object identities across steps, implicitly filtering task-irrelevant entities and reducing computational overhead in downstream modules. The resulting descriptors condition an open-vocabulary segmentation model to produce pixel-accurate masks $\{M_i\}_{i=1}^{N}=\mathrm{SAM3}(I_t,\mathcal{D})$~\cite{carion2025sam}, where $M_i\in\{0,1\}^{H\times W}$. Combined with the depth map $D_t$~\cite{depthanything}, each mask defines a spatially coherent region for 3D point cloud extraction, providing the inputs to $\Phi_{\text{enc}}$ for constructing $\mathcal{T}_t^{\text{STF}}$ as in Sec.~\ref{sec:stf}.

\subsection{Spatio-Temporal Token Fusion}
\label{sec:stf}

Given instance masks $\{M_i\}_{i=1}^{N}$ and the point cloud from the perception stage, we construct a per-object Spatio-Temporal Fusion Token $\tau_i^t$ by integrating visual evidence with metric geometry into a unified, object-centric representation; the full set $\mathcal{T}_t^{\text{STF}} = \{\tau_i^t\}_{i=1}^{N}$ constitutes the encoder output $\Phi_{\text{enc}}$ defined in Sec.~\ref{sec:overview}. For each object $o_i$, the mask $M_i$ is applied to the RGB observation $I_t$ to obtain a masked image $\hat{I}_i^t$, which is passed through the VLM visual encoder~\cite{radford2021learning, dosovitskiy2020image} to produce a $16 \times 16$ grid of patch tokens. Patches with an IoU exceeding $I_{th}$ relative to $M_i$ are selected, projected into the language space, and aggregated to form the visual evidence $\mathbf{v}_i^t$. Unlike methods that pass multiple masked images directly to the VLM, this design avoids redundant background processing by compressing each object's filtered visual tokens, spatial position, geometry, and temporal trajectory into a compact token sequence, shifting VLM reasoning from raw pixels to structured object instances.This compact-token design is orthogonal to recent efficient-inference methods that reduce VLA or vision-transformer computation through adaptive reasoning, model scheduling, token pruning, speculative decoding, or structural pruning\cite{li2026elegantvla,li2025sp,li2025prance,li2025ts}.

In parallel, we define a shape representation vector $\mathbf{s}_i^t$ constructed from Gaussian distribution parameters and geometric extrema along each spatial axis: $\mathbf{s}_i^t = \{(\mu_a, \sigma_a, a_{\min}, a_{\max}) \mid a \in \{x,y,z\}\}$. Unlike traditional bounding boxes, this captures geometric extent more completely, with the joint mean-standard-deviation encoding suppressing outlier noise while correcting Gaussian approximation bias. The centroid $\mathbf{c}_i^t = \mathrm{median}(\mathbf{P}_i) \in \mathbb{R}^3$ is extracted from the object-specific point cloud $\mathbf{P}_i$. Both $\mathbf{c}_i^t$ and $\mathbf{s}_i^t$ are serialized via the text tokenizer into the same language embedding space as $\mathbf{v}_i^t$, placing visual appearance and 3D geometry within a single context window. This shared space allows the VLM to jointly attend over heterogeneous visual and spatial signals without architectural modification, encoding the complete per-object state as:
\begin{equation}
\tau_i^t = \Big\langle \, \mathbf{v}_i^t, \; \mathbf{c}_i^t, \; \mathbf{s}_i^t, \; t \, \Big\rangle,
\label{eq:stf_token}
\end{equation}
where $t$ denotes the current timestamp. By grounding visual tokens in explicit 3D geometry and temporal context, STF-Tokens enable deterministic spatial queries and suppress cascading errors in long-horizon manipulation tasks.

\subsection{4D Causal Spatio-Temporal Graph}
\label{sec:cstg}

While STF-Tokens stabilize per-object spatial perception, long-horizon manipulation additionally requires reasoning about inter-object spatial relations and action causality. We address this by constructing the Causal Spatio-Temporal Graph (CSTG) $\mathcal{G}_t^{\text{CST}}$~\cite{armeni20193d, wald2020learning}, which realizes the graph update operator $\Psi_{\text{graph}}$ in Sec.~\ref{sec:overview} by integrating incoming STF-Tokens $\mathcal{T}_t^{\text{STF}}$ with the historical state $\mathcal{G}_{t-1}^{\text{CST}}$ through a spatial scene graph and a causal memory log. At each timestamp $t$, we construct a spatial scene graph $\mathcal{G}_t = (\mathcal{V}_t, \mathcal{E}_t)$ where each node $v_i \in \mathcal{V}_t$ corresponds to an object in $\mathcal{T}_t^{\text{STF}}$ and maintains a sliding-window buffer $\{\tau_i^{t-K+1}, \ldots, \tau_i^{t}\}$ of length $K$. Edges $e_{ij} \in \mathcal{E}_t$ encode pairwise spatial relations via Euclidean distance $d_{ij} = \|\mathbf{c}_i - \mathbf{c}_j\|_2$ and directional offset $\Delta\mathbf{c}_{ij} = \mathbf{c}_j - \mathbf{c}_i$.

Building upon this structure, semantic events are detected by comparing STF-Tokens across the sliding window, covering both object-level events (e.g., planned displacement, unintended collision, occlusion) and task-level events (e.g., action execution, subtask completion). Each event is recorded with its timestamp, location, and causal source in the memory log $\mathcal{H}_t^{K}$, which is integrated with the spatial graph to form the complete CSTG:
\begin{equation}
\mathcal{G}_t^{\text{CST}} = \big( \mathcal{G}_t,\; \mathcal{H}_t^{K} \big).
\label{eq:cstg}
\end{equation}
This structured representation enables deterministic verification of action preconditions at each planning step, triggering re-planning upon detected failure to prevent error accumulation over long horizons.

\subsection{VLM-based Causal Reasoning and Planning}
\label{sec:vlm_planning}

At each planning step, RoboStream assembles the context-augmented prompt $\mathcal{P}_t$ from the CSTG $\mathcal{G}_t^{\text{CST}}$ encoding current object states and spatial relations, the annotated observation with object identity labels overlaid~\cite{yang2023set}, and the goal specification $I_{\text{goal}}$. This fusion of structured causal memory with grounded visual evidence enables $\Pi_{\theta}$ to reason jointly over historical context and current scene geometry. The VLM then performs chain-of-thought spatial verification~\cite{wei2022chain}, reviewing historical states, checking action preconditions against object geometry, and resolving causal conflicts from occlusion or dependencies, before generating the semantic action directive $\hat{a}_t$ as defined in Eq.~\ref{eq:action_directive}. Finally, $\hat{a}_t$ is instantiated into the 6-DoF action $a_t$ via $\Phi_{\text{inst}}$ by resolving object identities against STF-Token geometry stored in $\mathcal{G}_t^{\text{CST}}$ (Sec.~\ref{sec:overview}).

\section{Experiments}
\label{sec:experiments}

\begin{figure}[t]
    \centering
    \includegraphics[width=\textwidth]{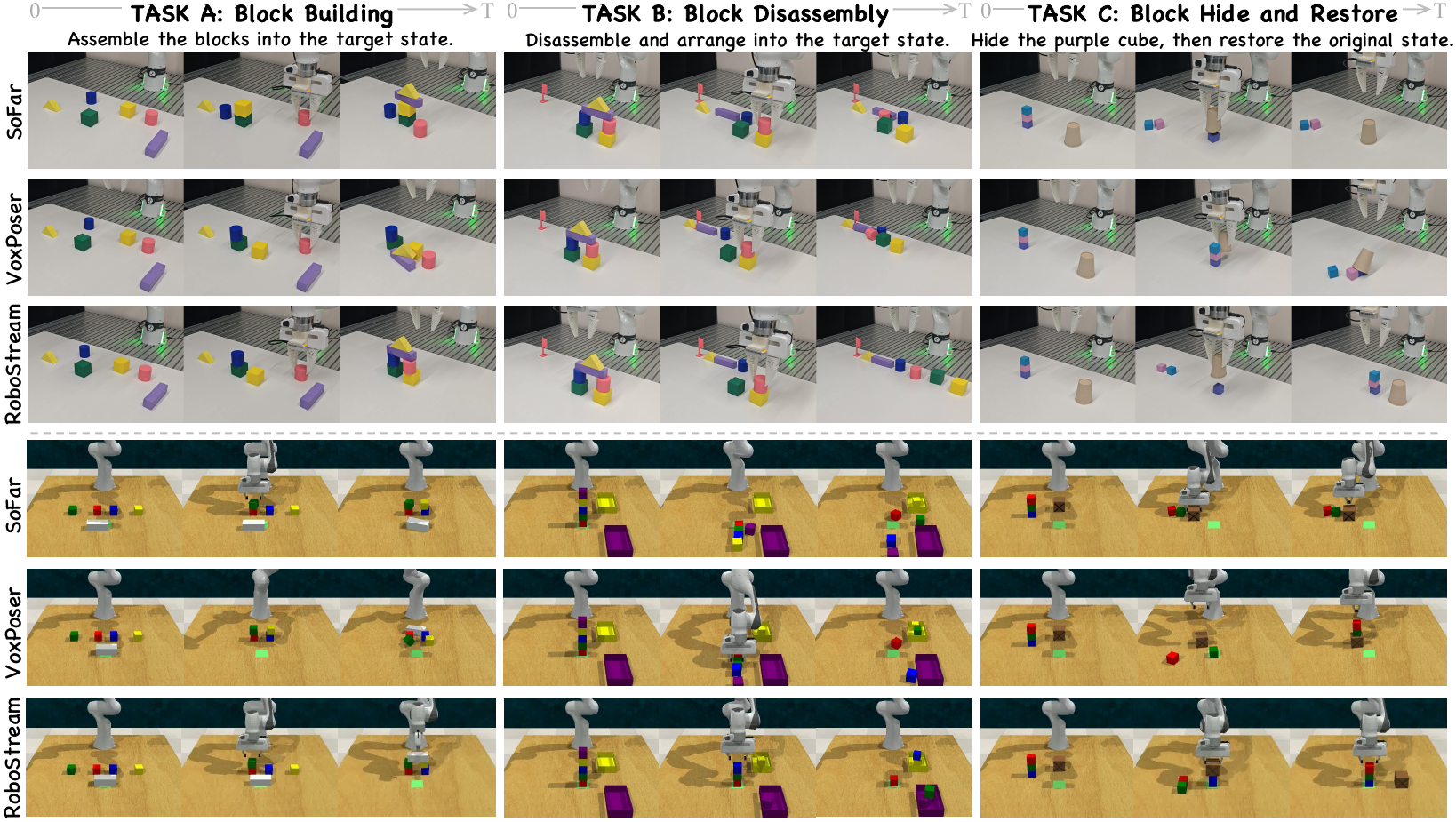}
    \caption{\textbf{Qualitative results in real-world and simulated manipulation tasks.} We evaluate on three challenging tasks: (a) Block Building, demanding precise bottom-to-top assembly toward a target configuration; (b) Block Disassembly, requiring structured top-to-bottom deconstruction and rearrangement; and (c) Block Hide and Restore, requiring causal memory maintenance under full occlusion. RoboStream consistently outperforms SoFar~\cite{qisofar} and VoxPoser~\cite{huang2023voxposer}, particularly in spatio-temporal perception and causal memory over extended horizons.}
    \label{fig:step_setting}

\end{figure}

\subsection{Experimental Setup}
\label{sec:setup}

We evaluate RoboStream across five benchmarks spanning physical and simulated environments, covering real-world manipulation on a Franka Research 3 arm (Sec.~\ref{sec:real_world}), long-horizon tasks on RLBench~\cite{james2020rlbench} (Sec.~\ref{sec:rlbench}), zero-shot generalization on SIMPLER~\cite{li2025evaluating} (Sec.~\ref{sec:simpler}), and 6-DoF spatial reasoning on SpatialBench~\cite{qisofar} and Open6DOR V2~\cite{ding2024open6dor} (Sec.~\ref{sec:spatial_reasoning}). Short-horizon evaluation validates that the geometry grounding of STF-Tokens suppresses spatial hallucination in individual actions; long-horizon evaluation tests whether the full framework sustains causal state tracking and correct action ordering. We instantiate RoboStream with three VLM backbones, Qwen3-VL-8B, Qwen3-VL-32B, and Qwen3-VL-235B~\cite{yang2025qwen3}, hereafter referred to as RoboStream-8B, RoboStream-32B, and RoboS\-tream-235B respectively. All variants operate without environment-specific adaptation or fine-tuning across any benchmark. We compare against state-of-the-art baselines including SoFar~\cite{qisofar}, VoxPoser~\cite{huang2023voxposer}, and RT-2-X~\cite{zitkovich2023rt}, measuring execution success rate for manipulation tasks and accuracy for spatial reasoning.

To accommodate task diversity, we adopt two goal-specification strategies. For real-world experiments and RLBench long-horizon tasks, where target configurations are difficult to fully capture in text, we employ image-guided goal specification with images depicting the desired final state, such as blocks arranged in a target tower configuration. For SIMPLER short-horizon tasks, and tasks involving occlusion or state restoration, we employ text-guided natural language instructions, such as ``hide the blue cube, then restore the original state.''

\begin{figure}[t]
   
    \centering
    \includegraphics[width=1\textwidth]{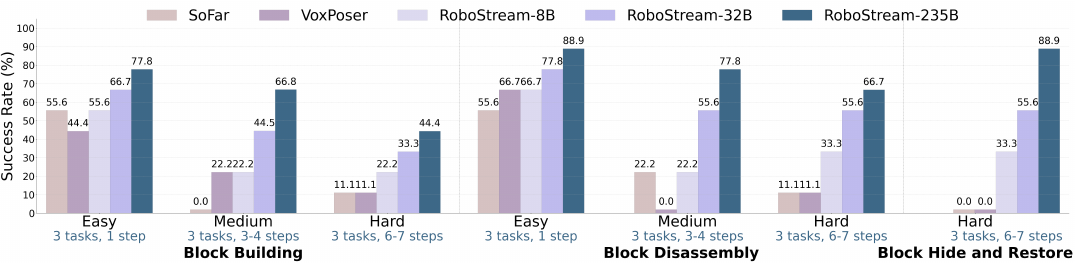}
    \caption{\textbf{Performance comparison on zero-shot real-world manipulation tasks.} We design 21 diverse tasks covering block building (Task A), block disassembly (Task B), and block hide and restore (Task C).}
    \label{fig:real_world}
\end{figure}

\subsection{Real-World Robotic Manipulation}
\label{sec:real_world}
\noindent{\textbf{Tasks and Evaluation.}}~
We construct 21 real-world tasks on a Franka Research 3 (FR3) arm equipped with a parallel gripper and a front-view Intel RealSense D435i RGB-D camera, utilizing 17 colored objects, as shown in Fig.~\ref{fig:step_setting}. The tasks are categorized into three types evaluating different cognitive and physical reasoning dimensions. Task A (Block Building) assesses spatio-temporal grounding and precondition adherence by stacking objects bottom-to-top across three difficulty levels (Easy, Medium, Hard), requiring the robot to verify base stability before proceeding. Task B (Block Disassembly) evaluates sequential state tracking and support relation maintenance during top-to-bottom removal, requiring accurate causal memory of stacking order and support dependencies. Task C (Block Hide and Restore) stress-tests causal memory and object permanence under full occlusion, where the robot must conceal a block beneath a cup, execute distractor steps, and restore the original configuration by retrieving object identity and last known pose from causal history. Tasks A and B are specified by goal images, while Task C is guided by natural language instructions. Each task is executed three times, with details in the Appendix.

\noindent{\textbf{Results.}}~
As shown in Fig.~\ref{fig:real_world}, all RoboStream variants consistently outperform baselines across all difficulty levels, with success rates scaling proportionally with backbone capacity. On Hard-level block building and disassembly tasks, RoboStream-235B achieves success rates of 44.4\% and 66.7\%, respectively. In the hard block-building setting, SoFar~\cite{qisofar} and VoxPoser~\cite{huang2023voxposer} struggle to exceed 11.1\% without persistent causal tracking and physical state evolution awareness. In Task C (Block Hide and Restore), requiring accurate tracking under total occlusion, RoboStream-235B achieves 88.9\% while even our lightweight 8B variant maintains 33.3\%, whereas SoFar and VoxPoser completely fail at 0\%. This contrast highlights that structured spatio-temporal memory is far more decisive for hidden-state tasks than scaling reactive systems. The monotonic gains across model scales confirm that STF-Tokens and CSTG provide complementary capabilities growing increasingly effective with model capacity, yielding robustness that purely reactive systems cannot replicate.

\begin{table*}[t]
    \centering
    \caption{\textbf{Quantitative success rate comparison on the SIMPLER benchmark}. We evaluate: (a) the Google Robot setup under Variant Aggregation (VA) and Visual Matching (VM) protocols, and (b) the WidowX + Bridge setup. For the WidowX tasks, we use the following abbreviations: Spoon (Put Spoon on Towel), Carrot (Put Carrot on Plate), Stack (Stack Green Block on Yellow Block), and Egg (Put Eggplant in Basket). ``Zero'' in the Data column indicates zero-shot evaluation. }
    \label{tab:simpler_env}

    \begin{minipage}[htbp]{0.42\linewidth}
        \centering
        \textbf{(a) Google Robot Setup}
        \vspace{2pt}

        \resizebox{\linewidth}{!}{%
        \renewcommand{\arraystretch}{1.1}
            \begin{tabular}{llcc}
                \toprule
                \textbf{Set} & \textbf{Policy} & \textbf{Pick Coke} & \textbf{Move Near} \\
                \midrule

                \multirow{6}{*}{\textbf{VA}}
                & RT-1-X~\cite{brohan2022rt}     & 49.0 & 32.3 \\
                & RT-2-X~\cite{zitkovich2023rt}     & 82.3 & 79.2 \\
                & Octo-B~\cite{team2024octo}     & 0.6  & 3.1 \\
                & OpenVLA~\cite{kim2025openvla}    & 54.5 & 47.7 \\
                & SoFar~\cite{qisofar}      & \cellcolor{secondpurple}\underline{90.7} & \cellcolor{secondpurple}\underline{74.0} \\
                & \textbf{RoboStream-8B} & \cellcolor{firstpurple}\textbf{94.2} & \cellcolor{firstpurple}\textbf{79.8} \\
                \midrule

                \multirow{6}{*}{\textbf{VM}}
                & RT-1-X~\cite{brohan2022rt}     & 56.7 & 31.7 \\
                & RT-2-X~\cite{zitkovich2023rt}     & 78.7 & 77.9 \\
                & Octo-B~\cite{team2024octo}     & 17.0 & 4.2 \\
                & OpenVLA~\cite{kim2025openvla}    & 16.3 & 46.2 \\
                & SoFar~\cite{qisofar}      & \cellcolor{secondpurple}\underline{92.3} & \cellcolor{secondpurple}\underline{91.7} \\
                & \textbf{RoboStream-8B} & \cellcolor{firstpurple}\textbf{95.7} & \cellcolor{firstpurple}\textbf{95.8} \\
                \bottomrule
            \end{tabular}
        }
    \end{minipage}
    \hfill
    \begin{minipage}[htbp]{0.56\linewidth}
        \centering
        \textbf{(b) WidowX + Bridge Setup}
        \vspace{2pt}

        \renewcommand{\arraystretch}{1.11}
        \resizebox{\linewidth}{!}{%
            \begin{tabular}{llccccc}
                \toprule
                \textbf{Policy} & \textbf{Data} & \textbf{Spoon} & \textbf{Carrot} & \textbf{Stack} & \textbf{Egg} & \textbf{Avg} \\
                \midrule
                RT-1-X~\cite{brohan2022rt}      & OXE    & 0.0 & 4.2 & 0.0 & 0.0 & 1.1 \\
                Octo-B~\cite{team2024octo}      & OXE    & 12.5 & 8.3 & 0.0 & 43.1 & 16.0 \\
                Octo-S~\cite{team2024octo}      & OXE    & 47.2 & 9.7 & 4.2 & 56.9 & 30.0 \\
                OpenVLA~\cite{kim2025openvla}    & OXE    & 0.0 & 0.0 & 0.0 & 4.1 & 1.0 \\
                RoboVLM~\cite{li2026robovlm}     & OXE    & 20.8 & 25.0 & 8.3 & 0.0 & 13.5 \\
                RoboVLM~\cite{li2026robovlm}     & Bridge & 29.2 & 25.0 & 12.5 & 58.3 & 31.3 \\
                SpatialVLA~\cite{qu2025spatialvla}  & OXE    & 20.8 & 20.8 & 25.0 & 70.8 & 34.4 \\
                SpatialVLA~\cite{qu2025spatialvla}  & Bridge & 16.7 & 25.0 & 29.2 & \cellcolor{firstpurple}\textbf{100.0} & 42.7 \\
                SoFar~\cite{qisofar}       & Zero   & \cellcolor{secondpurple}\underline{58.3} & \cellcolor{secondpurple}\underline{66.7} & \cellcolor{secondpurple}\underline{70.8} & 37.5 & \cellcolor{secondpurple}\underline{58.3} \\
                \midrule
                \textbf{RoboStream-8B} & \textbf{Zero} & \cellcolor{firstpurple}\textbf{62.5} & \cellcolor{firstpurple}\textbf{71.9} & \cellcolor{firstpurple}\textbf{77.1} & \cellcolor{secondpurple}\underline{87.5} & \cellcolor{firstpurple}\textbf{74.8} \\
                \bottomrule
            \end{tabular}
        }
    \end{minipage}
\end{table*}

\subsection{Simulation Object Manipulation Evaluation on SIMPLER}
\label{sec:simpler}

We evaluate zero-shot cross-embodiment generalization on the SIMPLER benchmark~\cite{li2025evaluating} across Google Robot and WidowX + Bridge configurations, representing substantial domain shifts from training environments. These short-horizon tasks isolate STF-Token contributions by testing whether spatial grounding transfers across embodiments and viewpoints, independent of long-horizon reasoning.

\noindent{\textbf{Google Robot Configuration.}}~
As shown in Tab.~\ref{tab:simpler_env}, RoboStream-8B achieves superior success rates on both ``Pick Coke Can'' and the spatially demanding ``Move Near'' tasks under VM evaluation, surpassing the prior state-of-the-art SoFar~\cite{qisofar}. Notably, RoboStream-8B also outperforms OXE-finetuned RT-2-X~\cite{zitkovich2023rt} by a substantial margin despite operating fully zero-shot. Under the more challenging VA evaluation, RoboStream-8B maintains competitive performance, again exceeding SoFar. Consistent gains across both protocols and tasks with varying geometry confirm stable spatial grounding independent of embodiment.

\noindent{\textbf{WidowX + Bridge Configuration.}}~
RoboStream achieves a higher average success rate across all subtasks compared to zero-shot SoFar and Bridge-finetuned SpatialVLA~\cite{qu2025spatialvla} without requiring any fine-tuning or domain adaptation. Notably, this advantage holds despite SpatialVLA being explicitly trained on in-domain Bridge data, highlighting the superior transferability of our geometry-grounded representation. These results confirm that spatio-temporal reasoning grounded in explicit geometric primitives transfers robustly across embodiments and extends naturally from short tasks to subsequent long-horizon benchmarks.

\subsection{Simulation Long-Horizon Task Evaluation on RLBench}
\label{sec:rlbench}

\begin{table}[t]
\centering
\caption{\textbf{Simulation evaluation of long-horizon manipulation on RLBench.} All RoboStream variants are evaluated in a zero-shot manner without in-domain fine-tuning. }
\label{tab:full_vertical_compact}
\resizebox{\textwidth}{!}{
\begin{tabular}{l ccccc}
\toprule
\textbf{Tasks} & 
\textbf{SoFar}~\cite{qisofar} & 
\textbf{VoxPoser}~\cite{huang2023voxposer} & 
\textbf{RoboStream-8B} & 
\textbf{RoboStream-32B} & 
\textbf{RoboStream-235B} \\
\midrule
Bridge Between Towers            & 52.0 & 8.0 & \cellcolor{secondpurple}\underline{76.0} & \cellcolor{firstpurple}\textbf{88.0} & \cellcolor{firstpurple}\textbf{88.0} \\
Cover Top Block with Box         & 4.0 & 4.0 & 40.0 & \cellcolor{secondpurple}\underline{84.0} & \cellcolor{firstpurple}\textbf{92.0} \\
Cover Bottom Block with Box      & 0.0 & 0.0 & 16.0 & \cellcolor{secondpurple}\underline{68.0} & \cellcolor{firstpurple}\textbf{96.0} \\
Place Blocks in Two Containers   & \cellcolor{firstpurple}\textbf{100.0} & \cellcolor{secondpurple}\underline{96.0} & \cellcolor{firstpurple}\textbf{100.0} & \cellcolor{firstpurple}\textbf{100.0} & \cellcolor{firstpurple}\textbf{100.0} \\
Place Blocks in Two Cont. (Hard) & 4.0 & 4.0 & 24.0 & \cellcolor{secondpurple}\underline{76.0} & \cellcolor{firstpurple}\textbf{88.0} \\
Stack Five Colors                & 4.0 & 4.0 & 60.0 & \cellcolor{secondpurple}\underline{72.0} & \cellcolor{firstpurple}\textbf{80.0} \\
Stack Three Colors               & \cellcolor{secondpurple}\underline{60.0} & \cellcolor{firstpurple}\textbf{96.0} & \cellcolor{firstpurple}\textbf{96.0} & \cellcolor{firstpurple}\textbf{96.0} & \cellcolor{firstpurple}\textbf{96.0} \\
Unstack then Stack Four Colors   & 0.0 & 0.0 & 52.0 & \cellcolor{secondpurple}\underline{76.0} & \cellcolor{firstpurple}\textbf{84.0} \\
\midrule
\textbf{Average}                     & 28.0 & 26.5 & 58.0 & \cellcolor{secondpurple}\underline{82.5} & \cellcolor{firstpurple}\textbf{90.5} \\
\bottomrule
\end{tabular}
}
\end{table}

\noindent{\textbf{Tasks and Evaluation.}}~
To assess long-horizon manipulation capabilities, we introduce 8 complex tasks within RLBench~\cite{james2020rlbench} (Tab.~\ref{tab:full_vertical_compact}), constructed by extending and reconfiguring native assets. Each method is evaluated over 25 episodes per task using different random seeds. Following Sec.~\ref{sec:setup}, tasks are categorized by guidance modality. Five goal-image-guided tasks require visually complex target configurations, including Bridge Between Towers, Place Blocks in Two Containers, Place Blocks in Two Cont. (Hard), and Stack Three/Five Colors. Three text-guided tasks stress-test action ordering, restoration, and object permanence under occlusion, including Cover Top/Bottom Block with Box and Unstack then Stack Four Colors. Detailed setups are provided in the Appendix.

\noindent{\textbf{Results.}}~
As shown in Tab.~\ref{tab:full_vertical_compact}, RoboStream consistently outperforms both SoFar~\cite{qisofar} and VoxPoser~\cite{huang2023voxposer} across these long-horizon scenarios. Notably, our framework demonstrates a strong scaling effect, with performance increasing alongside model capacity to reach a 90.5\% average success rate on RoboStream-235B without additional training, as larger capacity provides the reasoning depth necessary to maintain logical consistency over extended action sequences. However, scaling the VLM alone is not a panacea for long-horizon stability. Even our smallest variant, RoboStream-8B, substantially exceeds much larger baselines, demonstrating that spatio-temporal reasoning and structured memory are more critical for physical consistency than simply increasing parameter scale. We further compare against memory-augmented approaches SAM2Act~\cite{fang2025sam2act} and MemoryVLA~\cite{shi2026memoryvla} on four RLBench long-horizon cases; RoboStream outperforms them across all model scales, as detailed in \cref{sec:appendix_memory_baselines}.

\begin{table}[t]
    \centering
    \begin{minipage}[t]{0.56\textwidth} 
        \centering
        \vspace{0em}
        \setlength{\tabcolsep}{2.5pt} 
        \caption{\textbf{Spatial reasoning evaluation on 6-DoF SpatialBench~\cite{qisofar}.} Results are categorized by spatial VQA task types, covering absolute ($abs.$) and relative ($rel.$) reasoning for both position and orientation.}
        \label{tab:spatial_reasoning_final}
        \scriptsize
        \begin{tabular}{@{}lccccc@{}}
        \toprule
        \multirow{2}{*}{Method} & \multicolumn{2}{c}{Position} & \multicolumn{2}{c}{Orientation} & \multirow{2}{*}{Total} \\
        \cmidrule(lr){2-3} \cmidrule(lr){4-5}
        & rel. & abs. & rel. & abs. & \\
        \midrule
        GPT-4o~\cite{hurst2024gpt}      & 49.4 & 28.4 & 44.2 & 25.8 & 36.2 \\
        SpaceMantis~\cite{chen2024spatialvlm} & 33.6 & 29.2 & 27.2 & 25.0 & 28.9 \\
        RoboPoint~\cite{yuan2025robopoint}   & 43.8 & 30.8 & 33.8 & 25.8 & 33.5 \\
        SoFar~\cite{qisofar} & 59.6 & 33.8 & \cellcolor{firstpurple}\textbf{54.6} & 31.3 & 43.9 \\
        SoFar-8B~\cite{qisofar} & 41.5 & 24.3 & 27.1 & 31.3 & 30.5 \\
        SoFar-32B~\cite{qisofar} & \cellcolor{secondpurple}\underline{60.4} & \cellcolor{secondpurple}\underline{41.9} & 45.8 & 33.3 & \cellcolor{secondpurple}\underline{45.3} \\
        \textbf{RoboStream-8B} & 47.2 & 31.1 & 35.4 & \cellcolor{secondpurple}\underline{35.4} & 36.8 \\
        \textbf{RoboStream-32B} & \cellcolor{firstpurple}\textbf{66.0} & \cellcolor{firstpurple}\textbf{44.6} & \cellcolor{secondpurple}\underline{48.0} & \cellcolor{firstpurple}\textbf{37.5} & \cellcolor{firstpurple}\textbf{48.9} \\
        \bottomrule
        \end{tabular}
    \end{minipage}
    \hfill 
    \begin{minipage}[t]{0.42\textwidth} 
        \centering
        \setlength{\tabcolsep}{2pt} 
        \renewcommand{\arraystretch}{1.18} %
        \caption{\textbf{6-DoF object rea\-r\-r\-a\-n\-g\-e\-m\-e\-n\-t evaluation on O\-p\-e\-n\-6\-D\-O\-R~\cite{ding2024open6dor}.} Results are categorized by position ($Pos.$), rotation ($Rot.$), and integrated 6-DoF success rates. }
        \label{tab:open6dor_updated}
        \scriptsize 
        \begin{tabular}{@{}lccc@{}}
        \toprule
        Method & Pos.  & Rot.  & 6-DoF  \\
        \midrule
        Dream2Real~\cite{kapelyukh2024dream2real}         & 15.9 & 31.3 & 13.5 \\
        VoxPoser~\cite{huang2023voxposer}            & 32.6 & -    & -    \\
        Open6DOR-GPT~\cite{ding2024open6dor}       & 74.9 & 41.1 & 35.6 \\
        SoFar~\cite{qisofar}          & 93.0 & \cellcolor{secondpurple}\underline{57.0} & \cellcolor{secondpurple}\underline{48.7} \\
        SoFar-8B~\cite{qisofar}               & 92.4 & 42.9 & 45.1 \\
        SoFar-32B~\cite{qisofar}              & \cellcolor{secondpurple}\underline{93.1} & 54.6 & 48.4 \\
        \textbf{RoboStream-8B}          & \cellcolor{secondpurple}\underline{93.1} & 47.6 & 47.3 \\
        \textbf{RoboStream-32B}& \cellcolor{firstpurple}\textbf{93.8} & \cellcolor{firstpurple}\textbf{58.8} & \cellcolor{firstpurple}\textbf{52.2} \\
        \bottomrule
        \end{tabular}
    \end{minipage}
\end{table}

\subsection{6-DoF Spatial Reasoning Evaluation}
\label{sec:spatial_reasoning}
To evaluate how STF-Tokens bridge high-level visual features with precise instance-level 3D grounding, we extend our evaluation to rotation and full 6-DoF pose estimation. Following SoFar~\cite{qisofar}, we incorporate PointSO for 6-DoF prediction and evaluate on Open6DOR V2~\cite{ding2024open6dor} and 6-DoF SpatialBench~\cite{qisofar}. Since SoFar serves as our primary baseline, we include variants using Qwen3-VL-8B and Qwen3-VL-32B backbones for fair cross-scale comparison with RoboStream.

\noindent{\textbf{Spatial Reasoning on 6-DoF SpatialBench.}}
As shown in Tab.~\ref{tab:spatial_reasoning_final}, RoboStream-32B achieves superior overall performance, outperforming both SoFar and GPT-4o~\cite{hurst2024gpt}. The most pronounced gains appear on absolute and relative position tasks, where pure image-text alignment is most susceptible to spatial ambiguity. These results confirm that the core advantage of STF-Tokens lies not in coordinate injection alone, but in serving as a geometric anchor providing deterministic spatial reference when visual features and language descriptions conflict.

\noindent{\textbf{6-DoF Object Rearrangement on Open6DOR V2.}}
As shown in Tab.~\ref{tab:open6dor_updated}, RoboStream-32B consistently outperforms SoFar across position, rotation, and integrated 6-DoF metrics. Unlike baselines treating spatial coordinates as text annotations, STF-Tokens distill per-object visual evidence into structured instance representations encoding centroid and Gaussian shape. This shift from pixel-level to object-level reasoning underlies consistent gains across all tracks, most notably on 6-DoF where tight placement constraints demand precise geometric grounding and spatial alignment.

\begin{table*}[t]
\centering
\caption{\textbf{Ablation study on RoboStream-235B.} Success rates (\%) across 8 long-horizon tasks, isolating contributions of CSTG and STF-Tokens.}
\label{tab:ablation_235b_horizontal}
\resizebox{\textwidth}{!}{%
\renewcommand{\arraystretch}{1.2} 
\begin{tabular}{cc | ccccccccc}
\toprule
\thead{\textit{STF-} \\ \textit{Tokens}} & \thead{\textit{CSTG}} &
 \thead{Bridge Between\\Towers} & 
 \thead{Cover Top\\Block} & 
 \thead{Cover Bottom\\Block} & 
 \thead{Place in\\Containers} & 
 \thead{Place in\\Containers (Hard)} & 
 \thead{Stack Five\\Colors} & 
 \thead{Stack Three\\Colors} & 
 \thead{Unstack then\\Stack} & 
 \textbf{Average} \\
\midrule
\xmark & \xmark & 0.0 & 0.0 & 0.0 & 84.0 & 12.0 & 0.0 & 0.0 & 0.0 & 12.0 \\
\cmark & \xmark & 0.0 & 0.0 & 0.0 & \cellcolor{secondpurple}\underline{92.0} & 24.0 & 0.0 & 0.0 & 0.0 & 14.5 \\
\xmark & \cmark & \cellcolor{secondpurple}\underline{72.0} & \cellcolor{secondpurple}\underline{84.0} & \cellcolor{secondpurple}\underline{88.0} & \cellcolor{firstpurple}\textbf{100.0} & \cellcolor{secondpurple}\underline{72.0} & \cellcolor{secondpurple}\underline{60.0} & \cellcolor{firstpurple}\textbf{96.0} & \cellcolor{secondpurple}\underline{64.0} & \cellcolor{secondpurple}\underline{79.5} \\
\cmark & \cmark & \cellcolor{firstpurple}\textbf{88.0} & \cellcolor{firstpurple}\textbf{92.0} & \cellcolor{firstpurple}\textbf{96.0} & \cellcolor{firstpurple}\textbf{100.0} & \cellcolor{firstpurple}\textbf{88.0} & \cellcolor{firstpurple}\textbf{80.0} & \cellcolor{firstpurple}\textbf{96.0} & \cellcolor{firstpurple}\textbf{84.0} & \cellcolor{firstpurple}\textbf{90.5} \\
\bottomrule
\end{tabular}%
}
\end{table*}

\subsection{Ablation Study}
\label{sec:ablation_study}

We ablate RoboStream-235B across all 8 long-horizon RLBench~\cite{james2020rlbench} tasks to isolate the contribution of each architectural component (Tab.~\ref{tab:ablation_235b_horizontal}). Removing the CSTG module (\textit{w/o CSTG}) causes the average success rate to collapse from 90.5\% to 14.5\%, confirming that persistent state tracking is the foundational driver for long-horizon tasks. However, while CSTG provides logical structure, STF-Tokens provide physical precision. Removing STF-Tokens (\textit{w/o STF-Tokens}) results in an 11.0\% absolute drop (from 90.5\% to 79.5\%), demonstrating that without explicit 3D geometric grounding, even a CSTG-equipped agent struggles with compounding spatial inaccuracies during contact-rich execution. The mutual dependence of both components becomes evident when removing them simultaneously (\textit{w/o Both}), which yields only 12.0\% success rate. Notably, comparing this to the \textit{w/o CSTG} variant (14.5\%) shows that STF-Tokens still provide marginal improvement in short-horizon spatial accuracy even when long-horizon logic fails. Their full potential, however, is unlocked only when coupled with the CSTG, where CSTG ensures correct action sequencing and STF-Tokens ensure each step is executed with deterministic spatial precision.


\section{Limitations}
\label{sec:limitation}
As a decoupled planning-execution framework, RoboStream shares a limitation inherent to the broader VLM-for-robotics paradigm. Imperfections in any sub-module, including unstable grasping or perceptual uncertainty, can propagate into execution failures even when high-level planning is correct. This reflects a fundamental tension between the semantic generalization of VLM-based planners and the physical precision demanded by low-level control, a challenge broadly applicable to decoupled systems beyond RoboStream. Future work includes exploring hybrid approaches that bridge VLM planner generalization with end-to-end execution fluency, either by adopting a Vision-Language-Action (VLA) model as the downstream controller for smoother and more compliant motion, or by developing a fully end-to-end VLA that internalizes spatio-temporal reasoning and causal memory within the action generation loop. Extending RoboStream to contact-rich, dexterous, and open-world manipulation also constitutes a promising direction.

\section{Conclusion}
\label{sec:conclusion}
RoboStream addresses two key deficits behind long-horizon VLM planning failures: ungrounded spatial perception and untracked causal history. STF-Tokens convert per-object appearance and 3D geometry into identity-persistent primitives, while the CSTG records action-induced state transitions and preserves hidden object states under occlusion. Across RLBench, SIMPLER, SpatialBench, Open6DOR V2, and real-world Franka experiments, RoboStream consistently improves long-horizon stability, showing that explicit spatio-temporal grounding and persistent memory are more decisive than model scaling alone.

\noindent\textbf{Acknowledgments.}
This work is supported by the National Key R\&D Program of China (2022ZD0161700), the National Natural Science Foundation of China (Grant Nos. 92467204 and 62472249), and the Shenzhen Science and Technology Program (Grant No. KJZD20240903102300001). This study was also supported by the Shenzhen Science and Technology Program (Grant No. JCYJ20250604145014018) and the Natural Science Foundation of Top Talent of SZTU (Grant No. GDRC202413). We would like to thank YuanxingGuangnian Robotics for sponsoring this research.

{
    \small
    \bibliographystyle{splncs04}
    \bibliography{main}

@String(CVPR  = {IEEE Conf. Comput. Vis. Pattern Recog.})

@String(CVPR  = {CVPR})

@inproceedings{li2025evaluating,
  title={Evaluating Real-World Robot Manipulation Policies in Simulation},
  author={Li, Xuanlin and Hsu, Kyle and Gu, Jiayuan and Mees, Oier and Pertsch, Karl and Walke, Homer Rich and Fu, Chuyuan and Lunawat, Ishikaa and Sieh, Isabel and Kirmani, Sean and others},
  booktitle={Conference on Robot Learning},
  pages={3705--3728},
  year={2025},
  organization={PMLR}
}

@inproceedings{zitkovich2023rt,
  title={Rt-2: Vision-language-action models transfer web knowledge to robotic control},
  author={Zitkovich, Brianna and Yu, Tianhe and Xu, Sichun and Xu, Peng and Xiao, Ted and Xia, Fei and Wu, Jialin and Wohlhart, Paul and Welker, Stefan and Wahid, Ayzaan and others},
  booktitle={Conference on Robot Learning},
  pages={2165--2183},
  year={2023},
  organization={PMLR}
}

@inproceedings{kim2025openvla,
  title={OpenVLA: An Open-Source Vision-Language-Action Model},
  author={Kim, Moo Jin and Pertsch, Karl and Karamcheti, Siddharth and Xiao, Ted and Balakrishna, Ashwin and Nair, Suraj and Rafailov, Rafael and Foster, Ethan P and Sanketi, Pannag R and Vuong, Quan and others},
  booktitle={Conference on Robot Learning},
  pages={2679--2713},
  year={2025},
  organization={PMLR}
}

@inproceedings{zhen20243d,
  title={3D-VLA: A 3D Vision-Language-Action Generative World Model},
  author={Zhen, Haoyu and Qiu, Xiaowen and Chen, Peihao and Yang, Jincheng and Yan, Xin and Du, Yilun and Hong, Yining and Gan, Chuang},
  booktitle={International Conference on Machine Learning},
  pages={61229--61245},
  year={2024},
  organization={PMLR}
}

@inproceedings{shridhar2023perceiver,
  title={Perceiver-actor: A multi-task transformer for robotic manipulation},
  author={Shridhar, Mohit and Manuelli, Lucas and Fox, Dieter},
  booktitle={Conference on Robot Learning},
  pages={785--799},
  year={2023},
  organization={PMLR}
}

@inproceedings{wu2023tidybot,
  title={TidyBot: Personalized Robot Assistance with Large Language Models},
  author={Wu, Jimmy and Antonova, Rika and Kan, Adam and Lepert, Marion and Zeng, Andy and Song, Shuran and Bohg, Jeannette and Rusinkiewicz, Szymon and Funkhouser, Thomas},
  booktitle={2023 IEEE/RSJ International Conference on Intelligent Robots and Systems (IROS)},
  pages={3546--3553},
  year={2023},
  organization={IEEE}
}

@inproceedings{jiang2023vima,
  title={VIMA: General Robot Manipulation with Multimodal Prompts},
  author={Jiang, Yunfan and Gupta, Agrim and Zhang, Zichen and Wang, Guanzhi and Dou, Yongqiang and Chen, Yanjun and Fei-Fei, Li and Anandkumar, Anima and Zhu, Yuke and Fan, Linxi},
  booktitle={International Conference on Machine Learning},
  pages={14975--15022},
  year={2023},
  organization={PMLR}
}

@article{lake2017building,
  title={Building machines that learn and think like people},
  author={Lake, Brenden M and Ullman, Tomer D and Tenenbaum, Joshua B and Gershman, Samuel J},
  journal={Behavioral and brain sciences},
  volume={40},
  pages={e253},
  year={2017},
  publisher={Cambridge University Press}
}

@inproceedings{qin2025robofactory,
  title={Robofactory: Exploring embodied agent collaboration with compositional constraints},
  author={Qin, Yiran and Kang, Li and Song, Xiufeng and Yin, Zhenfei and Liu, Xiaohong and Liu, Xihui and Zhang, Ruimao and Bai, Lei},
  booktitle={Proceedings of the IEEE/CVF International Conference on Computer Vision},
  pages={10075--10085},
  year={2025}
}

@inproceedings{zhudexflywheel,
  title={DexFlyWheel: A Scalable and Self-improving Data Generation Framework for Dexterous Manipulation},
  author={Zhu, Kefei and Bai, Fengshuo and Xiang, YuanHao and Cai, Yishuai and Chen, Xinglin and Li, Ruochong and Wang, Xingtao and Dong, Hao and Yang, Yaodong and Fan, Xiaopeng and others},
  booktitle={The Thirty-ninth Annual Conference on Neural Information Processing Systems},
  year={2025}
}

@article{tan2025roboos,
  title={Roboos: A hierarchical embodied framework for cross-embodiment and multi-agent collaboration},
  author={Tan, Huajie and Hao, Xiaoshuai and Chi, Cheng and Lin, Minglan and Lyu, Yaoxu and Cao, Mingyu and Liang, Dong and Chen, Zhuo and Lyu, Mengsi and Peng, Cheng and others},
  journal={arXiv preprint arXiv:2505.03673},
  year={2025}
}

@article{cheng2024spatialrgpt,
  title={Spatialrgpt: Grounded spatial reasoning in vision-language models},
  author={Cheng, An-Chieh and Yin, Hongxu and Fu, Yang and Guo, Qiushan and Yang, Ruihan and Kautz, Jan and Wang, Xiaolong and Liu, Sifei},
  journal={Advances in Neural Information Processing Systems},
  volume={37},
  pages={135062--135093},
  year={2024}
}

@inproceedings{song2025robospatial,
  title={Robospatial: Teaching spatial understanding to 2d and 3d vision-language models for robotics},
  author={Song, Chan Hee and Blukis, Valts and Tremblay, Jonathan and Tyree, Stephen and Su, Yu and Birchfield, Stan},
  booktitle={Proceedings of the Computer Vision and Pattern Recognition Conference},
  pages={15768--15780},
  year={2025}
}

@inproceedings{daxberger2025mm,
  title={Mm-spatial: Exploring 3d spatial understanding in multimodal llms},
  author={Daxberger, Erik and Wenzel, Nina and Griffiths, David and Gang, Haiming and Lazarow, Justin and Kohavi, Gefen and Kang, Kai and Eichner, Marcin and Yang, Yinfei and Dehghan, Afshin and others},
  booktitle={Proceedings of the IEEE/CVF International Conference on Computer Vision},
  pages={7395--7408},
  year={2025}
}

@inproceedings{raysat,
  title={SAT: Dynamic Spatial Aptitude Training for Multimodal Language Models},
  author={Ray, Arijit and Duan, Jiafei and Brown II, Ellis L and Tan, Reuben and Bashkirova, Dina and Hendrix, Rose and Ehsani, Kiana and Kembhavi, Aniruddha and Plummer, Bryan A and Krishna, Ranjay and others},
  booktitle={Second Conference on Language Modeling},
  year={2025}
}

@inproceedings{wang2025spatial457,
  title={Spatial457: A Diagnostic Benchmark for 6D Spatial Reasoning of Large Multimodal Models},
  author={Wang, Xingrui and Ma, Wufei and Zhang, Tiezheng and de Melo, Celso M and Chen, Jieneng and Yuille, Alan},
  booktitle={2025 IEEE/CVF Conference on Computer Vision and Pattern Recognition (CVPR)},
  pages={24669--24679},
  year={2025},
  organization={IEEE}
}

@inproceedings{cai2025spatialbot,
  title={Spatialbot: Precise spatial understanding with vision language models},
  author={Cai, Wenxiao and Ponomarenko, Iaroslav and Yuan, Jianhao and Li, Xiaoqi and Yang, Wankou and Dong, Hao and Zhao, Bo},
  booktitle={2025 IEEE International Conference on Robotics and Automation (ICRA)},
  pages={9490--9498},
  year={2025},
  organization={IEEE}
}

@article{chen2025spacetools,
  title={SpaceTools: Tool-Augmented Spatial Reasoning via Double Interactive RL},
  author={Chen, Siyi and Uy, Mikaela Angelina and Song, Chan Hee and Ladhak, Faisal and Murali, Adithyavairavan and Qu, Qing and Birchfield, Stan and Blukis, Valts and Tremblay, Jonathan},
  journal={arXiv preprint arXiv:2512.04069},
  year={2025}
}

@inproceedings{chen2024spatialvlm,
  title={Spatialvlm: Endowing vision-language models with spatial reasoning capabilities},
  author={Chen, Boyuan and Xu, Zhuo and Kirmani, Sean and Ichter, Brain and Sadigh, Dorsa and Guibas, Leonidas and Xia, Fei},
  booktitle={Proceedings of the IEEE/CVF Conference on Computer Vision and Pattern Recognition},
  pages={14455--14465},
  year={2024}
}

@inproceedings{huang2022language,
  title={Language models as zero-shot planners: Extracting actionable knowledge for embodied agents},
  author={Huang, Wenlong and Abbeel, Pieter and Pathak, Deepak and Mordatch, Igor},
  booktitle={International conference on machine learning},
  pages={9118--9147},
  year={2022},
  organization={PMLR}
}

@inproceedings{singh2023progprompt,
  title={ProgPrompt: Generating Situated Robot Task Plans using Large Language Models},
  author={Singh, Ishika and Blukis, Valts and Mousavian, Arsalan and Goyal, Ankit and Xu, Danfei and Tremblay, Jonathan and Fox, Dieter and Thomason, Jesse and Garg, Animesh},
  booktitle={2023 IEEE International Conference on Robotics and Automation (ICRA)},
  pages={11523--11530},
  year={2023},
  organization={IEEE}
}

@inproceedings{zengsocratic,
  title={Socratic Models: Composing Zero-Shot Multimodal Reasoning with Language},
  author={Zeng, Andy and Attarian, Maria and Choromanski, Krzysztof Marcin and Wong, Adrian and Welker, Stefan and Tombari, Federico and Purohit, Aveek and Ryoo, Michael S and Sindhwani, Vikas and Lee, Johnny and others},
  booktitle={The Eleventh International Conference on Learning Representations},
  year={2023}
}

@article{huang2023grounded,
  title={Grounded decoding: Guiding text generation with grounded models for embodied agents},
  author={Huang, Wenlong and Xia, Fei and Shah, Dhruv and Driess, Danny and Zeng, Andy and Lu, Yao and Florence, Pete and Mordatch, Igor and Levine, Sergey and Hausman, Karol and others},
  journal={Advances in Neural Information Processing Systems},
  volume={36},
  pages={59636--59661},
  year={2023}
}

@inproceedings{brohan2023saycan,
  title={Do as i can, not as i say: Grounding language in robotic affordances},
  author={Brohan, Anthony and Chebotar, Yevgen and Finn, Chelsea and Hausman, Karol and Herzog, Alexander and Ho, Daniel and Ibarz, Julian and Irpan, Alex and Jang, Eric and Julian, Ryan and others},
  booktitle={Conference on robot learning},
  pages={287--318},
  year={2023},
  organization={PMLR}
}

@inproceedings{liang2023code,
  title={Code as policies: Language model programs for embodied control},
  author={Liang, Jacky and Huang, Wenlong and Xia, Fei and Xu, Peng and Hausman, Karol and Ichter, Brian and Florence, Pete and Zeng, Andy},
  booktitle={2023 IEEE International conference on robotics and automation (ICRA)},
  pages={9493--9500},
  year={2023},
  organization={IEEE}
}

@inproceedings{yu2023language,
  title={Language to Rewards for Robotic Skill Synthesis},
  author={Yu, Wenhao and Gileadi, Nimrod and Fu, Chuyuan and Kirmani, Sean and Lee, Kuang-Huei and Arenas, Montserrat Gonzalez and Chiang, Hao-Tien Lewis and Erez, Tom and Hasenclever, Leonard and Humplik, Jan and others},
  booktitle={Conference on Robot Learning},
  pages={374--404},
  year={2023},
  organization={PMLR}
}

@article{sharma2022correcting,
  title={Correcting robot plans with natural language feedback},
  author={Sharma, Pratyusha and Sundaralingam, Balakumar and Blukis, Valts and Paxton, Chris and Hermans, Tucker and Torralba, Antonio and Andreas, Jacob and Fox, Dieter},
  journal={arXiv preprint arXiv:2204.05186},
  year={2022}
}

@inproceedings{yuan2025robopoint,
  title={RoboPoint: A Vision-Language Model for Spatial Affordance Prediction in Robotics},
  author={Yuan, Wentao and Duan, Jiafei and Blukis, Valts and Pumacay, Wilbert and Krishna, Ranjay and Murali, Adithyavairavan and Mousavian, Arsalan and Fox, Dieter},
  booktitle={Conference on Robot Learning},
  pages={4005--4020},
  year={2025},
  organization={PMLR}
}

@inproceedings{huang2025rekep,
  title={ReKep: Spatio-Temporal Reasoning of Relational Keypoint Constraints for Robotic Manipulation},
  author={Huang, Wenlong and Wang, Chen and Li, Yunzhu and Zhang, Ruohan and Fei-Fei, Li},
  booktitle={Conference on Robot Learning},
  pages={4573--4602},
  year={2025},
  organization={PMLR}
}

@inproceedings{huang2025a3vlm,
  title={A3VLM: Actionable Articulation-Aware Vision Language Model},
  author={Huang, Siyuan and Chang, Haonan and Liu, Yuhan and Zhu, Yimeng and Dong, Hao and Boularias, Abdeslam and Gao, Peng and Li, Hongsheng},
  booktitle={Conference on Robot Learning},
  pages={1675--1690},
  year={2025},
  organization={PMLR}
}

@article{fang2024moka,
  title={Moka: Open-world robotic manipulation through mark-based visual prompting},
  author={Fang, Kuan and Liu, Fangchen and Abbeel, Pieter and Levine, Sergey},
  journal={Robotics: Science and Systems XX},
  year={2024},
  publisher={Robotics: Science and Systems Foundation}
}

@article{du2023vision,
  title={Vision-Language Models as Success Detectors},
  author={Du, Yuqing and Konyushkova, Ksenia and Denil, Misha and Raju, Akhil and Landon, Jessica and Hill, Felix and de Freitas, Nando and Cabi, Serkan},
  journal={arXiv preprint arXiv:2303.07280},
  year={2023}
}

@inproceedings{zhou2025code,
  title={Code-as-monitor: Constraint-aware visual programming for reactive and proactive robotic failure detection},
  author={Zhou, Enshen and Su, Qi and Chi, Cheng and Zhang, Zhizheng and Wang, Zhongyuan and Huang, Tiejun and Sheng, Lu and Wang, He},
  booktitle={Proceedings of the Computer Vision and Pattern Recognition Conference},
  pages={6919--6929},
  year={2025}
}

@inproceedings{driess2023palm,
  title={PaLM-E: An Embodied Multimodal Language Model},
  author={Driess, Danny and Xia, Fei and Sajjadi, Mehdi SM and Lynch, Corey and Chowdhery, Aakanksha and Ichter, Brian and Wahid, Ayzaan and Tompson, Jonathan and Vuong, Quan and Yu, Tianhe and others},
  booktitle={International Conference on Machine Learning},
  pages={8469--8488},
  year={2023},
  organization={PMLR}
}

@inproceedings{huang2023inner,
  title={Inner Monologue: Embodied Reasoning through Planning with Language Models},
  author={Huang, Wenlong and Xia, Fei and Xiao, Ted and Chan, Harris and Liang, Jacky and Florence, Pete and Zeng, Andy and Tompson, Jonathan and Mordatch, Igor and Chebotar, Yevgen and others},
  booktitle={Conference on Robot Learning},
  pages={1769--1782},
  year={2023},
  organization={PMLR}
}

@inproceedings{hu2024look,
  title={Look Before You Leap: Unveiling the Power of GPT-4V in Robotic Vision-Language Planning},
  author={Hu, Yingdong and Lin, Fanqi and Zhang, Tong and Yi, Li and Gao, Yang},
  booktitle={First Workshop on Vision-Language Models for Navigation and Manipulation at ICRA 2024},
  year={2024}
}

@inproceedings{zhouroborefer,
  title={RoboRefer: Towards Spatial Referring with Reasoning in Vision-Language Models for Robotics},
  author={Zhou, Enshen and An, Jingkun and Chi, Cheng and Han, Yi and Rong, Shanyu and Zhang, Chi and Wang, Pengwei and Wang, Zhongyuan and Huang, Tiejun and Sheng, Lu and others},
  booktitle={The Thirty-ninth Annual Conference on Neural Information Processing Systems},
  year={2025}
}

@article{han2025tiger,
  title={TIGeR: Tool-Integrated Geometric Reasoning in Vision-Language Models for Robotics},
  author={Han, Yi and Chi, Cheng and Zhou, Enshen and Rong, Shanyu and An, Jingkun and Wang, Pengwei and Wang, Zhongyuan and Sheng, Lu and Zhang, Shanghang},
  journal={arXiv preprint arXiv:2510.07181},
  year={2025}
}

@inproceedings{qisofar,
  title={SoFar: Language-Grounded Orientation Bridges Spatial Reasoning and Object Manipulation},
  author={Qi, Zekun and Zhang, Wenyao and Ding, Yufei and Dong, Runpei and Yu, XinQiang and Li, Jingwen and Xu, Lingyun and Li, Baoyu and He, Xialin and Fan, Guofan and others},
  booktitle={The Thirty-ninth Annual Conference on Neural Information Processing Systems},
  year={2025}
}

@inproceedings{huang2023voxposer,
  title={VoxPoser: Composable 3D Value Maps for Robotic Manipulation with Language Models},
  author={Huang, Wenlong and Wang, Chen and Zhang, Ruohan and Li, Yunzhu and Wu, Jiajun and Fei-Fei, Li},
  booktitle={Conference on Robot Learning},
  pages={540--562},
  year={2023},
  organization={PMLR}
}

@inproceedings{hanrobocerebra,
  title={RoboCerebra: A Large-scale Benchmark for Long-horizon Robotic Manipulation Evaluation},
  author={Han, Songhao and Qiu, Boxiang and Liao, Yue and Huang, Siyuan and Gao, Chen and YAN, Shuicheng and Liu, Si},
  booktitle={The Thirty-ninth Annual Conference on Neural Information Processing Systems Datasets and Benchmarks Track},
  year={2025}
}

@article{gu2025manualvla,
  title={ManualVLA: A Unified VLA Model for Chain-of-Thought Manual Generation and Robotic Manipulation},
  author={Gu, Chenyang and Liu, Jiaming and Chen, Hao and Huang, Runzhong and Wuwu, Qingpo and Liu, Zhuoyang and Li, Xiaoqi and Li, Ying and Zhang, Renrui and Jia, Peng and others},
  journal={arXiv preprint arXiv:2512.02013},
  year={2025}
}

@inproceedings{huang2024copa,
  title={Copa: General robotic manipulation through spatial constraints of parts with foundation models},
  author={Huang, Haoxu and Lin, Fanqi and Hu, Yingdong and Wang, Shengjie and Gao, Yang},
  booktitle={2024 IEEE/RSJ International Conference on Intelligent Robots and Systems (IROS)},
  pages={9488--9495},
  year={2024},
  organization={IEEE}
}

@inproceedings{depthanything,
  title={Depth Anything: Unleashing the Power of Large-Scale Unlabeled Data}, 
  author={Yang, Lihe and Kang, Bingyi and Huang, Zilong and Xu, Xiaogang and Feng, Jiashi and Zhao, Hengshuang},
  booktitle=CVPR,
  year={2024}
}

@article{shi2024yell,
  title={Yell At Your Robot: Improving On-the-Fly from Language Corrections},
  author={Shi, Lucy and Hu, Zheyuan and Zhao, Tony and Sharma, Archit and Pertsch, Karl and Luo, Jianlan and Levine, Sergey and Finn, Chelsea},
  journal={Robotics: Science and Systems XX},
  year={2024},
  publisher={Robotics: Science and Systems Foundation}
}

@inproceedings{liuinteractive,
  title={Interactive Robot Learning from Verbal Correction},
  author={Liu, Huihan and Chen, Alice and Zhu, Yuke and Swaminathan, Adith and Kolobov, Andrey and Cheng, Ching-An},
  booktitle={2nd Workshop on Language and Robot Learning: Language as Grounding},
  year={2023}
}

@article{belkhale2024rt,
  title={Rt-h: Action hierarchies using language},
  author={Belkhale, Suneel and Ding, Tianli and Xiao, Ted and Sermanet, Pierre and Vuong, Quon and Tompson, Jonathan and Chebotar, Yevgen and Dwibedi, Debidatta and Sadigh, Dorsa},
  journal={arXiv preprint arXiv:2403.01823},
  year={2024}
}

@inproceedings{lihamster,
  title={HAMSTER: Hierarchical Action Models for Open-World Robot Manipulation},
  author={Li, Yi and Deng, Yuquan and Zhang, Jesse and Jang, Joel and Memmel, Marius and Garrett, Caelan Reed and Ramos, Fabio and Fox, Dieter and Li, Anqi and Gupta, Abhishek and others},
  booktitle={The Thirteenth International Conference on Learning Representations},
  year={2025}
}

@inproceedings{shi2025hi,
  title={Hi Robot: Open-Ended Instruction Following with Hierarchical Vision-Language-Action Models},
  author={Shi, Lucy Xiaoyang and Ichter, Brian and Equi, Michael Robert and Ke, Liyiming and Pertsch, Karl and Vuong, Quan and Tanner, James and Walling, Anna and Wang, Haohuan and Fusai, Niccolo and others},
  booktitle={International Conference on Machine Learning},
  pages={54919--54933},
  year={2025},
  organization={PMLR}
}

@inproceedings{wen2025dexvla,
  title={DexVLA: Vision-Language Model with Plug-In Diffusion Expert for General Robot Control},
  author={Wen, Junjie and Zhu, Yichen and Li, Jinming and Tang, Zhibin and Shen, Chaomin and Feng, Feifei},
  booktitle={Conference on Robot Learning},
  pages={3094--3114},
  year={2025},
  organization={PMLR}
}

@article{mao2024robomatrix,
  title={Robomatrix: A skill-centric hierarchical framework for scalable robot task planning and execution in open-world},
  author={Mao, Weixin and Zhong, Weiheng and Jiang, Zhou and Fang, Dong and Zhang, Zhongyue and Lan, Zihan and Li, Haosheng and Jia, Fan and Wang, Tiancai and Fan, Haoqiang and others},
  journal={arXiv preprint arXiv:2412.00171},
  year={2024}
}

@article{yang2025lohovla,
  title={Lohovla: A unified vision-language-action model for long-horizon embodied tasks},
  author={Yang, Yi and Sun, Jiaxuan and Kou, Siqi and Wang, Yihan and Deng, Zhijie},
  journal={arXiv preprint arXiv:2506.00411},
  year={2025}
}

@article{li2026elegantvla, title={ElegantVLA: Learning When to Think for Efficient Vision-Language-Action Models}, author={Li, Ye and Liu, Huanan and Ji, Kangye and Meng, Yuan and Fan, Jiajun and Wang, Yuansong and Qin, Shiyu and Wu, Chenglei and Xia, Shu-Tao and Wang, Zhi}, journal={arXiv preprint arXiv:2605.29438}, year={2026} }

@article{li2025sp, title={Sp-vla: A joint model scheduling and token pruning approach for vla model acceleration}, author={Li, Ye and Meng, Yuan and Sun, Zewen and Ji, Kangye and Tang, Chen and Fan, Jiajun and Ma, Xinzhu and Xia, Shutao and Wang, Zhi and Zhu, Wenwu}, journal={arXiv preprint arXiv:2506.12723}, year={2025} }

@article{li2025ts, title={TS-DP: Reinforcement Speculative Decoding For Temporal Adaptive Diffusion Policy Acceleration}, author={Li, Ye and Feng, Jiahe and Meng, Yuan and Ji, Kangye and Tang, Chen and Wen, Xinwan and Xia, Shutao and Wang, Zhi and Zhu, Wenwu}, journal={arXiv preprint arXiv:2512.15773}, year={2025} }

@article{li2025prance, title={Prance: Joint token-optimization and structural channel-pruning for adaptive vit inference}, author={Li, Ye and Tang, Chen and Meng, Yuan and Fan, Jiajun and Chai, Zenghao and Ma, Xinzhu and Wang, Zhi and Zhu, Wenwu}, journal={IEEE Transactions on Pattern Analysis and Machine Intelligence}, year={2025}, publisher={IEEE} }

@article{wei2022chain,
  title={Chain-of-thought prompting elicits reasoning in large language models},
  author={Wei, Jason and Wang, Xuezhi and Schuurmans, Dale and Bosma, Maarten and Xia, Fei and Chi, Ed and Le, Quoc V and Zhou, Denny and others},
  journal={Advances in neural information processing systems},
  volume={35},
  pages={24824--24837},
  year={2022}
}

@article{mu2023embodiedgpt,
  title={Embodiedgpt: Vision-language pre-training via embodied chain of thought},
  author={Mu, Yao and Zhang, Qinglong and Hu, Mengkang and Wang, Wenhai and Ding, Mingyu and Jin, Jun and Wang, Bin and Dai, Jifeng and Qiao, Yu and Luo, Ping},
  journal={Advances in Neural Information Processing Systems},
  volume={36},
  pages={25081--25094},
  year={2023}
}

@inproceedings{zawalski2025robotic,
  title={Robotic Control via Embodied Chain-of-Thought Reasoning},
  author={Zawalski, Micha{\l} and Chen, William and Pertsch, Karl and Mees, Oier and Finn, Chelsea and Levine, Sergey},
  booktitle={Conference on Robot Learning},
  pages={3157--3181},
  year={2025},
  organization={PMLR}
}

@inproceedings{zhang2025learning,
  title={Learning manipulation skills through robot chain-of-thought with sparse failure guidance},
  author={Zhang, Kaifeng and Yin, Zhao-Heng and Ye, Weirui and Gao, Yang},
  booktitle={2025 IEEE/RSJ International Conference on Intelligent Robots and Systems (IROS)},
  pages={8012--8018},
  year={2025},
  organization={IEEE}
}

@inproceedings{zhao2025cot,
  title={Cot-vla: Visual chain-of-thought reasoning for vision-language-action models},
  author={Zhao, Qingqing and Lu, Yao and Kim, Moo Jin and Fu, Zipeng and Zhang, Zhuoyang and Wu, Yecheng and Li, Zhaoshuo and Ma, Qianli and Han, Song and Finn, Chelsea and others},
  booktitle={Proceedings of the Computer Vision and Pattern Recognition Conference},
  pages={1702--1713},
  year={2025}
}

@article{cheang2025gr,
  title={Gr-3 technical report},
  author={Cheang, Chilam and Chen, Sijin and Cui, Zhongren and Hu, Yingdong and Huang, Liqun and Kong, Tao and Li, Hang and Li, Yifeng and Liu, Yuxiao and Ma, Xiao and others},
  journal={arXiv preprint arXiv:2507.15493},
  year={2025}
}

@inproceedings{kapelyukh2024dream2real,
  title={{Dream2Real}: Zero-shot {3D} object rearrangement with vision-language models},
  author={Kapelyukh, Ivan and Ren, Yifei and Alzugaray, Ignacio and Johns, Edward},
  booktitle={2024 IEEE International Conference on Robotics and Automation (ICRA)},
  pages={4796--4803},
  year={2024},
  organization={IEEE}
}

@inproceedings{radford2021learning,
  title={Learning transferable visual models from natural language supervision},
  author={Radford, Alec and Kim, Jong Wook and Hallacy, Chris and Ramesh, Aditya and Goh, Gabriel and Agarwal, Sandhini and Sastry, Girish and Askell, Amanda and Mishkin, Pamela and Clark, Jack and others},
  booktitle={International conference on machine learning},
  pages={8748--8763},
  year={2021},
  organization={PmLR}
}

@article{dosovitskiy2020image,
  title={An image is worth 16x16 words: Transformers for image recognition at scale},
  author={Dosovitskiy, Alexey and Beyer, Lucas and Kolesnikov, Alexander and Weissenborn, Dirk and Zhai, Xiaohua and Unterthiner, Thomas and Dehghani, Mostafa and Minderer, Matthias and Heigold, Georg and Gelly, Sylvain and others},
  journal={arXiv preprint arXiv:2010.11929},
  year={2020}
}

@inproceedings{armeni20193d,
  title={3d scene graph: A structure for unified semantics, 3d space, and camera},
  author={Armeni, Iro and He, Zhi-Yang and Gwak, JunYoung and Zamir, Amir R and Fischer, Martin and Malik, Jitendra and Savarese, Silvio},
  booktitle={Proceedings of the IEEE/CVF international conference on computer vision},
  pages={5664--5673},
  year={2019}
}

@inproceedings{wald2020learning,
  title={Learning 3d semantic scene graphs from 3d indoor reconstructions},
  author={Wald, Johanna and Dhamo, Helisa and Navab, Nassir and Tombari, Federico},
  booktitle={Proceedings of the IEEE/CVF Conference on Computer Vision and Pattern Recognition},
  pages={3961--3970},
  year={2020}
}

@article{yang2023set,
  title={Set-of-mark prompting unleashes extraordinary visual grounding in gpt-4v},
  author={Yang, Jianwei and Zhang, Hao and Li, Feng and Zou, Xueyan and Li, Chunyuan and Gao, Jianfeng},
  journal={arXiv preprint arXiv:2310.11441},
  year={2023}
}

@article{james2020rlbench,
  title={Rlbench: The robot learning benchmark \& learning environment},
  author={James, Stephen and Ma, Zicong and Arrojo, David Rovick and Davison, Andrew J},
  journal={IEEE Robotics and Automation Letters},
  volume={5},
  number={2},
  pages={3019--3026},
  year={2020},
  publisher={IEEE}
}

@article{brohan2022rt,
  title={Rt-1: Robotics transformer for real-world control at scale},
  author={Brohan, Anthony and Brown, Noah and Carbajal, Justice and Chebotar, Yevgen and Dabis, Joseph and Finn, Chelsea and Gopalakrishnan, Keerthana and Hausman, Karol and Herzog, Alex and Hsu, Jasmine and others},
  journal={arXiv preprint arXiv:2212.06817},
  year={2022}
}

@article{team2024octo,
  title={Octo: An open-source generalist robot policy},
  author={Team, Octo Model and Ghosh, Dibya and Walke, Homer and Pertsch, Karl and Black, Kevin and Mees, Oier and Dasari, Sudeep and Hejna, Joey and Kreiman, Tobias and Xu, Charles and others},
  journal={arXiv preprint arXiv:2405.12213},
  year={2024}
}

@article{qu2025spatialvla,
  title={Spatialvla: Exploring spatial representations for visual-language-action model},
  author={Qu, Delin and Song, Haoming and Chen, Qizhi and Yao, Yuanqi and Ye, Xinyi and Ding, Yan and Wang, Zhigang and Gu, JiaYuan and Zhao, Bin and Wang, Dong and others},
  journal={arXiv preprint arXiv:2501.15830},
  year={2025}
}

@article{yang2025qwen3,
  title={Qwen3 technical report},
  author={Yang, An and Li, Anfeng and Yang, Baosong and Zhang, Beichen and Hui, Binyuan and Zheng, Bo and Yu, Bowen and Gao, Chang and Huang, Chengen and Lv, Chenxu and others},
  journal={arXiv preprint arXiv:2505.09388},
  year={2025}
}

@inproceedings{ding2024open6dor,
  title={Open6DOR: Benchmarking open-instruction 6-DoF object rearrangement and a VLM-based approach},
  author={Ding, Yufei and Geng, Haoran and Xu, Chaoyi and Fang, Xiaomeng and Zhang, Jiazhao and Wei, Songlin and Dai, Qiyu and Zhang, Zhizheng and Wang, He},
  booktitle={2024 IEEE/RSJ International Conference on Intelligent Robots and Systems (IROS)},
  pages={7359--7366},
  year={2024},
  organization={IEEE}
}

@article{hurst2024gpt,
  title={Gpt-4o system card},
  author={Hurst, Aaron and Lerer, Adam and Goucher, Adam P and Perelman, Adam and Ramesh, Aditya and Clark, Aidan and Ostrow, AJ and Welihinda, Akila and Hayes, Alan and Radford, Alec and others},
  journal={arXiv preprint arXiv:2410.21276},
  year={2024}
}

@article{li2026robovlm,
  title={What matters in building vision--language--action models for generalist robots},
  author={Li, Xinghang and Li, Peiyan and Qian, Long and Liu, Minghuan and Wang, Dong and Liu, Jirong and Kang, Bingyi and Ma, Xiao and Wang, Xinlong and Guo, Di and others},
  journal={Nature Machine Intelligence},
  volume={8},
  pages={158--172},
  year={2026},
  doi={10.1038/s42256-025-01168-7},
  publisher={Nature Publishing Group UK London}
}

@article{carion2025sam,
  title={Sam 3: Segment anything with concepts},
  author={Carion, Nicolas and Gustafson, Laura and Hu, Yuan-Ting and Debnath, Shoubhik and Hu, Ronghang and Suris, Didac and Ryali, Chaitanya and Alwala, Kalyan Vasudev and Khedr, Haitham and Huang, Andrew and others},
  journal={arXiv preprint arXiv:2511.16719},
  year={2025}
}

@article{fang2025sam2act,
  title={SAM2Act: Integrating Visual Foundation Model with A Memory Architecture for Robotic Manipulation},
  author={Fang, Haoquan and Grotz, Markus and Pumacay, Wilbert and Wang, Yi Ru and Fox, Dieter and Krishna, Ranjay and Duan, Jiafei},
  journal={arXiv preprint arXiv:2501.18564},
  year={2025}
}

@inproceedings{shi2026memoryvla,
  title={MemoryVLA: Perceptual-Cognitive Memory in Vision-Language-Action Models for Robotic Manipulation},
  author={Shi, Hao and Xie, Bin and Liu, Yingfei and Sun, Lin and Liu, Fengrong and Wang, Tiancai and Zhou, Erjin and Fan, Haoqiang and Zhang, Xiangyu and Huang, Gao},
  booktitle={International Conference on Learning Representations},
  year={2026},
  url={https://openreview.net/forum?id=54U3XHf7qq}
}

@inproceedings{huang2026thinking,
  title={Thinking in Dynamics: How Multimodal Large Language Models Perceive, Track, and Reason Dynamics in Physical 4D World},
  author={Huang, Yuzhi and Wen, Kairun and Gao, Rongxin and Liu, Dongxuan and Lou, Yibin and Wu, Jie and Xu, Jing and Zhang, Jian and Yang, Zheng and Lin, Yunlong and Li, Chenxin and Pan, Panwang and Lu, Junbin and Jiang, Jingyan and Ding, Xinghao and Huang, Yue and Wang, Zhi},
  booktitle={Proceedings of the IEEE/CVF Conference on Computer Vision and Pattern Recognition},
  pages={33446--33456},
  year={2026}
}

@inproceedings{wen2025dynamicverse,
  title={DynamicVerse: A Physically-Aware Multimodal Framework for 4D World Modeling},
  author={Wen, Kairun and Huang, Yuzhi and Chen, Runyu and Zheng, Hui and Lin, Yunlong and Pan, Panwang and Li, Chenxin and Cong, Wenyan and Zhang, Jian and Lu, Junbin and Lin, Chenguo and Wang, Dilin and Yan, Zhicheng and Xu, Hongyu and Theiss, Justin and Huang, Yue and Ding, Xinghao and Ranjan, Rakesh and Fan, Zhiwen},
  booktitle={Advances in Neural Information Processing Systems},
  volume={38},
  year={2025},
  url={https://proceedings.neurips.cc/paper_files/paper/2025/hash/9c20f16b05f5e5e70fa07e2a4364b80e-Abstract-Conference.html}
}
}

\clearpage
\appendix

\section*{Appendix}
\addcontentsline{toc}{section}{Appendix}
\section{Robot Setups}
\subsection{Simulation Robot Setups}
We run all simulation experiments in two benchmarks: RLBench~\cite{james2020rlbench} and SIMPLER~\cite{li2025evaluating}.  

\textbf{(a) RLBench setup.} We use the standard RLBench environment equipped with a Franka Emika Panda (7-DoF) arm and a Panda Gripper, utilizing a planning-based control stack. Our approach relies primarily on the front RGB-D camera as the main visual observation for planning. We report results on 15 short-horizon tasks and 8 long-horizon tasks.

\textbf{(b) SIMPLER setup.} We evaluate two embodiments in SIMPLER: Google Robot and WidowX + Bridge. For visual observations, we utilize the overhead camera for the Google Robot and the third-person-view camera for WidowX. Crucially, we apply our training-free planning pipeline directly across both configurations without any environment-specific fine-tuning.

\subsection{Real World Robot Setups}

\begin{figure}[htbp]
    \centering
    \newlength{\robotheight}\settoheight{\robotheight}{\includegraphics[width=0.72\textwidth]{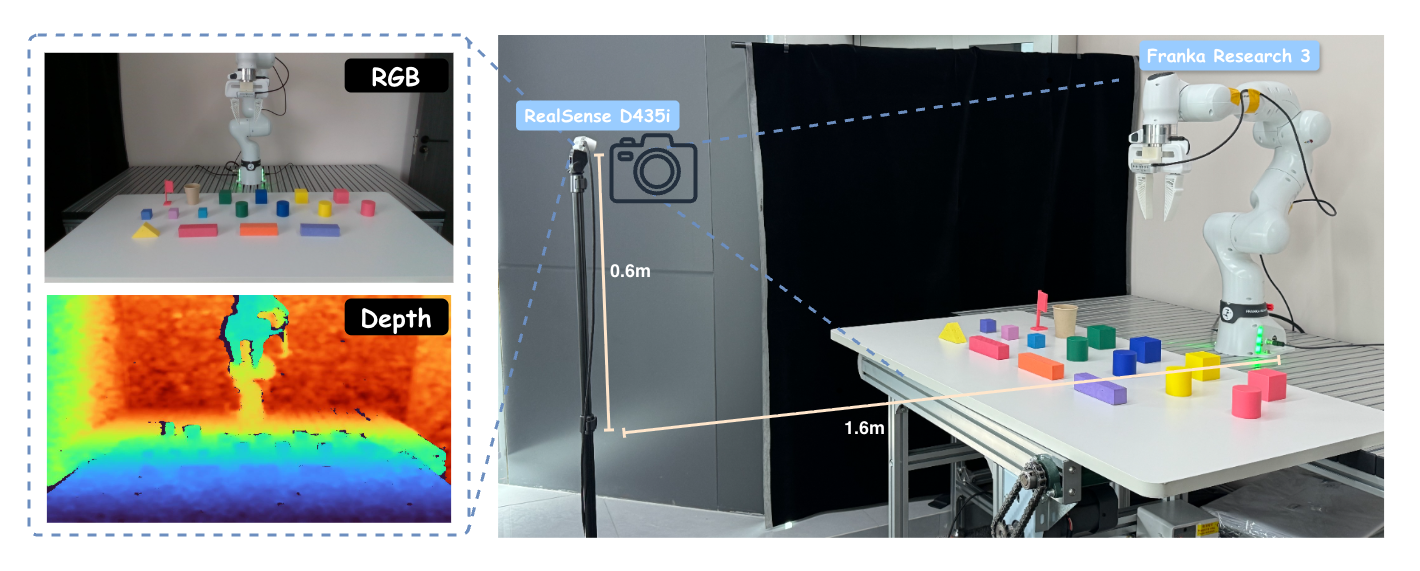}}%
    \begin{subfigure}[b]{0.25\textwidth}
        \centering
        \includegraphics[height=\robotheight]{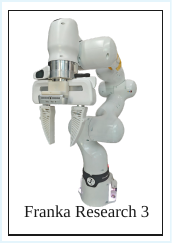}
        \caption{Franka Research 3.}
        \label{fig:appendix_franka}
    \end{subfigure}
    \hfill
    \begin{subfigure}[b]{0.72\textwidth}
        \centering
        \includegraphics[width=\textwidth]{Figures/appendix_robotenv.png}
        \caption{Workspace setup with RGB and depth views.}
        \label{fig:appendix_robotenv}
    \end{subfigure}
    \caption{\textbf{Real-world experimental setup.} (a) The Franka Research 3 robot arm used in our experiments. (b) The physical workspace equipped with the robot arm and an Intel RealSense D435i camera; the insets show the corresponding RGB view and depth map.}
    \label{fig:appendix_robot_setup}
\end{figure}

In our real-world experiments, we uniformly utilized a Franka Research 3 (FR3) robotic arm equipped with a parallel gripper. A front-view Intel RealSense D435i RGB-D camera was deployed to capture both visual and depth information for point cloud reconstruction. The detailed workspace and robot configurations are illustrated in Fig.~\ref{fig:appendix_robot_setup}.

\vspace{-2em}

\section{Experimental Assets and Task Design}
\begin{figure}[htbp]
    \centering
    \includegraphics[width=0.6\textwidth]{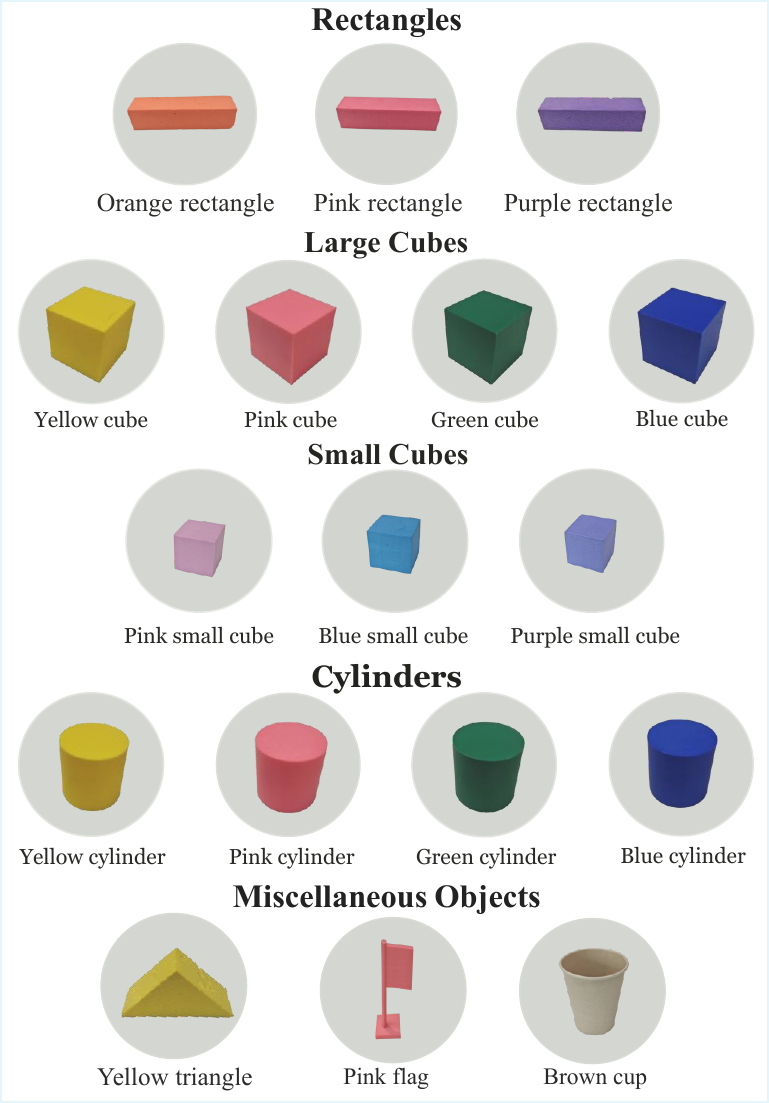}
    \caption{\textbf{Assets for real-world experiments.} A collection of 17 manipulation objects including geometric primitives, a triangular prism, and functional objects used across all task difficulty levels.}
    \label{fig:appendix_assets}
\end{figure}

\noindent\textbf{Real-World Experimental Assets.} 
As shown in Fig.~\ref{fig:appendix_assets}, our real-world experiments utilize a diverse set of manipulation objects designed to evaluate precise spatial reasoning and sequential stacking. These assets primarily consist of: 
(i) \textit{Geometric primitives} including rectangular blocks, cubes, and cylinders in various sizes and distinct colors (pink, blue, green, yellow, purple, and orange); 
(ii) \textit{A triangular prism} (yellow triangle in Fig.~\ref{fig:appendix_assets}) used for assessing high-precision alignment and challenging placement tasks; and 
(iii) \textit{Functional objects} (e.g., a pink flag and a brown cup) to create dynamic occlusion and nesting scenarios. 
In total, these 17 unique pieces allow for a vast combinatorial space of long-horizon tasks, such as multi-layer block building and disassembling, providing a rigorous testbed for the model's object permanence and spatial consistency.

\vspace{-2em}

\subsection{Design of Real-World Tasks}

To rigorously validate the core contributions of RoboStream, specifically the geometry-grounded perception of STF-Tokens and the causal state tracking of the CSTG, we engineered a hierarchical suite of 21 real-world tasks. Unlike random pick-and-place scenarios, these tasks are hypothesis-driven stress tests designed to probe the boundaries of spatio-temporal reasoning. The evaluation is structured into three task categories (corresponding to Task~A, B, and C in the main text), organized along two primary evaluation axes. (i)~\textit{Spatio-temporal grounding and sequential reasoning:} Tasks~A (Block Building) and~B (Block Disassembly) assess whether STF-Tokens and the CSTG sustain accurate spatial reasoning across progressively longer construction and deconstruction sequences, with target configurations specified by goal images. (ii)~\textit{Causal memory and object permanence:} Task~C (Block Hide and Restore) stress-tests whether the CSTG preserves latent object states when the corresponding objects are fully occluded, guided by natural language instructions. Each task configuration is executed three times.

\begin{figure}[htbp]
    \centering
    \includegraphics[width=1\textwidth]{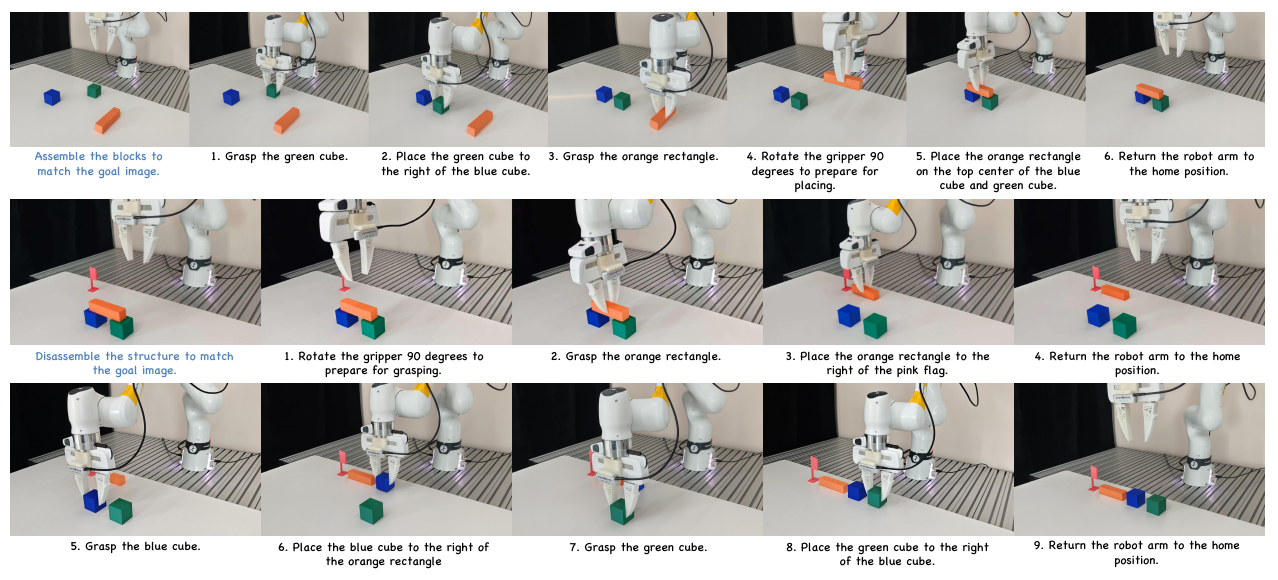} 
    \caption{\textbf{Long-horizon block manipulation (Easy).} Execution of basic assembly and disassembly involving 2--3 objects, demonstrating fundamental spatial grounding and pick-and-place precision.}
    \label{fig:appendix_step1}
\end{figure}

\begin{figure}[htbp]
    \centering
    \includegraphics[width=1\textwidth]{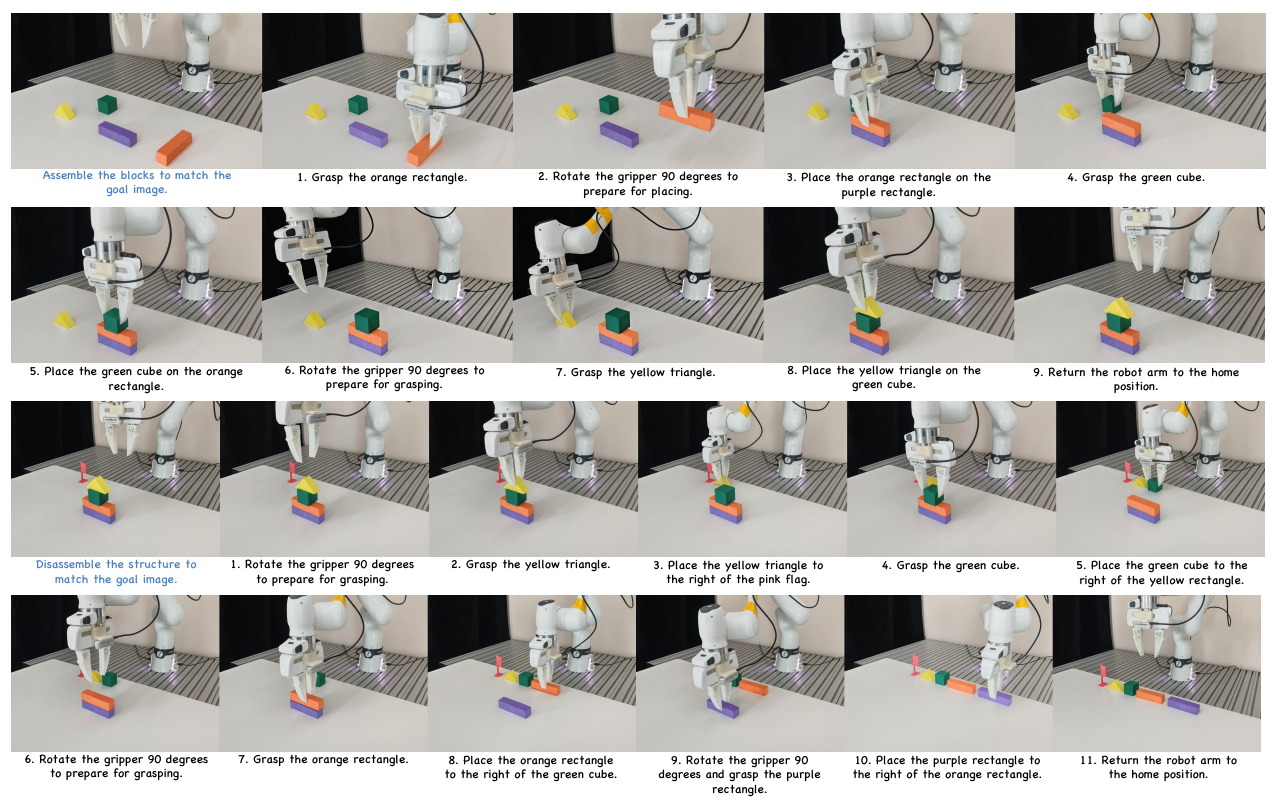} 
    \caption{\textbf{Long-horizon block manipulation (Medium).} Intermediate execution requiring a strict sequence of 3--4 steps to manipulate 4 objects, verifying sequential planning and state tracking.}
    \label{fig:appendix_step2}
\end{figure}

\begin{figure}[htbp]
    \centering
    \includegraphics[width=1\textwidth]{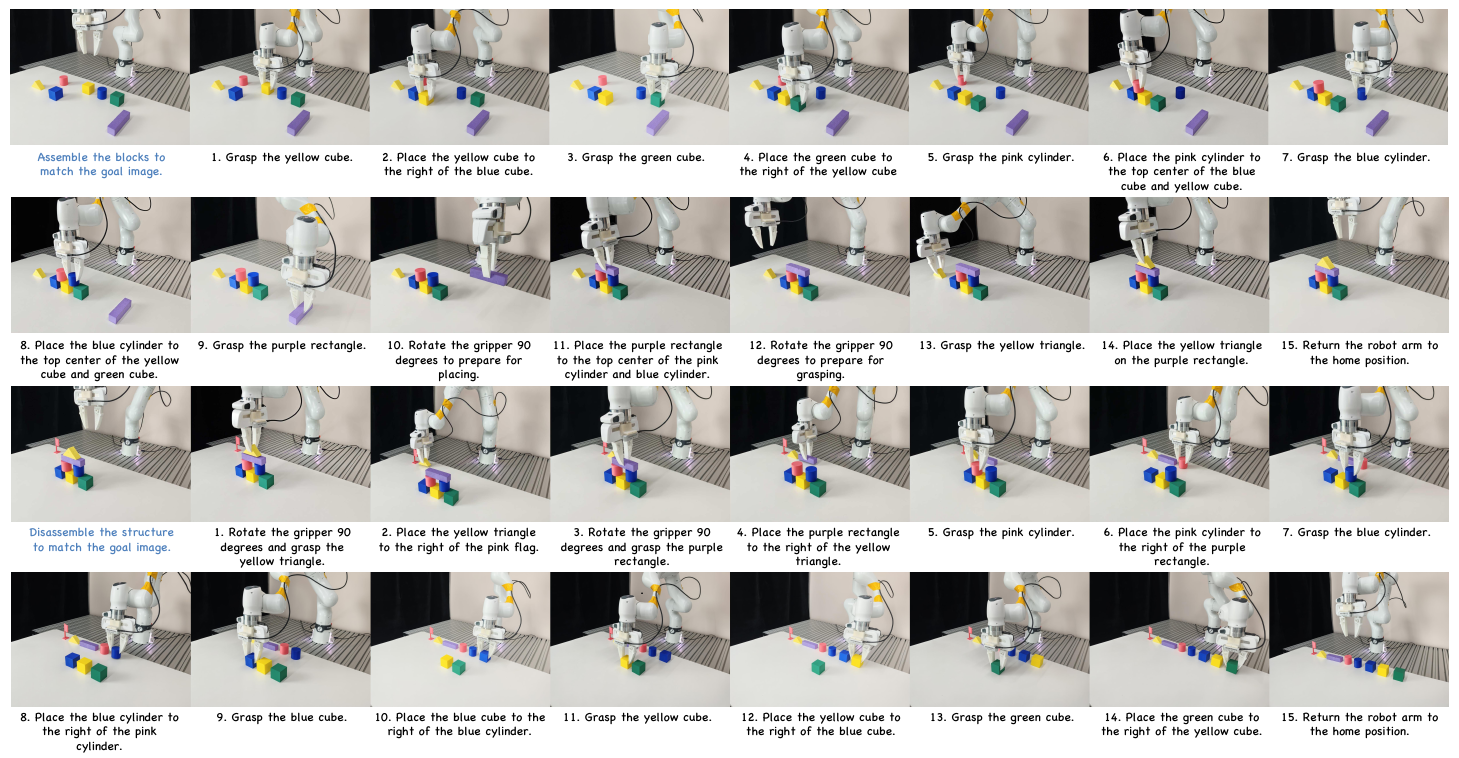} 
    \caption{\textbf{Long-horizon block manipulation (Hard).} Extreme long-horizon execution involving 7 objects and 6--7 steps. The sequence highlights the model's ability to maintain physical stability during complex ``bottom-to-top'' building and ``top-to-bottom'' disassembly without collapsing the structure.}
    \label{fig:appendix_step3}
\end{figure}

\noindent\textbf{Block Building and Disassembly (Fig.~\ref{fig:appendix_step1}, \ref{fig:appendix_step2}, \ref{fig:appendix_step3}).} 
These two categories jointly evaluate spatio-temporal grounding and sequential state tracking through symmetric construction and deconstruction sequences: bottom-to-top stacking (Block Building) and top-to-bottom removal (Block Disassembly), each across three synchronized difficulty levels. At the \textit{Easy and Medium} levels, distractors sharing colors but differing in shape are deliberately introduced to test whether the model resists visual-semantic grounding errors; at each step the robot must verify base stability and placement preconditions before proceeding.

The \textit{Hard} level scales to 7 objects and 6--7 sequential steps, imposing strict physical constraints from opposite directions. During building, the VLM must continuously update its scene representation via the CSTG, treating newly placed objects as valid geometric anchors for subsequent layers to suppress cascading spatial hallucinations. During disassembly, the model must instead deduce the correct top-to-bottom dependency chain from the CSTG; a single greedy mistake (such as extracting a bottom support block) causes immediate structural collapse, proving the necessity of explicit 3D structural reasoning and accurate causal memory of stacking order and support dependencies, which simple semantic matching cannot provide.

\begin{figure}[htbp]
    \centering
    \includegraphics[width=1\textwidth]{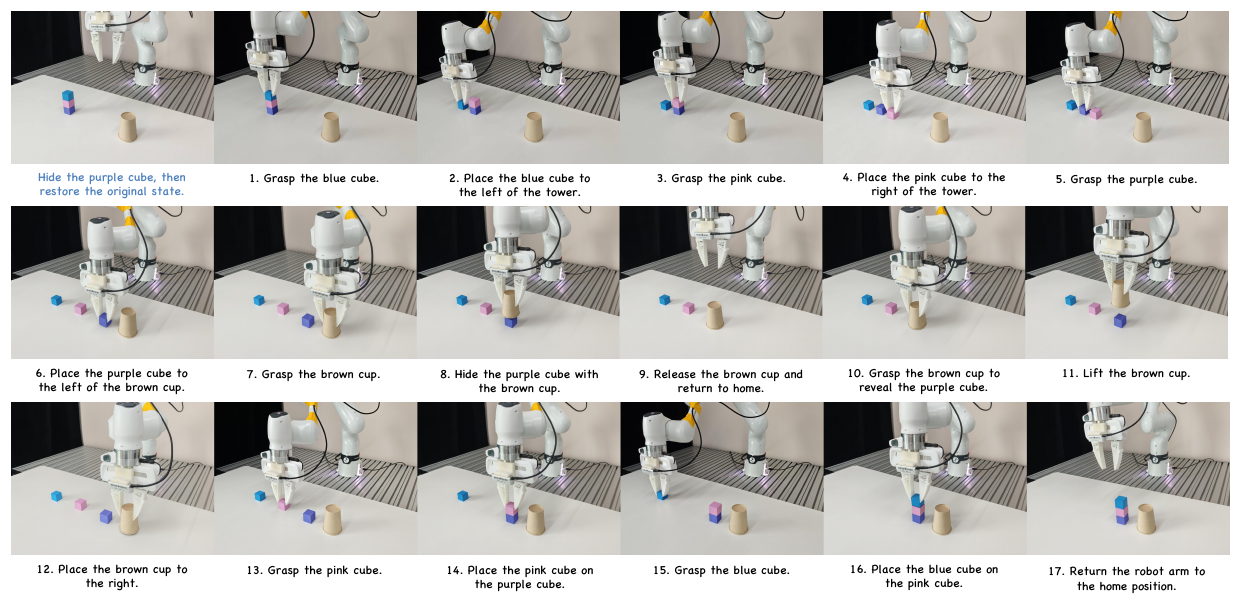} 
    \caption{\textbf{Long-horizon Block Hide and Restore task (Occlusion).} A stress-test for causal memory and object permanence. The robot hides the purple cube beneath a brown cup, executes intermediate distractor steps, and successfully retrieves the hidden object to restore the original configuration.}
    \label{fig:appendix_step4}
\end{figure}

\noindent\textbf{Block Hide and Restore (Fig.~\ref{fig:appendix_step4}).} 
To explicitly validate the necessity of the Causal Spatio-Temporal Graph (CSTG), we engineered Block Hide and Restore tasks that stress-test causal memory and object permanence under full occlusion. In the most challenging configuration, the robot is instructed to hide the \textit{bottom} block of an existing tower. This forces a deep causal dependency chain: the model must autonomously deduce the physical necessity to first disassemble the blocking upper layers, conceal the target under an opaque cup, execute intermediate distractor steps, and ultimately reverse the entire sequence to restore the exact original tower. 

This design creates an extreme perceptual bottleneck where direct visual evidence of the hidden object and its original stacking order is unavailable because the object is fully occluded. Successfully completing this task proves that RoboStream maintains a persistent causal spatio-temporal representation via the CSTG, relying on its causal memory log to track latent object identities and last-known poses when reactive visual perception fundamentally fails.

\vspace{2em}
\subsection{Design of Simulation Tasks}
To comprehensively evaluate the capabilities of RoboStream, we carefully select and design tasks across two simulation platforms: SIMPLER~\cite{li2025evaluating} and RLBench~\cite{james2020rlbench}. Rather than treating all tasks uniformly, we deliberately categorize them into two distinct evaluation axes to independently stress-test zero-shot generalization, high-precision spatial grounding, and long-horizon causal memory.

\begin{figure}[htpb]
    \centering
    \includegraphics[width=1\textwidth]{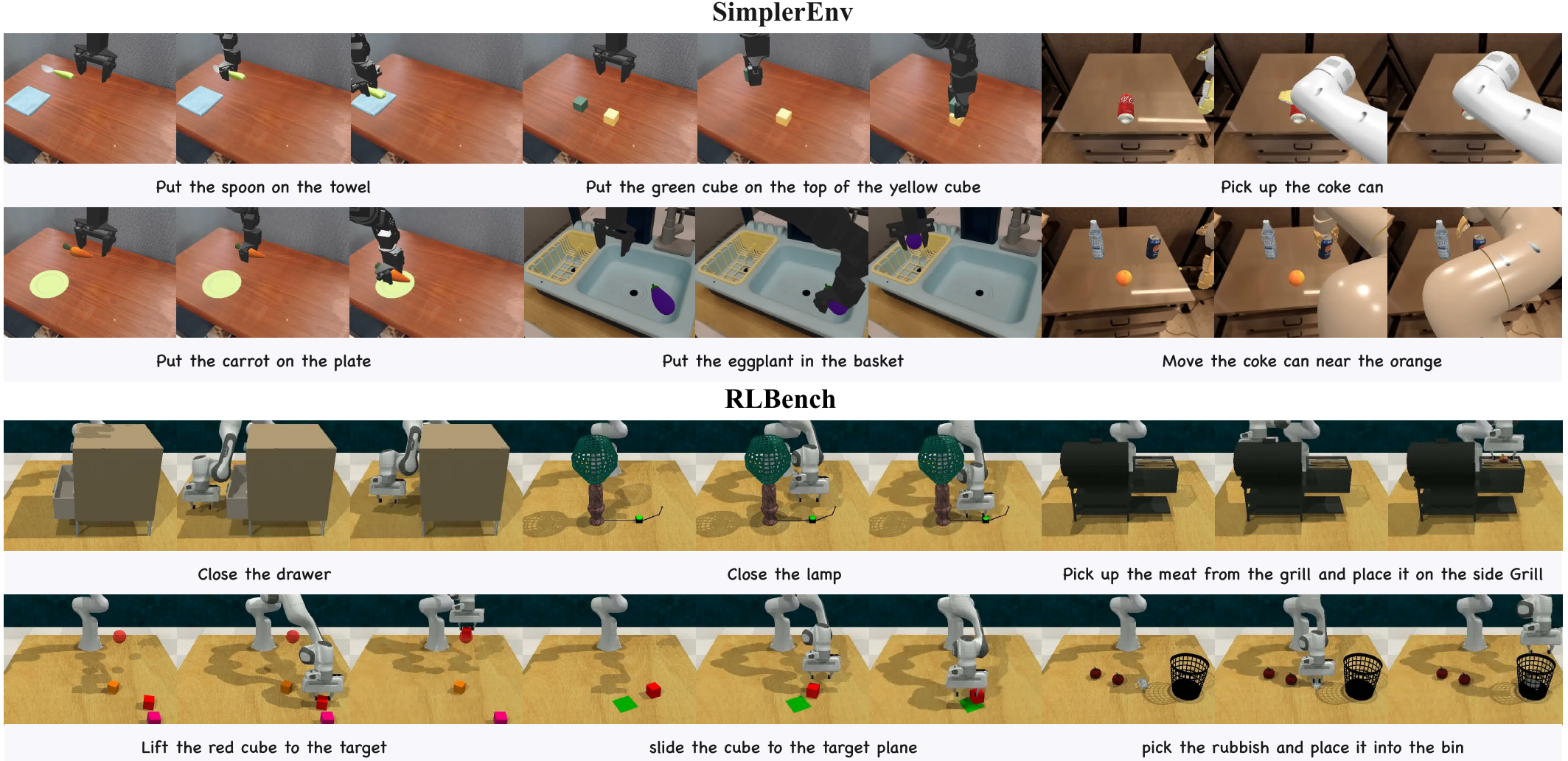} 
    \caption{\textbf{Visualization of short-horizon evaluation tasks.} Execution examples from the SIMPLER environment (top) and RLBench (bottom). These tasks are designed to demonstrate the model's zero-shot cross-embodiment generalization and its capability for high-precision spatial grounding in contact-rich scenarios.}
    \label{fig:appendix_sim1}
\end{figure}

\noindent\textbf{Evaluating Spatial Grounding and Zero-Shot Generalization.} 
The first suite of tasks is designed to isolate and evaluate the explicit 3D geometric grounding provided by STF-Tokens, independent of long-term memory requirements. To assess zero-shot cross-embodiment generalization, we deploy our framework in the SIMPLER benchmark (visualized in Fig.~\ref{fig:appendix_sim1}, top). We evaluate the Google Robot embodiment on tasks like ``Pick Coke Can'' and ``Move Near'', alongside the WidowX + Bridge embodiment on tasks such as ``Put Spoon on Towel'', ``Put Carrot on Plate'', ``Stack Green Block on Yellow Block'', and ``Put Eggplant in Basket''. The design rationale here is to verify that the spatio-temporal reasoning enabled by STF-Tokens is not overfitted to a specific robot kinematic structure or camera viewpoint. By deploying RoboStream on unseen embodiments without fine-tuning, we demonstrate its robustness to visual domain shifts and novel spatial layouts. 

Complementing this cross-embodiment evaluation, we select 15 diverse short-horizon tasks from the standard RLBench suite (e.g., \textit{Close Drawer, Lamp Off, Meat Off Grill, Put Rubbish in Bin, Slide Block to Target}, as shown in Fig.~\ref{fig:appendix_sim1}, bottom). The primary motivation for including these short-horizon tasks is their demand for fine-grained geometric understanding and contact-rich interactions. Testing on these specific scenarios allows us to prove that explicit 3D geometric representations—captured via centroids and Gaussian shapes—surpass pixel-level implicit reasoning even before complex sequential planning is required.

\begin{figure}[htbp]
    \centering
    \includegraphics[width=0.93\textwidth]{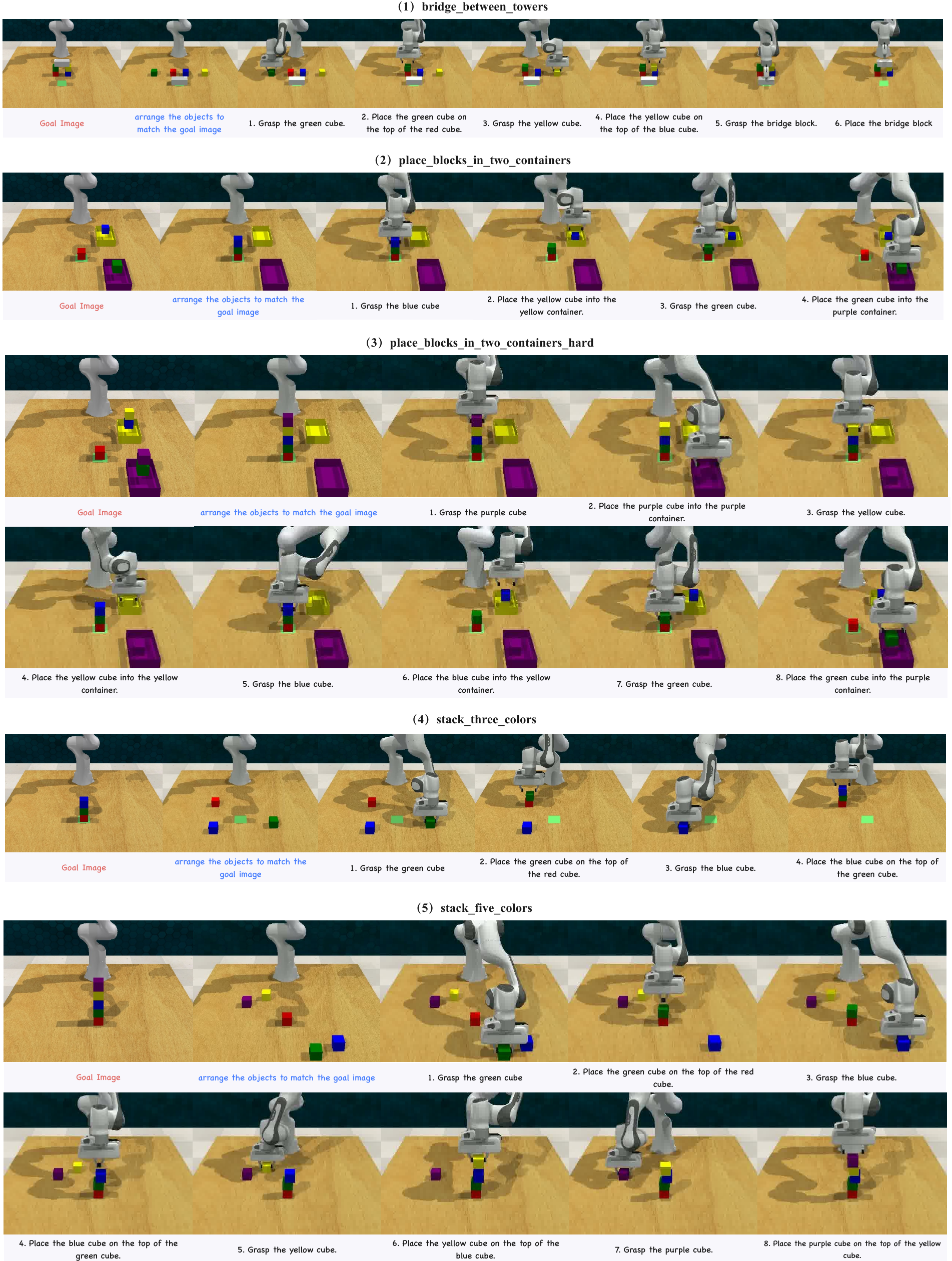} 
    \caption{\textbf{Goal-Image Guided long-horizon Tasks. } Execution sequences for five long-horizon tasks in RLBench where the target state is provided via a visual goal image. These tasks systematically evaluate the model's ability to maintain spatial coherence and track multiple objects across extended dynamic trajectories.}
    \label{fig:appendix_sim2}
\end{figure}

\noindent\textbf{Evaluating Causal Memory and Object Permanence.} 
To explicitly stress-test the Causal Spatio-Temporal Graph (CSTG) and its ability to maintain object permanence, we engineer a custom suite of 8 complex, long-horizon tasks in RLBench. These tasks are structured around two distinct guidance modalities to evaluate different aspects of persistent memory. The first category consists of goal-image guided tasks (\textit{Bridge Between Towers, Place Blocks in Two Containers (Normal/Hard), and Stack Three/Five Colors}, visualized in Fig.~\ref{fig:appendix_sim2}). In these scenarios, the robot must deduce the correct long-horizon execution sequence from a visually complex target state. This design tests the model's capacity to maintain spatial coherence and track multiple object states simultaneously as the scene dynamically evolves. 

\begin{figure}[htbp]
    \centering
    \includegraphics[width=0.91\textwidth]{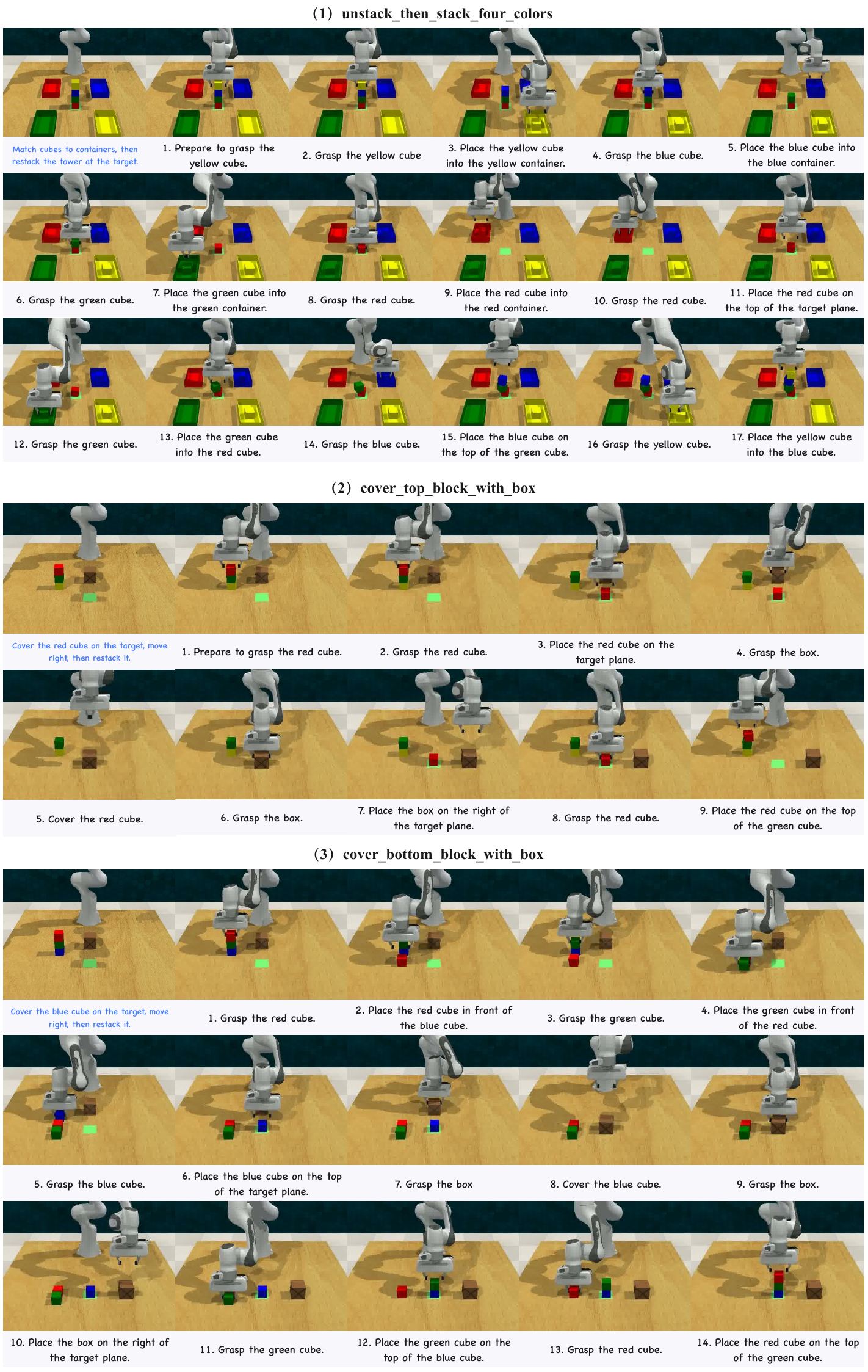} 
    \caption{\textbf{Text-Guided long-horizon Tasks.} Execution sequences in RLBench driven by natural language instructions. Tasks involving severe occlusion explicitly evaluate the CSTG's capacity for causal memory and object permanence.}
    \label{fig:appendix_sim3}
\end{figure}

The second category introduces text-guided tasks (\textit{Cover Top/Bottom Block with Box, and Unstack then Stack Four Colors}, visualized in Fig.~\ref{fig:appendix_sim3}), which are specifically designed to evaluate causal memory under severe occlusion. By instructing the robot to hide a target object and subsequently restore the environment, we force the system to rely entirely on its causal memory log rather than immediate visual perception. This task design acts as a rigorous stress test, providing strong evidence for the necessity of the CSTG in preventing spatial hallucinations when visual evidence is temporarily lost.

\section{Detailed Experimental Results}
\subsection{Detailed Results of Real-World Manipulation}

To validate the effectiveness of our framework in physical environments, we report detailed success rates across three categories of manipulation tasks in Tab.~\ref{tab:real_world_detailed} and Tab.~\ref{tab:real_world_hide}. Our evaluation spans 21 unique task configurations, ranging from simple 2-object assembly to complex 7-object long-horizon restoration under occlusion. 

\begin{table}[htbp]
    \centering
    \caption{\textbf{Detailed results of real-world block building and block disassembly tasks.}}
    \label{tab:real_world_detailed}
    \renewcommand{\arraystretch}{1.1} 
    \resizebox{\textwidth}{!}{
    \begin{tabular}{m{0.3\textwidth} c c c c c c}
        \toprule
        \multirow{2}{*}{\textbf{Task}} & \multirow{2}{*}{\textbf{Difficulty}} & \multicolumn{5}{c}{\textbf{Success Rate}} \\
        \cmidrule(lr){3-7}
        & & \textbf{SoFar} & \textbf{VoxPoser} & \textbf{RoboStream-8B} & \textbf{RoboStream-32B} & \textbf{RoboStream-235B} \\
        \midrule
        
        \multicolumn{7}{c}{\textit{Block Building (Bottom-to-Top)} {\scriptsize $\vert$ \textit{Task $=$ Goal Image}}} \\
        \midrule
        \begin{minipage}{0.3\textwidth}
            \centering \vspace{1pt}
            \includegraphics[height=1.0cm]{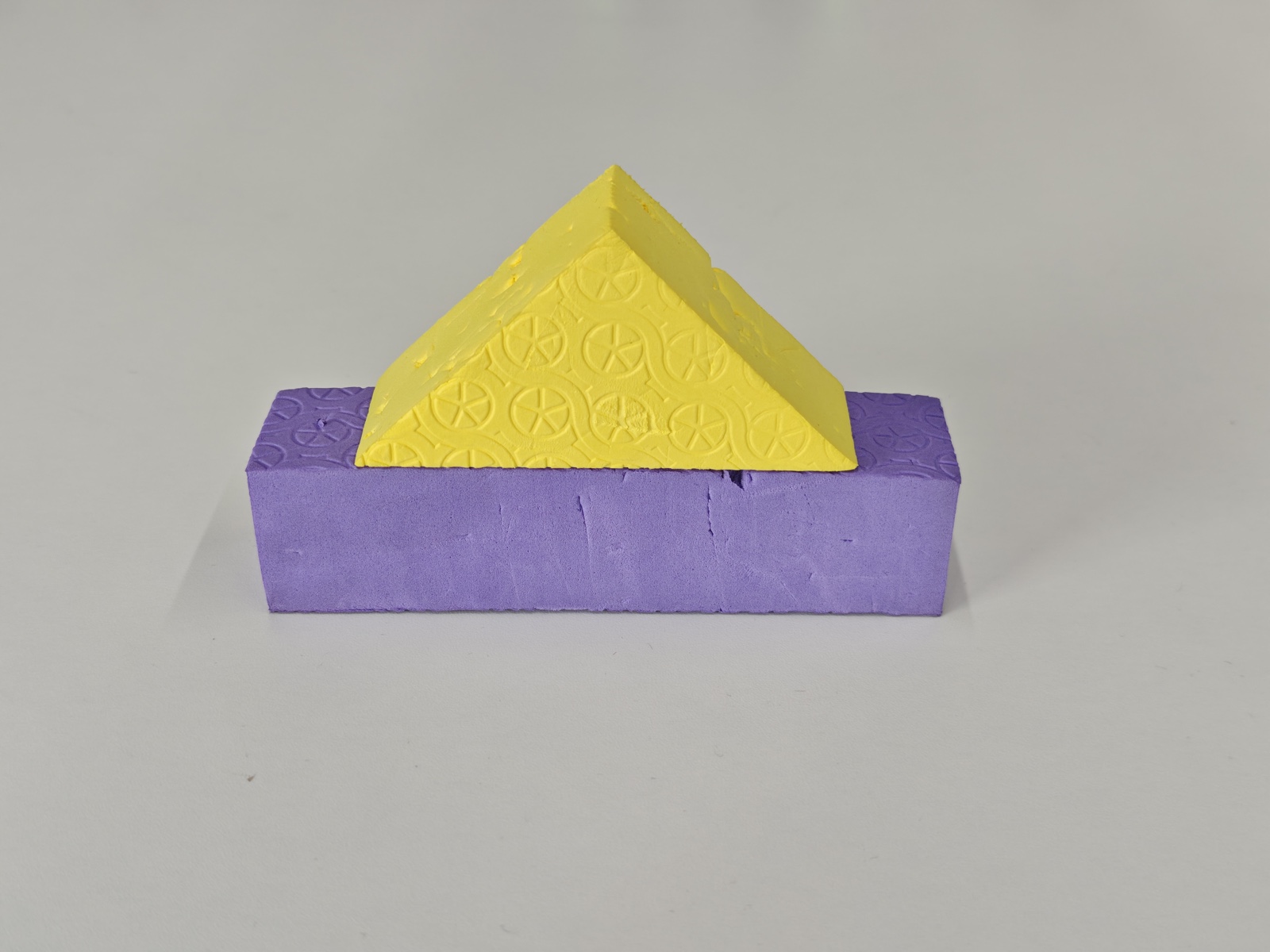}
            \vspace{1pt}
        \end{minipage} 
        & Easy & 3/3 & 2/3 & 1/3 & 2/3 & \textbf{2/3} \\
        
        \begin{minipage}{0.3\textwidth}
            \centering \vspace{1pt}
            \includegraphics[height=1.0cm]{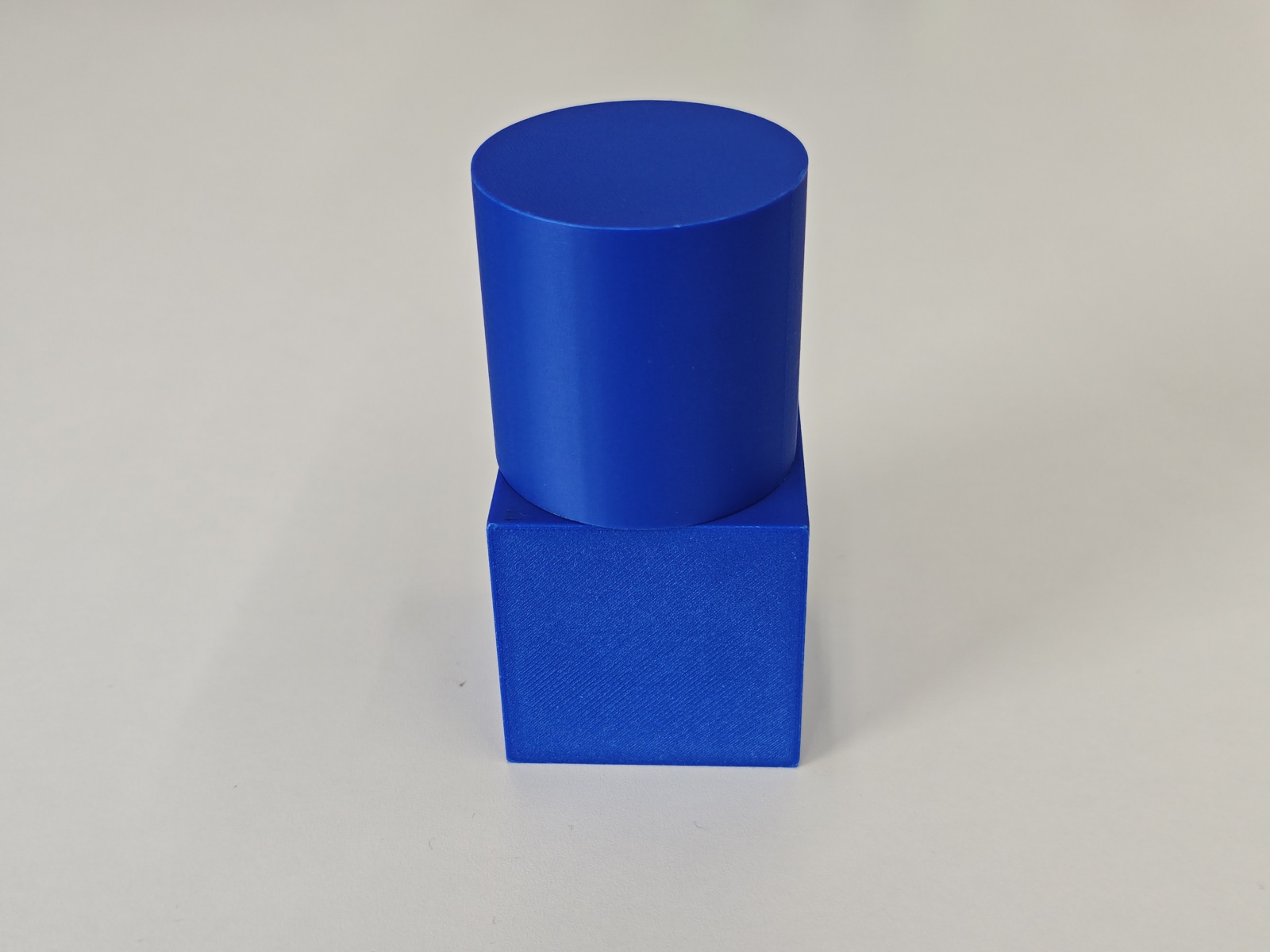}
            \vspace{1pt}
        \end{minipage} 
        & Easy & 2/3 & 2/3 & 3/3 & 1/3 & \textbf{3/3} \\
        
        \begin{minipage}{0.3\textwidth}
            \centering \vspace{1pt}
            \includegraphics[height=1.0cm]{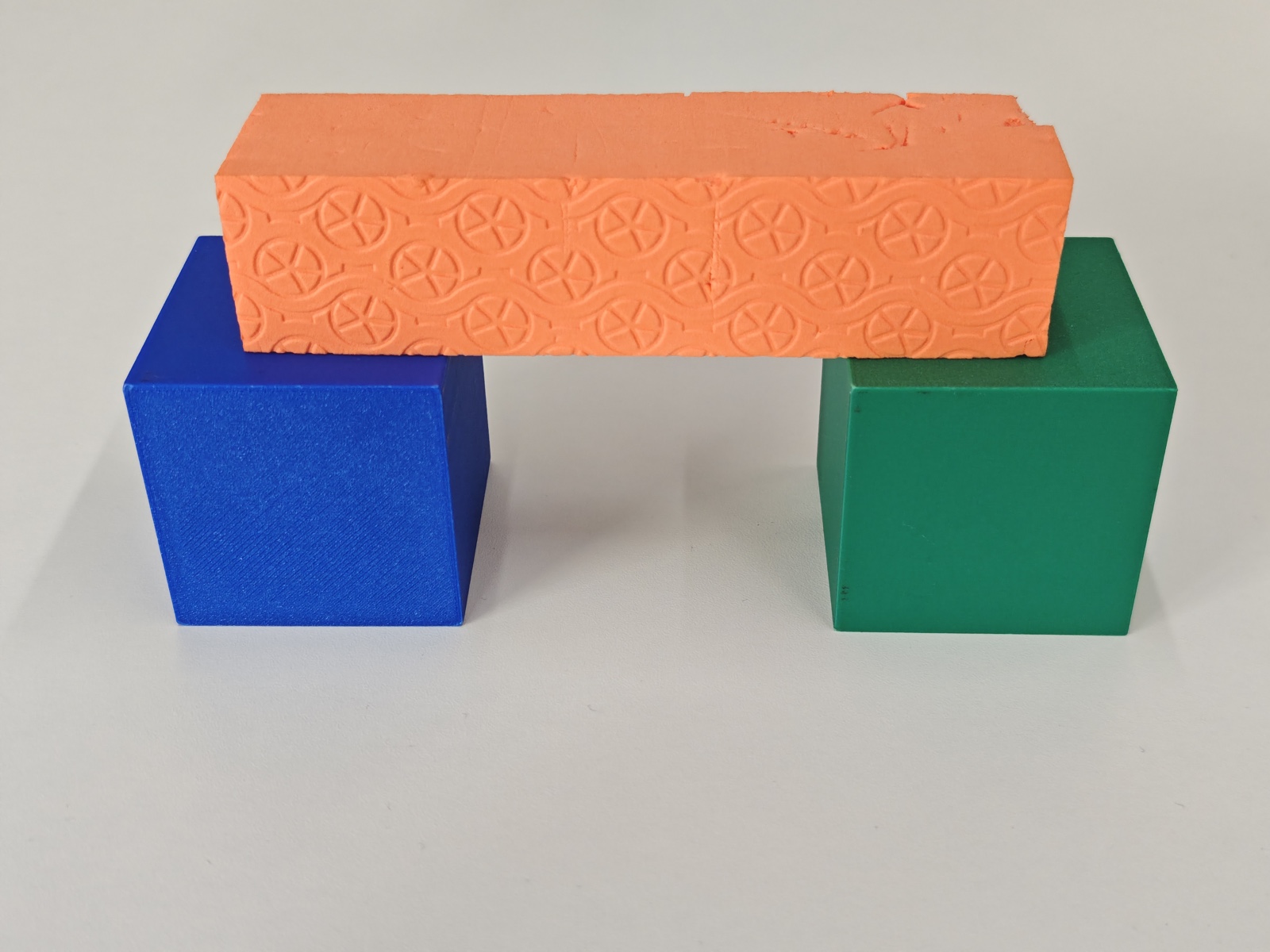}
            \vspace{1pt}
        \end{minipage} 
        & Easy & 0/3 & 0/3 & 1/3 & 3/3 & \textbf{2/3} \\
        
        \begin{minipage}{0.3\textwidth}
            \centering \vspace{1pt}
            \includegraphics[height=1.0cm]{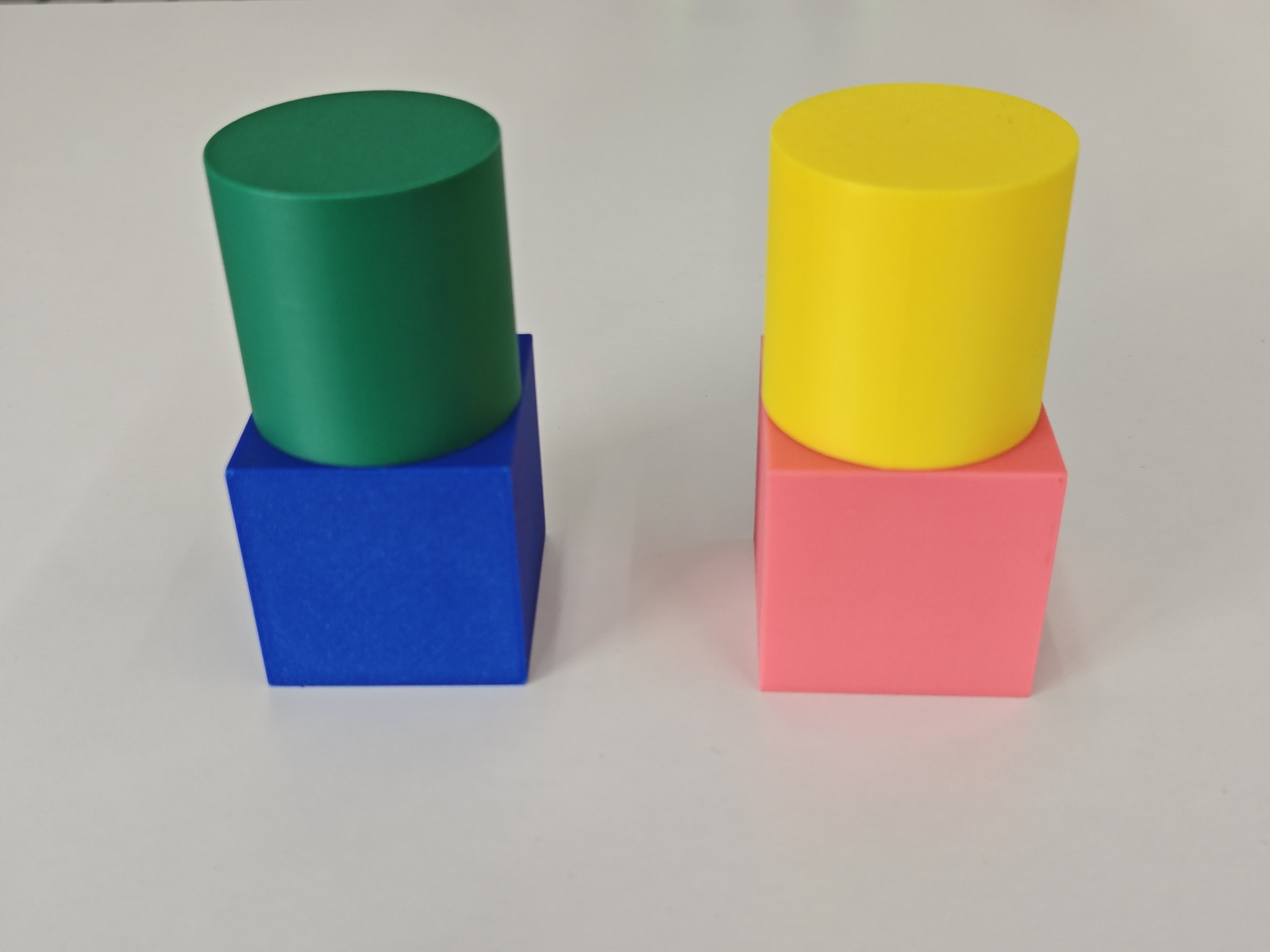}
            \vspace{1pt}
        \end{minipage} 
        & Medium & 0/3 & 2/3 & 2/3 & 2/3 & \textbf{3/3} \\
        
        \begin{minipage}{0.3\textwidth}
            \centering \vspace{1pt}
            \includegraphics[height=1.0cm]{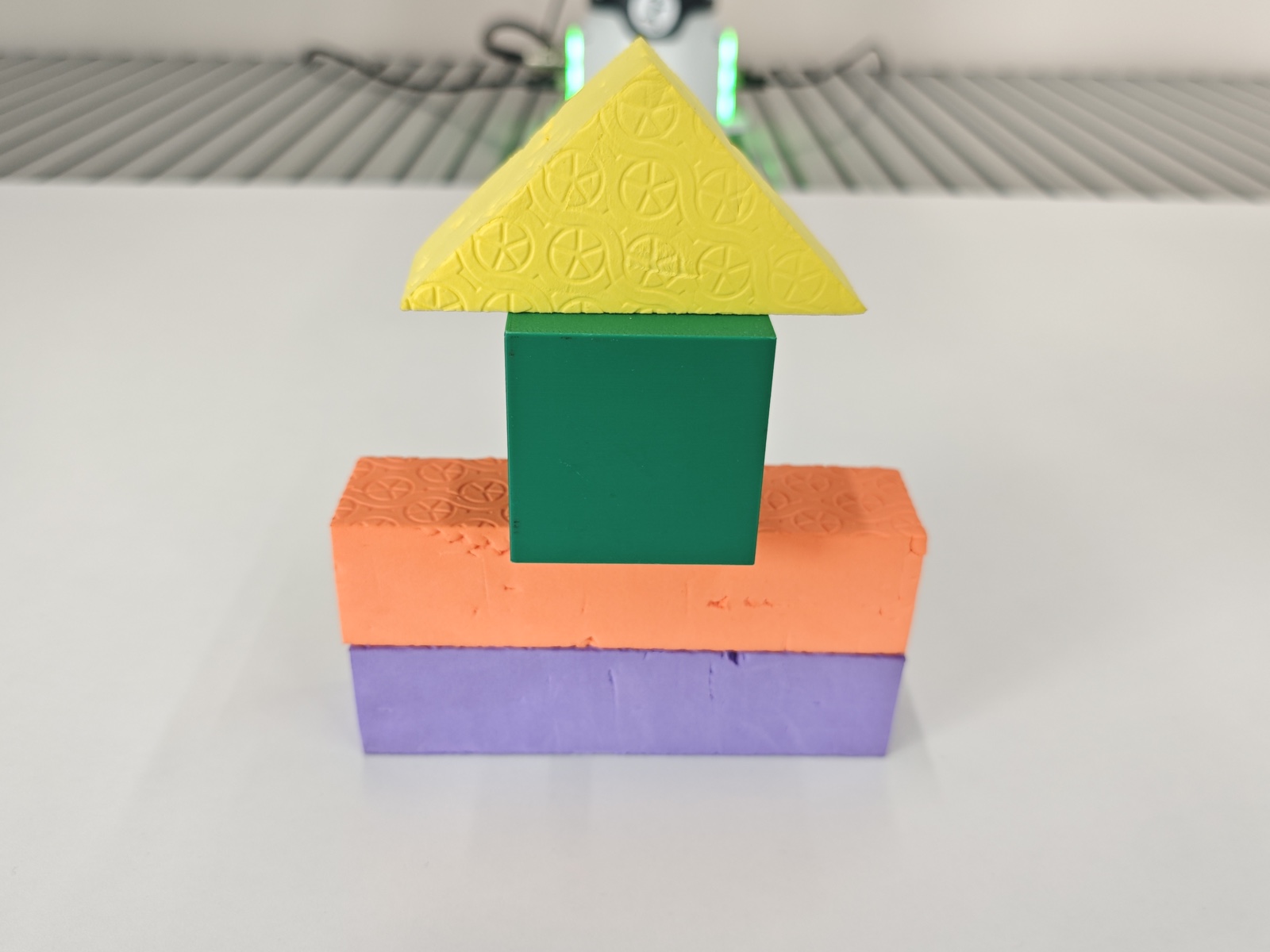}
            \vspace{1pt}
        \end{minipage} 
        & Medium & 0/3 & 0/3 & 0/3 & 1/3 & \textbf{1/3} \\
        
        \begin{minipage}{0.3\textwidth}
            \centering \vspace{1pt}
            \includegraphics[height=1.0cm]{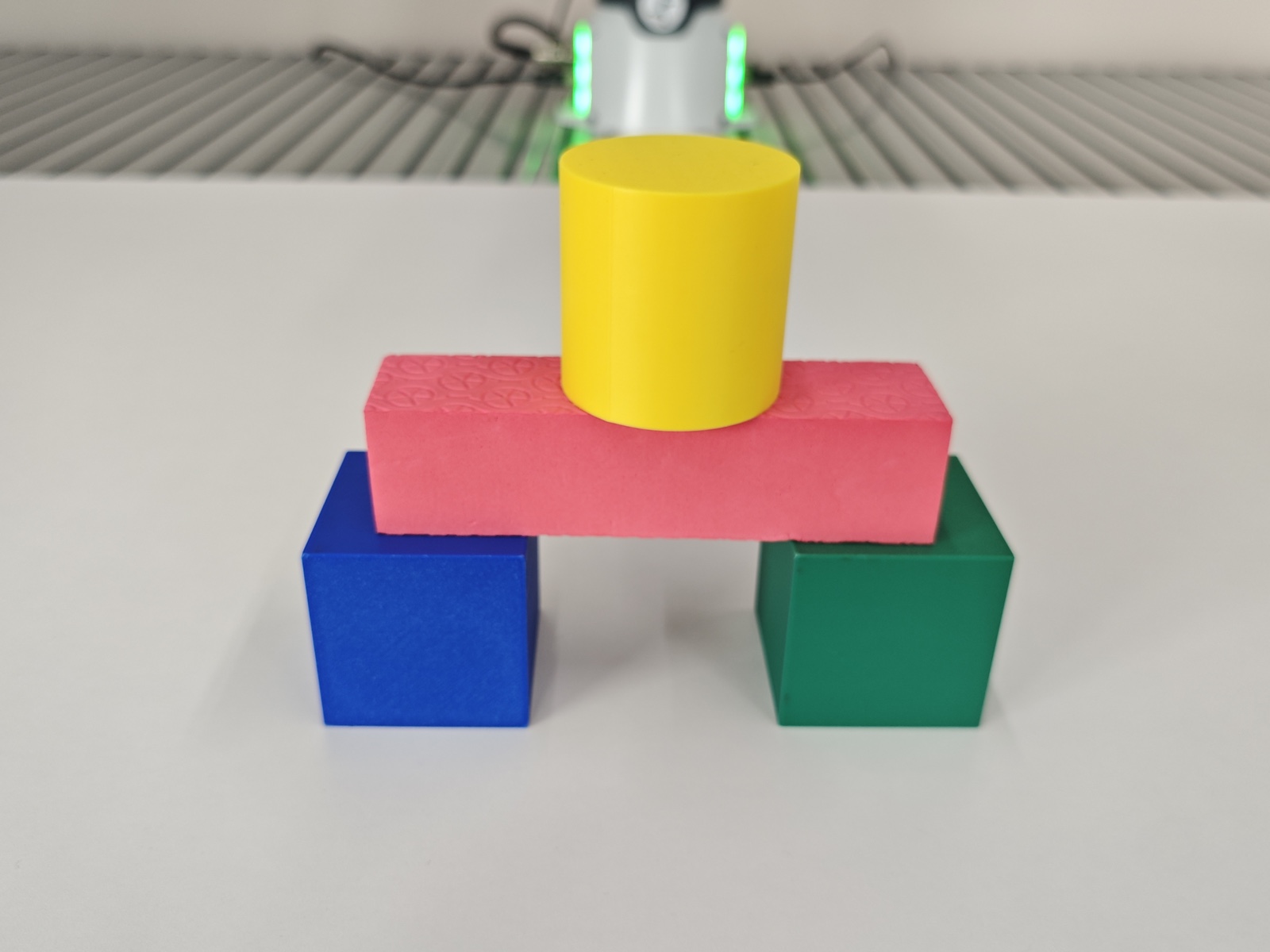}
            \vspace{1pt}
        \end{minipage} 
        & Medium & 0/3 & 0/3 & 0/3 & 1/3 & \textbf{2/3} \\
        
        \begin{minipage}{0.3\textwidth}
            \centering \vspace{1pt}
            \includegraphics[height=1.0cm]{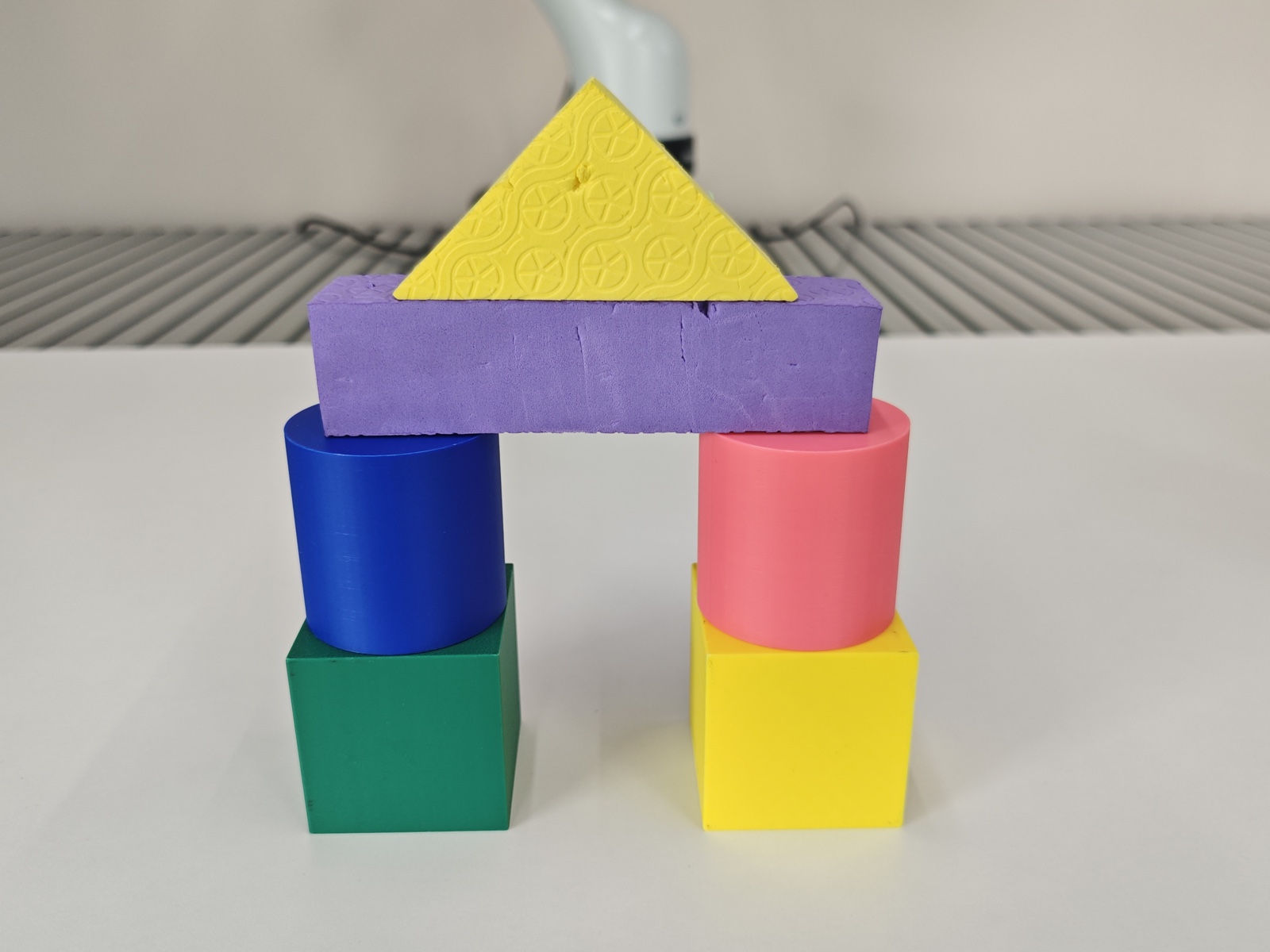}
            \vspace{1pt}
        \end{minipage} 
        & Hard & 1/3 & 1/3 & 2/3 & 2/3 & \textbf{2/3} \\
        
        \begin{minipage}{0.3\textwidth}
            \centering \vspace{1pt}
            \includegraphics[height=1.0cm]{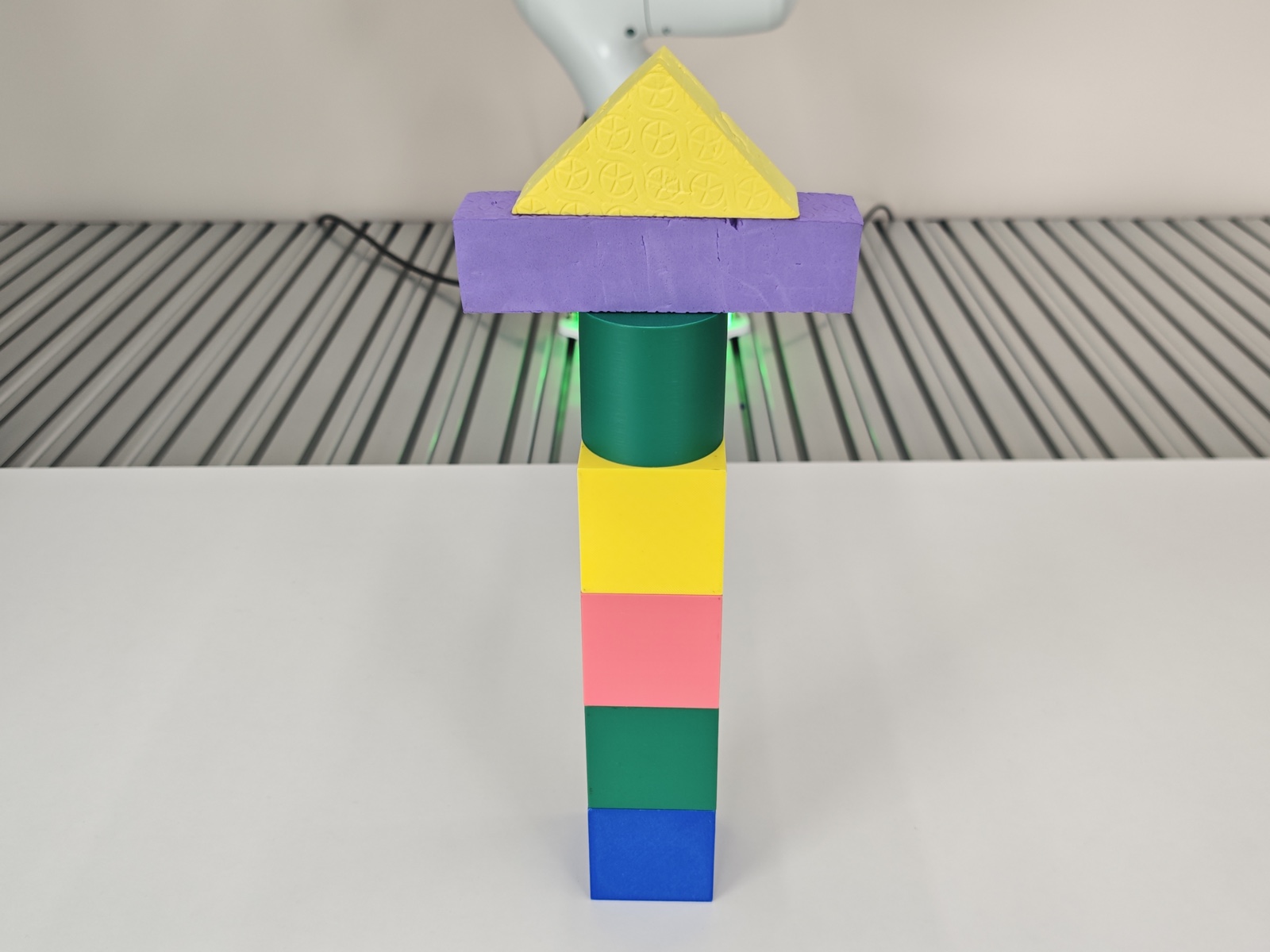}
            \vspace{1pt}
        \end{minipage} 
        & Hard & 0/3 & 0/3 & 0/3 & 1/3 & \textbf{2/3} \\
        
        \begin{minipage}{0.3\textwidth}
            \centering \vspace{1pt}
            \includegraphics[height=1.0cm]{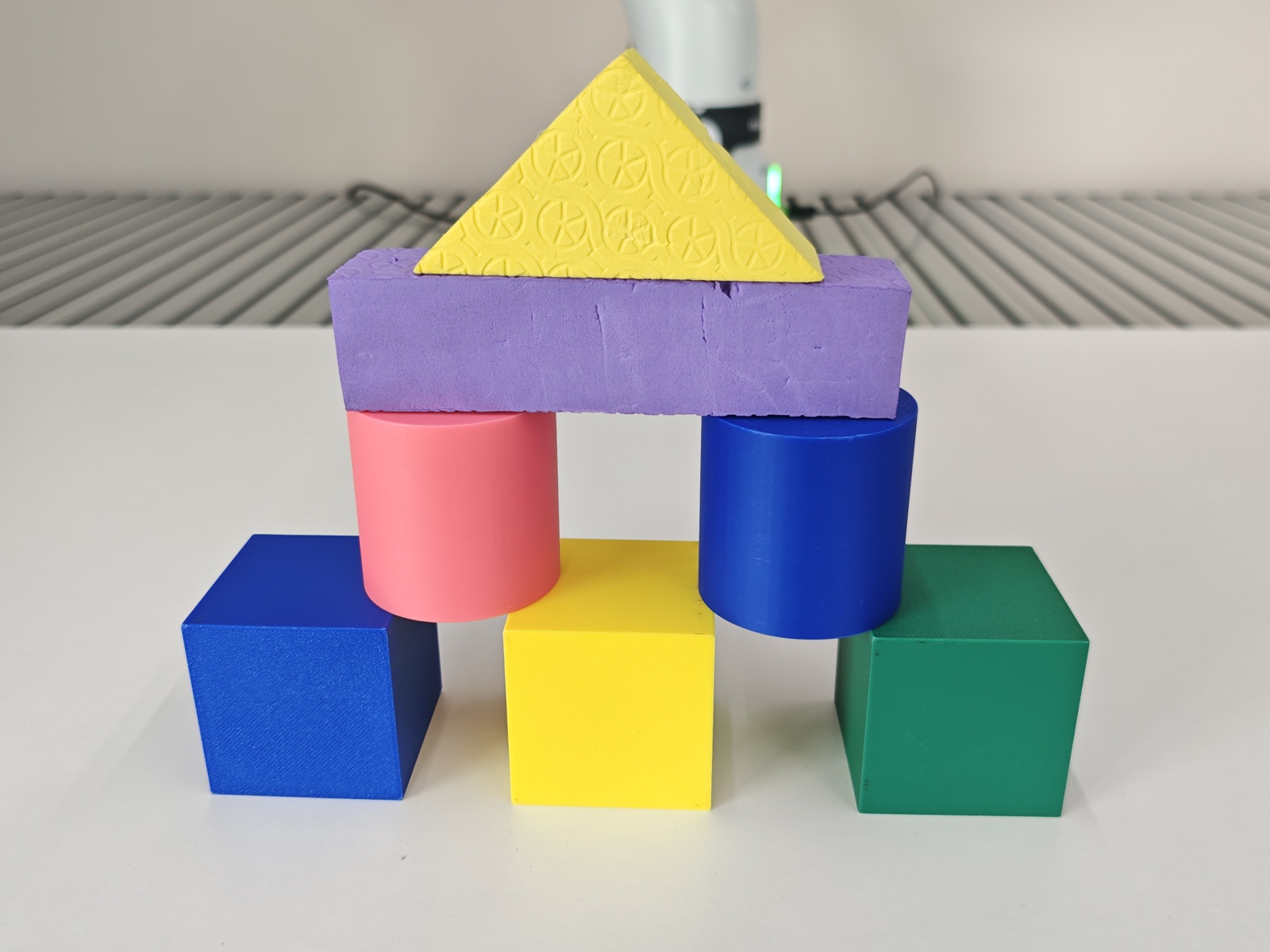}
            \vspace{1pt}
        \end{minipage} 
        & Hard & 0/3 & 0/3 & 0/3 & 0/3 & \textbf{0/3} \\
        
        \multicolumn{2}{l}{\textbf{Building Avg (\%)}} & 22.2 & 25.9 & 33.3 & \cellcolor{secondpurple}\underline{48.1} & \cellcolor{firstpurple}\textbf{63.0} \\
        \midrule
        
        \multicolumn{7}{c}{\textit{Block Disassembly (Top-to-Bottom)} {\scriptsize $\vert$ \textit{Task $=$ Goal Image}}} \\
        \midrule
        \begin{minipage}{0.3\textwidth}
            \centering \vspace{1pt}
            \includegraphics[height=0.8cm]{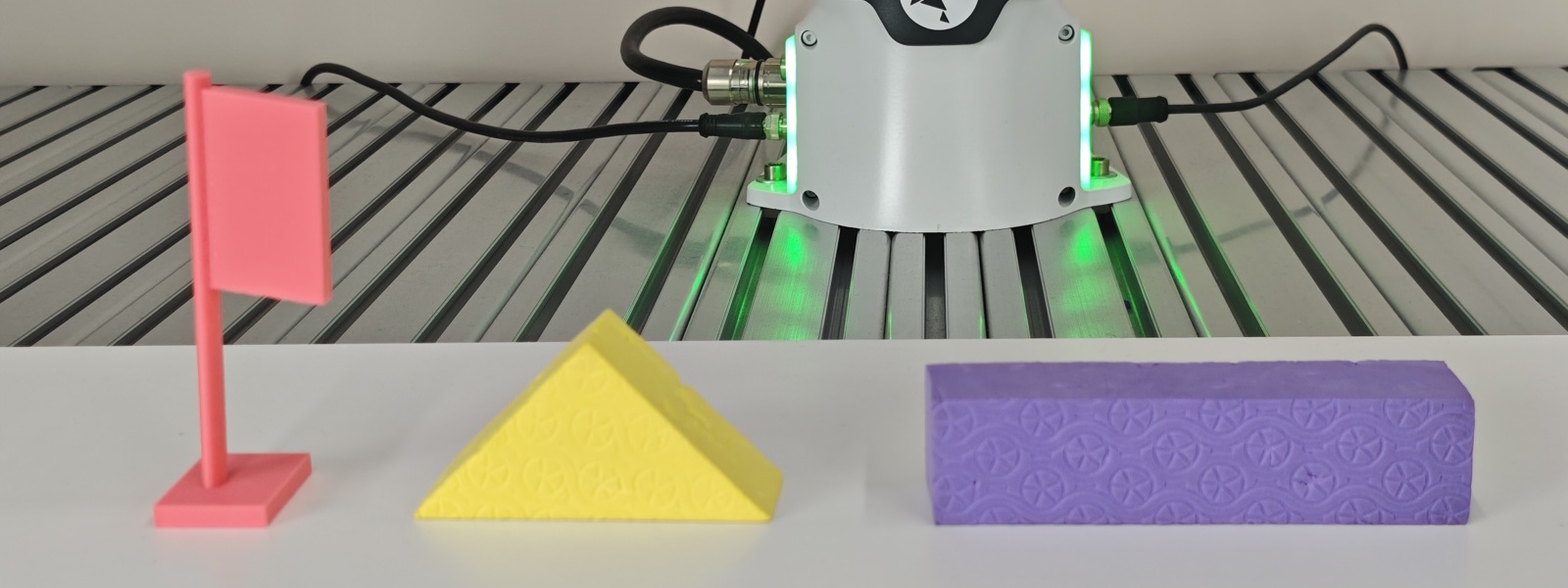}
            \vspace{1pt}
        \end{minipage} 
        & Easy & 2/3 & 3/3 & 2/3 & 2/3 & \textbf{3/3} \\
        
        \begin{minipage}{0.3\textwidth}
            \centering \vspace{1pt}
            \includegraphics[height=0.8cm]{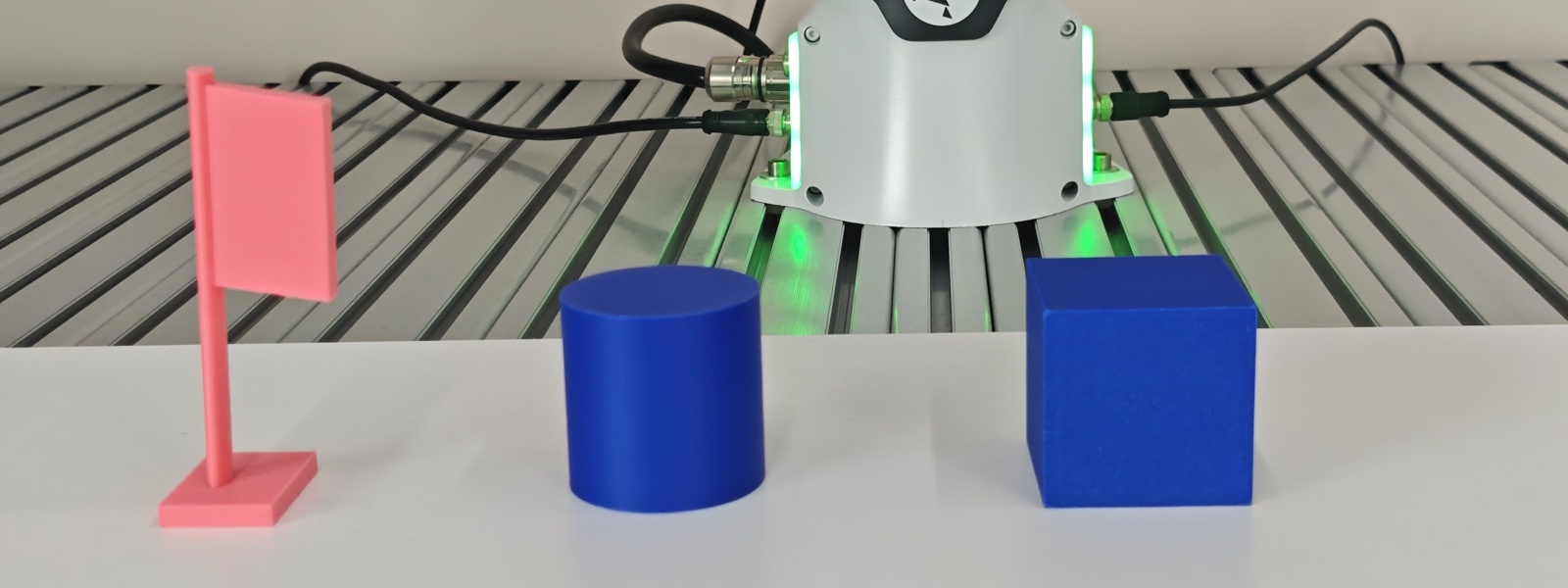}
            \vspace{1pt}
        \end{minipage} 
        & Easy & 2/3 & 2/3 & 3/3 & 3/3 & \textbf{3/3} \\
        
        \begin{minipage}{0.3\textwidth}
            \centering \vspace{1pt}
            \includegraphics[height=0.8cm]{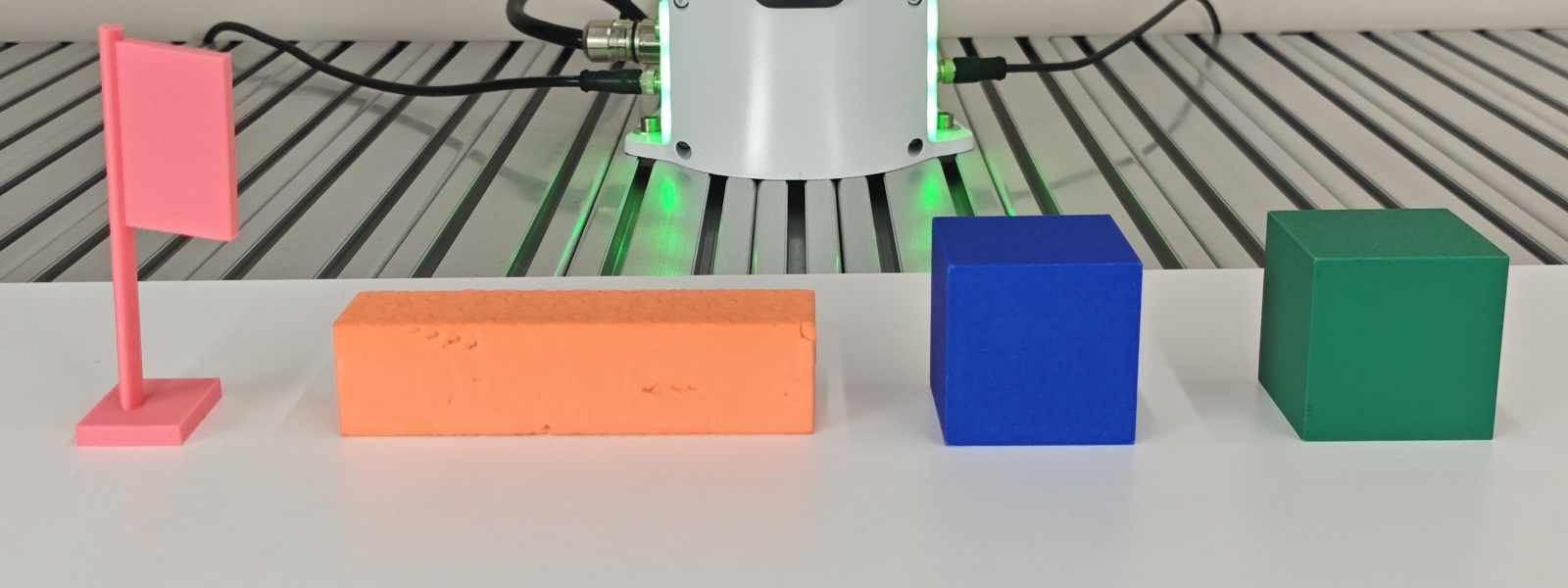}
            \vspace{1pt}
        \end{minipage} 
        & Easy & 1/3 & 1/3 & 1/3 & 2/3 & \textbf{2/3} \\
        
        \begin{minipage}{0.3\textwidth}
            \centering \vspace{1pt}
            \includegraphics[height=0.8cm]{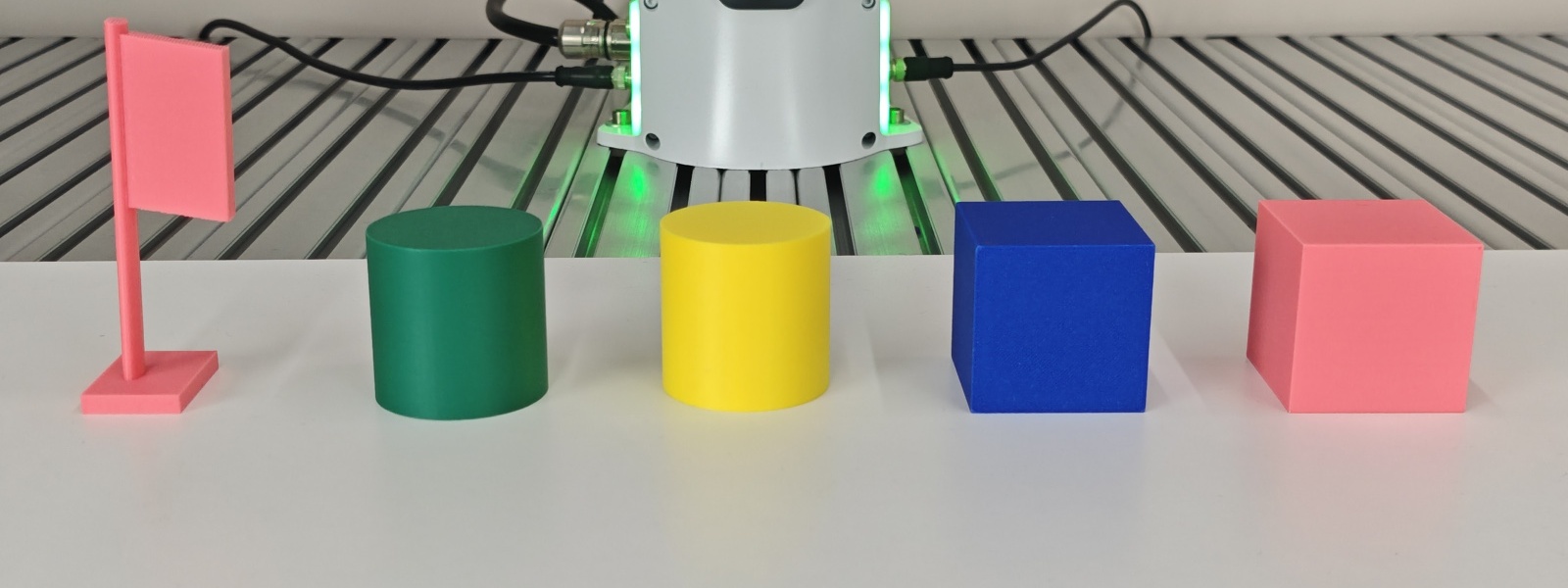}
            \vspace{1pt}
        \end{minipage} 
        & Medium & 1/3 & 0/3 & 0/3 & 1/3 & \textbf{2/3} \\
        
        \begin{minipage}{0.3\textwidth}
            \centering \vspace{1pt}
            \includegraphics[height=0.8cm]{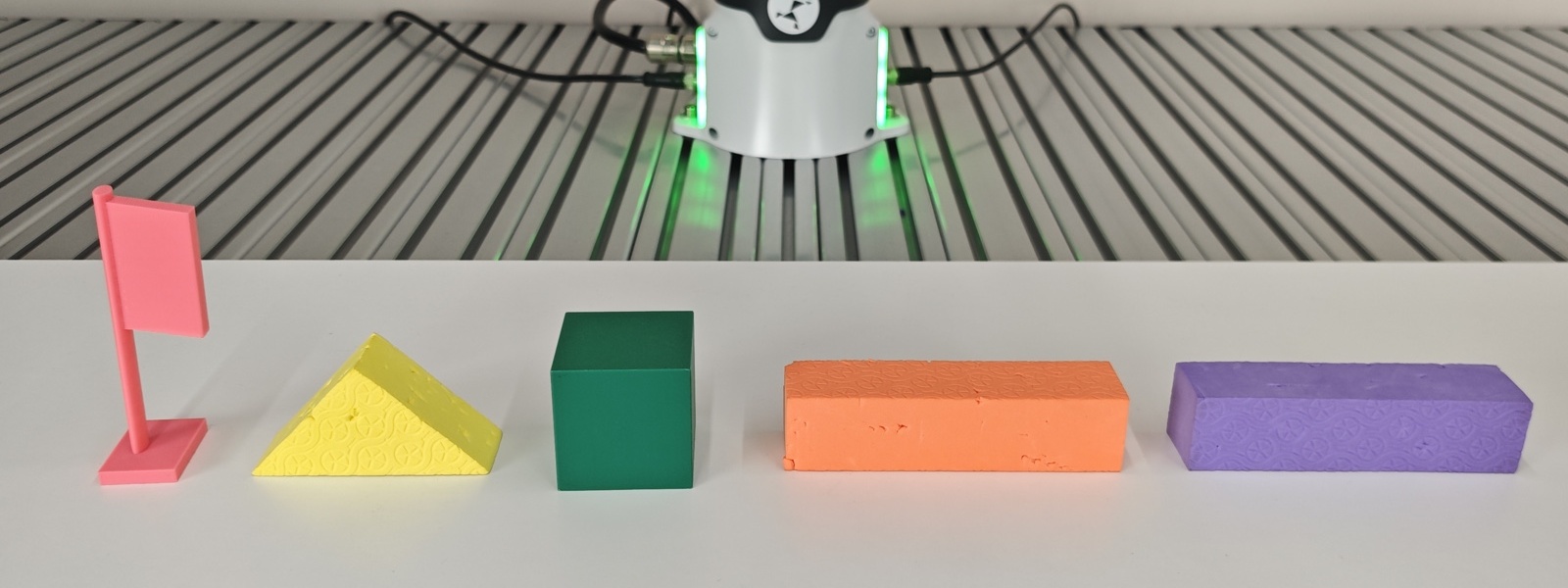}
            \vspace{1pt}
        \end{minipage} 
        & Medium & 0/3 & 0/3 & 1/3 & 2/3 & \textbf{2/3} \\
        
        \begin{minipage}{0.3\textwidth}
            \centering \vspace{1pt}
            \includegraphics[height=0.8cm]{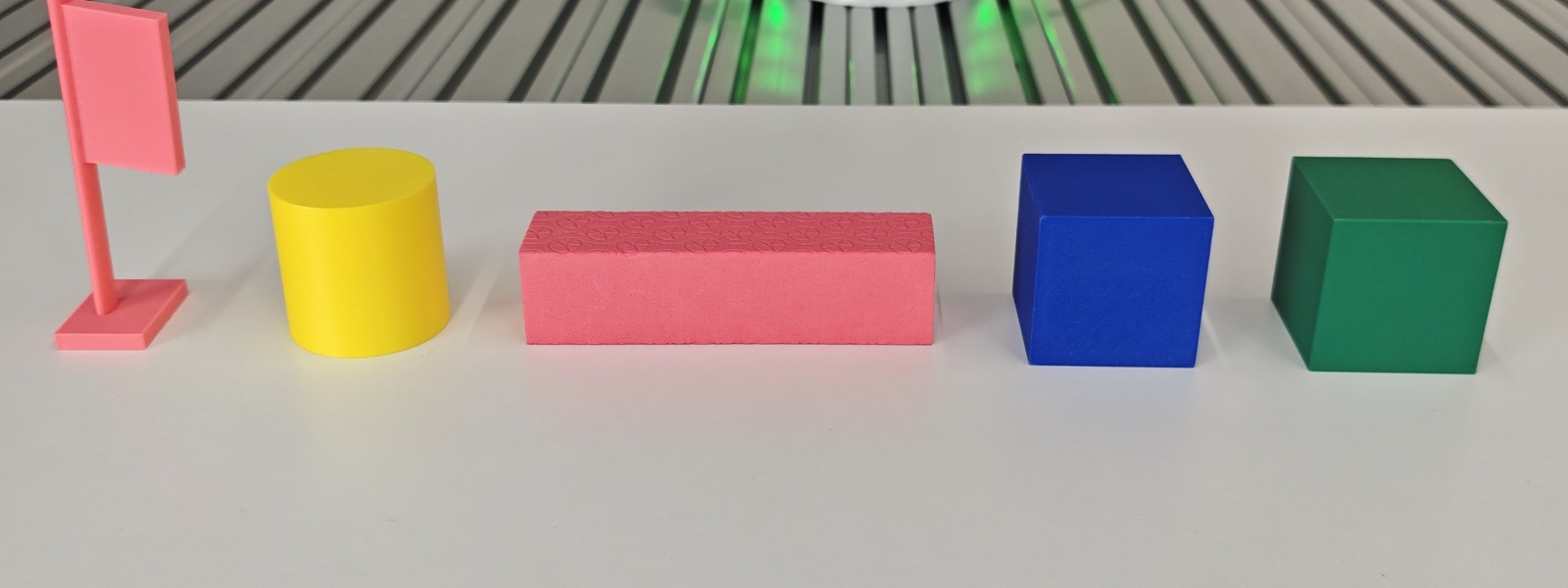}
            \vspace{1pt}
        \end{minipage} 
        & Medium & 1/3 & 0/3 & 1/3 & 2/3 & \textbf{3/3} \\
        
        \begin{minipage}{0.3\textwidth}
            \centering \vspace{1pt}
            \includegraphics[height=0.8cm]{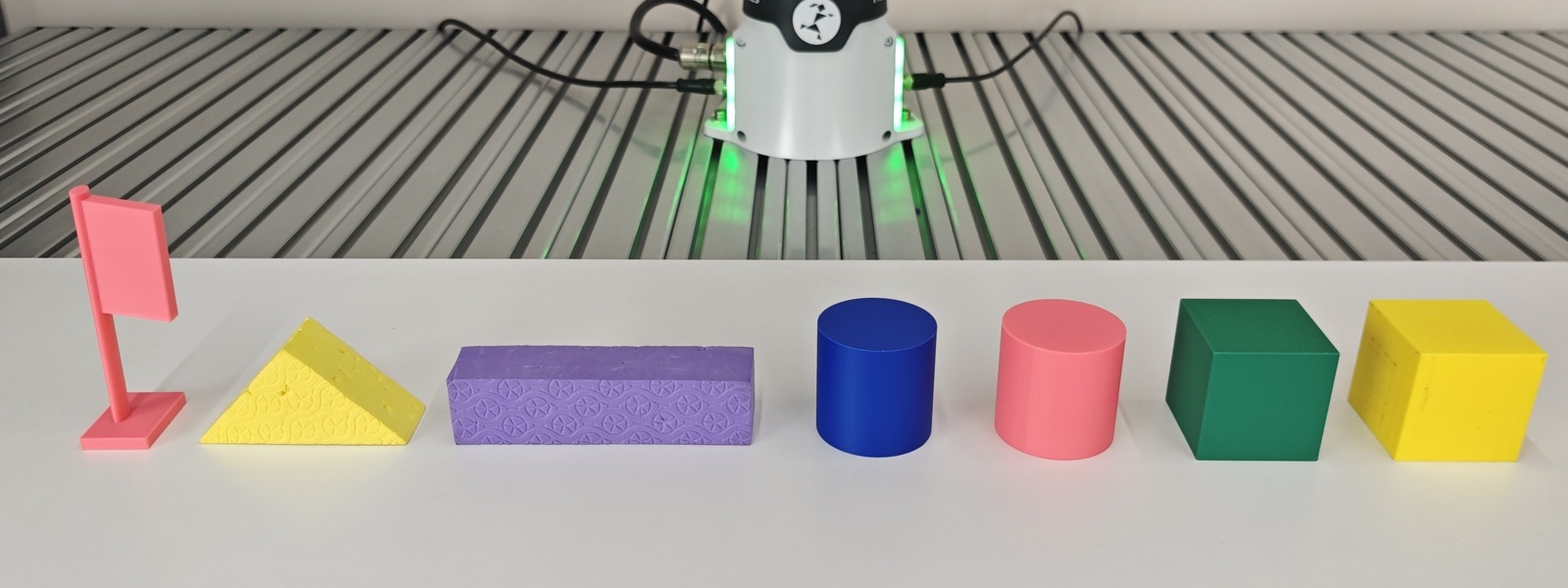}
            \vspace{1pt}
        \end{minipage} 
        & Hard & 1/3 & 1/3 & 2/3 & 2/3 & \textbf{2/3} \\
        
        \begin{minipage}{0.3\textwidth}
            \centering \vspace{1pt}
            \includegraphics[height=0.8cm]{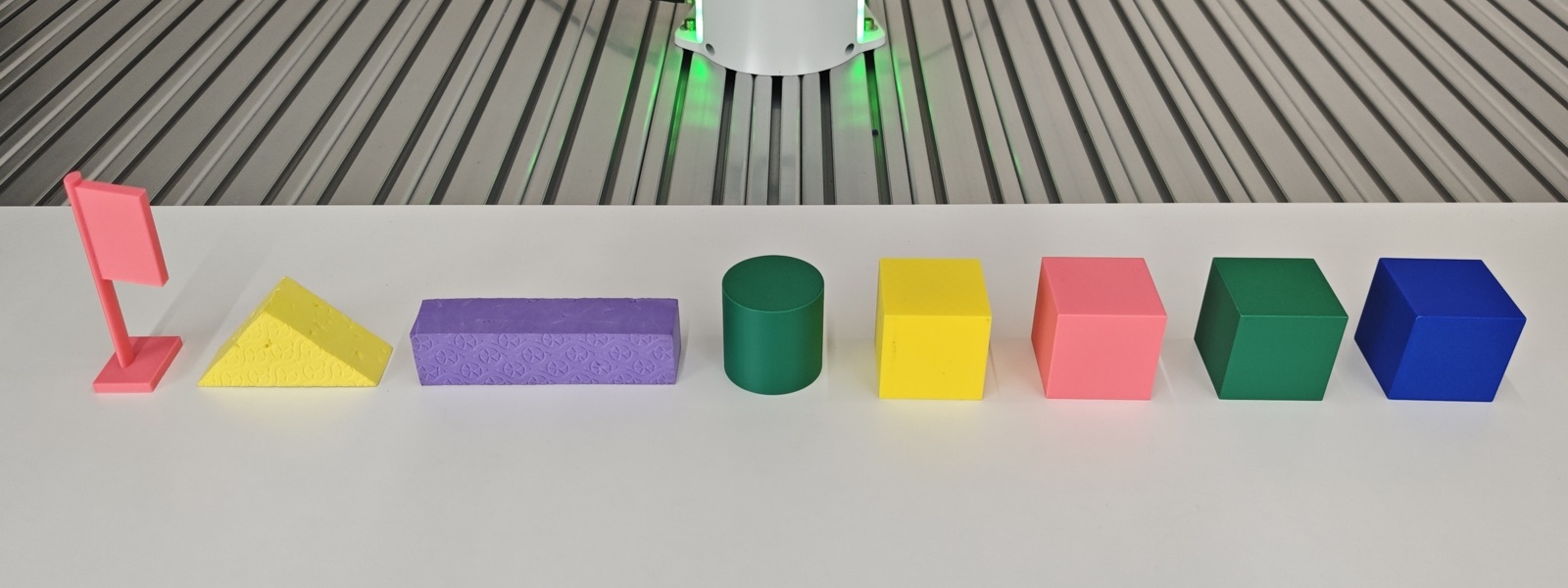}
            \vspace{1pt}
        \end{minipage} 
        & Hard & 0/3 & 0/3 & 0/3 & 2/3 & \textbf{2/3} \\
        
        \begin{minipage}{0.3\textwidth}
            \centering \vspace{1pt}
            \includegraphics[height=0.8cm]{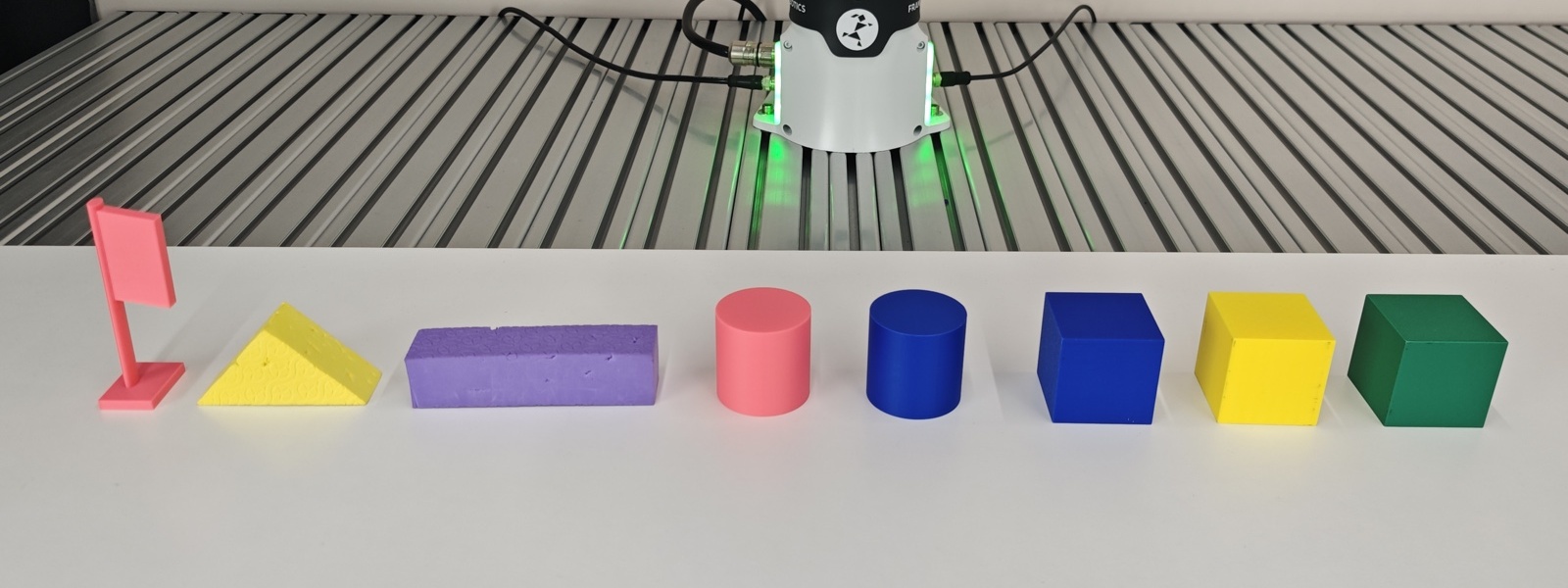}
            \vspace{1pt}
        \end{minipage} 
        & Hard & 0/3 & 0/3 & 1/3 & 1/3 & \textbf{2/3} \\
        
        \multicolumn{2}{l}{\textbf{Disassembly Avg (\%)}} & 29.6 & 25.9 & 40.7 & \cellcolor{secondpurple}\underline{63.0} & \cellcolor{firstpurple}\textbf{77.8} \\
        \bottomrule
    \end{tabular}
}
\end{table}

As shown in the results, RoboStream consistently outperforms existing baselines (SoFar and VoxPoser) across all difficulty levels. Notably, the success rate gap widens significantly as tasks transition from \textit{Easy} to \textit{Hard}. While baseline models often falter in the \textit{Medium} and \textit{Hard} stages due to compounding errors in sequential planning and loss of spatial awareness, RoboStream-235B maintains robust performance by leveraging the causal memory provided by the CSTG. 

The results further underscore two critical findings: 
(i) \textbf{Generalization under Occlusion:} In the \textit{Block Hide and Restore} tasks, RoboStream successfully retrieves hidden objects after intermediate distractor actions, a capability that neither SoFar nor VoxPoser possesses. This confirms that the causal memory log successfully mitigates the "visual amnesia" typically encountered by reactive models.
(ii) \textbf{Scaling Benefits:} We observe a clear performance gain as we scale from 8B to 235B parameters. The larger model capacity enables more nuanced semantic understanding of the assembly instructions, allowing it to handle complex physical constraints in the \textit{top-to-bottom} disassembly tasks, where maintaining structural stability is paramount. 

Overall, these real-world results demonstrate that by integrating explicit 3D geometric tokens (STF-Tokens) with causal temporal memory (CSTG), RoboStream achieves a superior balance between precise spatial control and long-horizon reasoning.

\begin{table}[htbp]
    \centering
    \caption{\textbf{Detailed results of real-world block hide and restore tasks.}}
    \label{tab:real_world_hide}
    \renewcommand{\arraystretch}{1.1} 
    \resizebox{\textwidth}{!}{
    \begin{tabular}{m{0.3\textwidth} c c c c c c}
        \toprule
        \multirow{2}{*}{\textbf{Task}} & \multirow{2}{*}{\textbf{Difficulty}} & \multicolumn{5}{c}{\textbf{Success Rate}} \\
        \cmidrule(lr){3-7}
        & & \textbf{SoFar} & \textbf{VoxPoser} & \textbf{RoboStream-8B} & \textbf{RoboStream-32B} & \textbf{RoboStream-235B} \\
        \midrule
        
        \multicolumn{7}{c}{\textit{Long-Horizon Memorize and Restore} {\scriptsize $\vert$ \textit{Task $=$ Initial State + Instruction}}} \\
        \midrule

        \begin{minipage}{0.3\textwidth}
            \centering \vspace{1pt}
            \includegraphics[height=1.2cm]{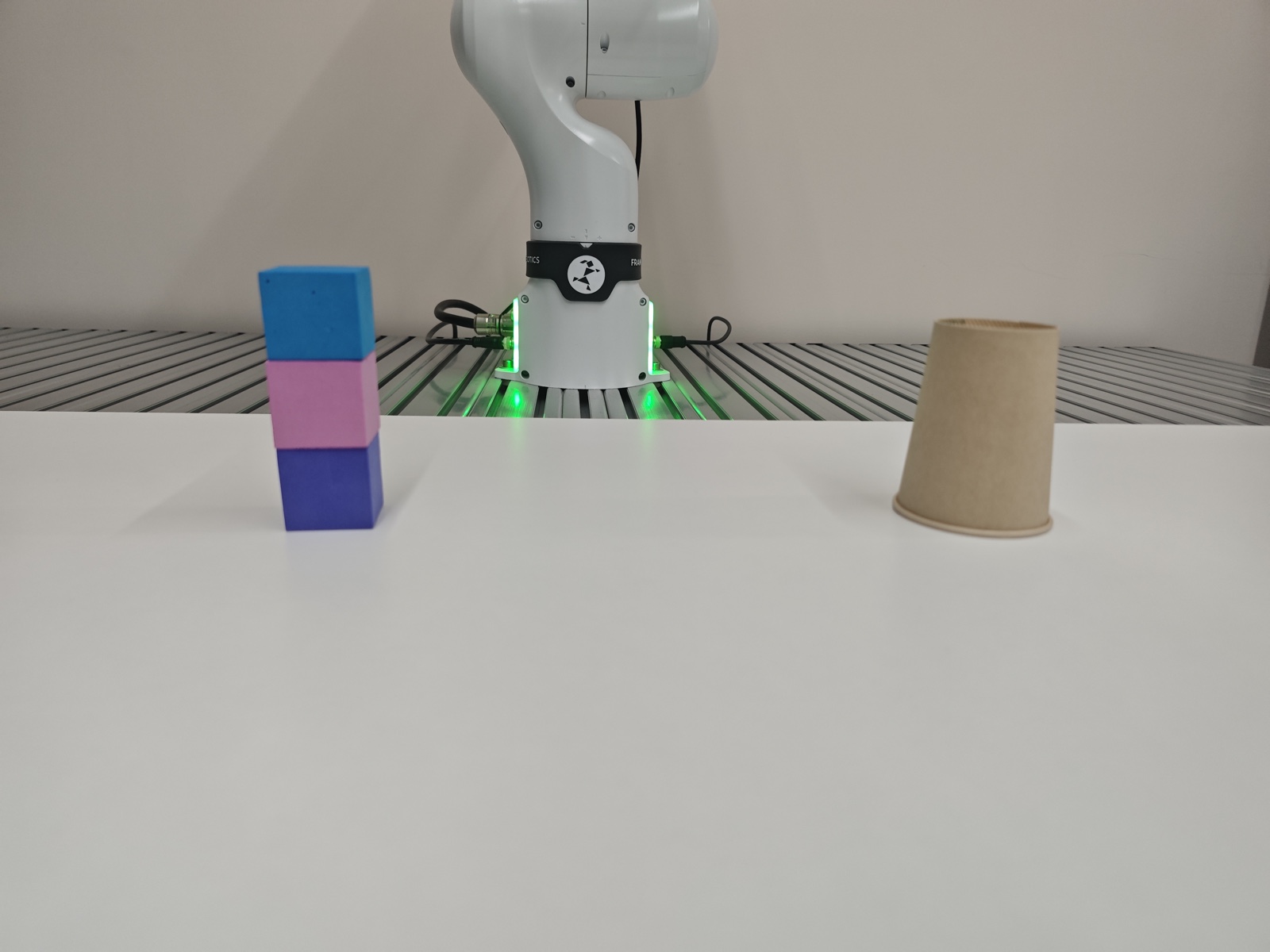}
            \\[-3pt]
            {\scriptsize \textit{Hide the \textbf{purple} cube with the brown cup, then restore.}}
            \vspace{1pt}
        \end{minipage}
        & Hard & 0/3 & 0/3 & 2/3 & 2/3 & \textbf{3/3} \\

        \begin{minipage}{0.3\textwidth}
            \centering \vspace{1pt}
            \includegraphics[height=1.2cm]{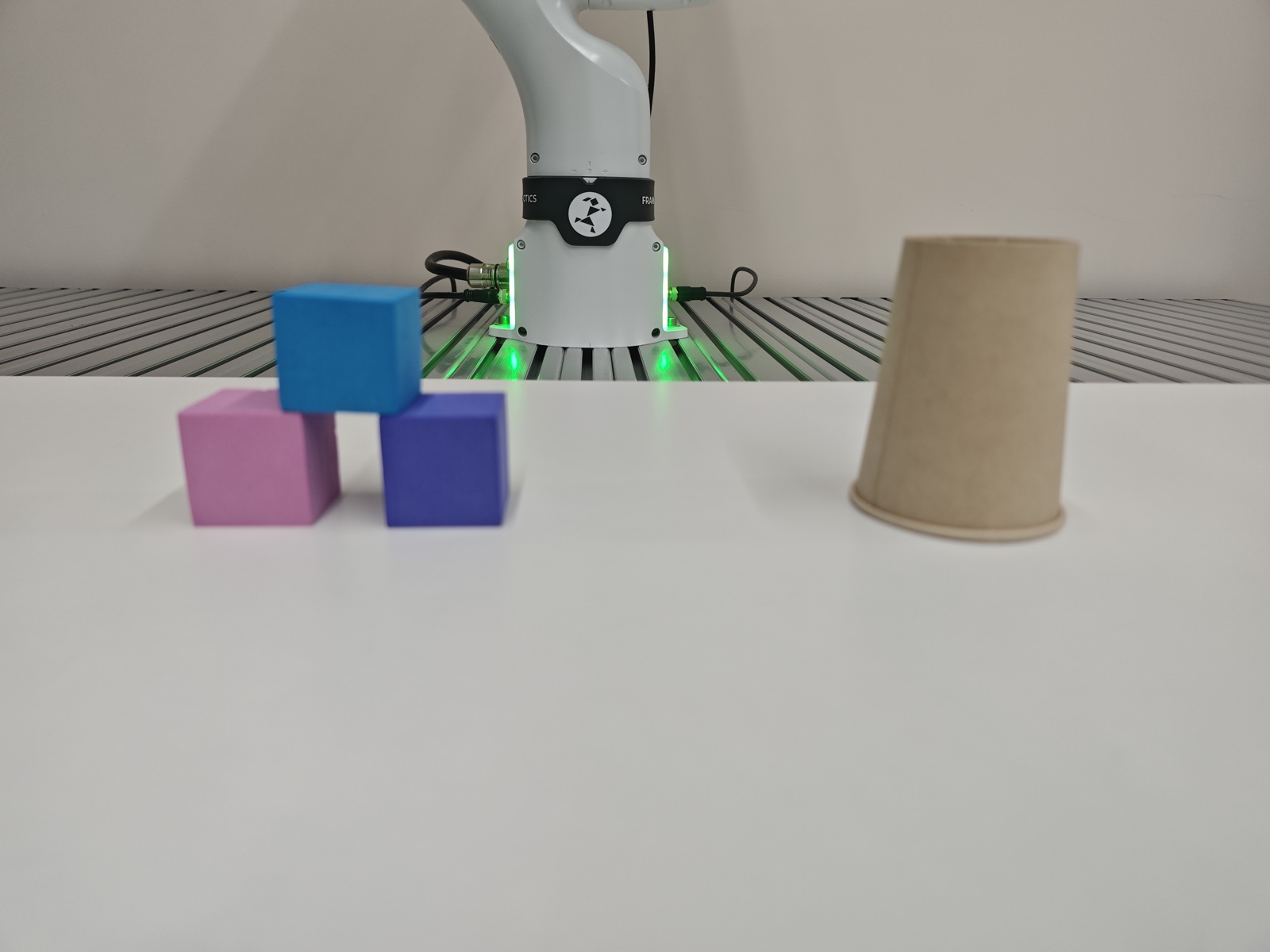}
            \\[-3pt]
            {\scriptsize \textit{Hide the \textbf{purple} cube with the brown cup, then restore.}}
            \vspace{1pt}
        \end{minipage}
        & Hard & 0/3 & 0/3 & 1/3 & 2/3 & \textbf{3/3} \\

        \begin{minipage}{0.3\textwidth}
            \centering \vspace{1pt}
            \includegraphics[height=1.2cm]{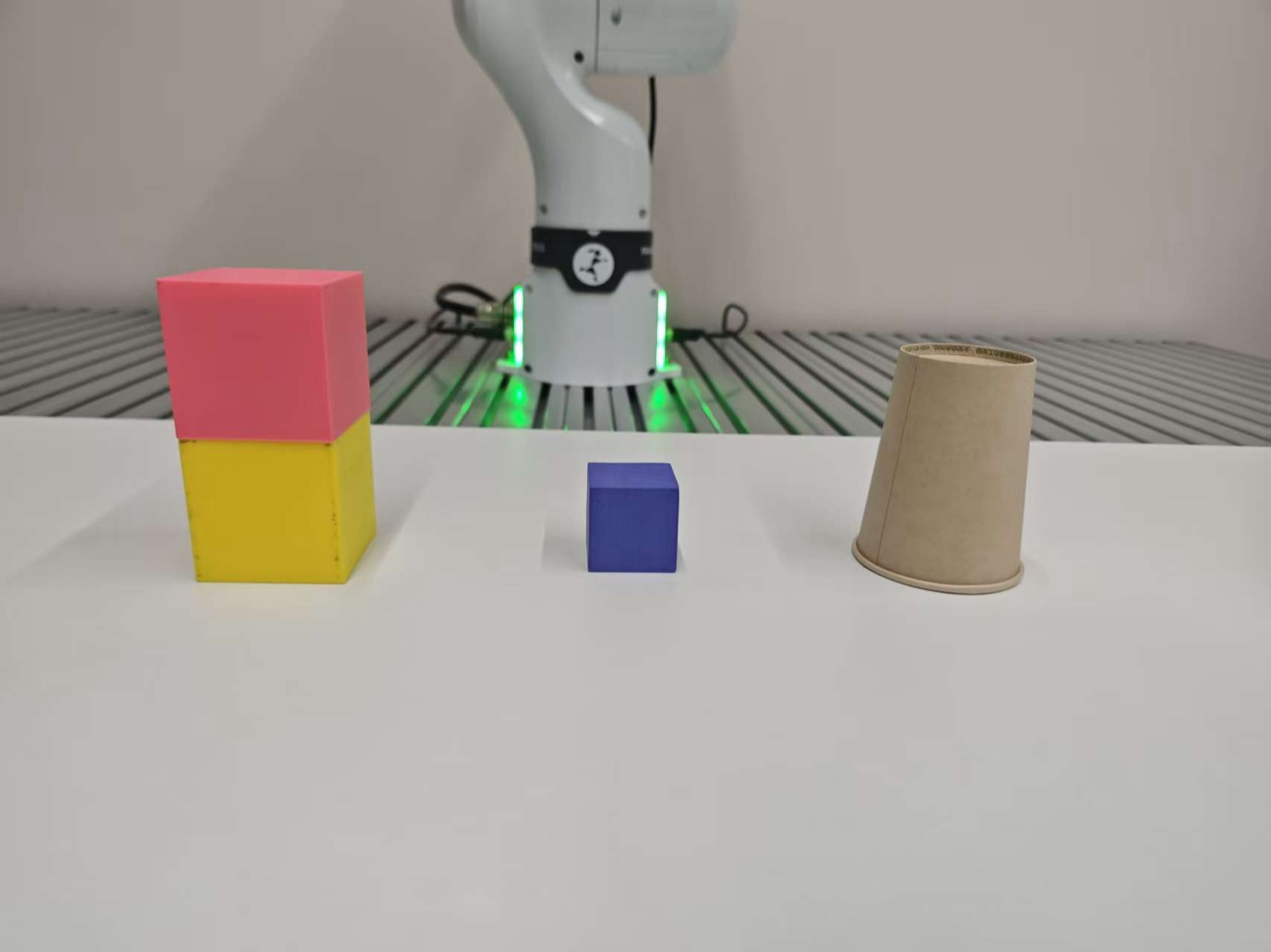}
            \\[-3pt]
            {\scriptsize \textit{Hide the \textbf{purple} cube with the brown cup, place the yellow cube to the right of the cup, then restore.}}
            \vspace{1pt}
        \end{minipage}
        & Hard & 0/3 & 0/3 & 0/3 & 1/3 & \textbf{2/3} \\

        \multicolumn{2}{l}{\textbf{Hide task Avg (\%)}} & 0.0 & 0.0 & \cellcolor{secondpurple}\underline{33.3} & \cellcolor{secondpurple}\underline{55.6} & \cellcolor{firstpurple}\textbf{88.9} \\
        \bottomrule
    \end{tabular}
}
\end{table}

\FloatBarrier
\begin{figure}[htbp]
    \centering
    \includegraphics[width=0.6\textwidth]{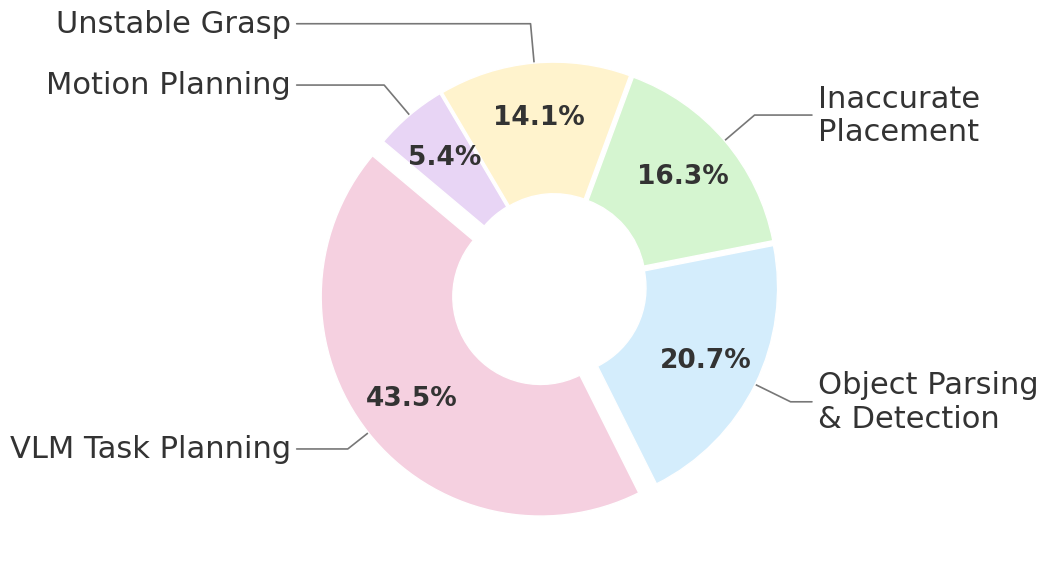}
    \caption{\textbf{Failure case distribution analysis} of RoboStream from real-world experiments.}
    \label{fig:failure_distribution}
\end{figure}

\subsection{Failure Case Distribution Analysis}

Based on the failure cases from real-world experiments, we conducted a quantitative analysis of the failure case distribution for RoboStream, with the results shown in Fig.~\ref{fig:failure_distribution}. It can be observed that 20.7\% of the failures were due to Object Parsing \& Detection errors, where open-vocabulary segmentation models such as SAM3~\cite{carion2025sam} failed to accurately localize targets under color-similar distractors or severe stacking occlusion, directly causing downstream grasp point deviations. Inaccurate Placement accounted for approximately 16.3\% of the errors; although the planning and detection were correct, small pose offsets during the final release caused center-of-gravity instability and tower collapse in precision-demanding block-stacking tasks. In addition, 14.1\% of failures arose from Unstable Grasp, including unreasonable 6-DoF grasp poses and object slippage during transport. Motion Planning contributed 5.4\% of failures, covering inverse kinematics solver failures, workspace limit violations, and unforeseen collisions during trajectory execution. Finally, the dominant source of failure was VLM task planning, accounting for 43.5\% of all errors, where the vision-language model produced spatial hallucinations, derived incorrect left-right spatial relationships or disassembly orders, failed to reason about occlusion-induced causal dependencies, or ``forgot'' previous states during long-horizon operations. The prevalence of this failure mode reflects the inherent difficulty of long-horizon spatio-temporal reasoning.

\begin{table*}[htbp]
    \centering
    \caption{\textbf{Comprehensive Ablation Results in Simulation.} Success rates (\%) across 8 long-horizon RLBench tasks. We compare the full RoboStream pipeline against variants removing Spatio-Temporal Fusion Tokens (w/o STF-Tokens), CSTG (w/o CSTG), or both (w/o Both) across 8B, 32B, and 235B scales. Data is averaged over 25 independent episodes per task. The 235B results correspond to the macro ablation study presented in the main text.}
    \label{tab:sim_detailed_ablation}
    
    \resizebox{\textwidth}{!}{%
    \renewcommand{\arraystretch}{1.2} 
    \begin{tabular}{c cc | cccccccc c}
        \toprule
        \textbf{Scale} & \thead{\textit{STF-} \\ \textit{Tokens}} & \thead{\textit{CSTG}} & 
        \thead{Bridge\\Between} & 
        \thead{Cover\\Bottom} & 
        \thead{Cover\\Top} & 
        \thead{Place in\\Cont.} & 
        \thead{Place in\\Cont. (H)} & 
        \thead{Stack\\Five} & 
        \thead{Stack\\Three} & 
        \thead{Unstack\\\& Stack} & 
        \textbf{Average} \\
        \midrule
        
        \multirow{4}{*}{\textbf{8B}} 
        & \xmark & \xmark & 0.0 & 0.0 & 0.0 & 88.0 & 0.0 & 0.0 & 0.0 & 0.0 & 11.0 \\
        & \cmark & \xmark & 0.0 & 0.0 & 0.0 & \cellcolor{firstpurple}\textbf{100.0} & 0.0 & 0.0 & 0.0 & 0.0 & 12.5 \\
        & \xmark & \cmark & \cellcolor{secondpurple}\underline{68.0} & \cellcolor{secondpurple}\underline{8.0} & \cellcolor{secondpurple}\underline{24.0} & \cellcolor{firstpurple}\textbf{100.0} & \cellcolor{secondpurple}\underline{12.0} & \cellcolor{secondpurple}\underline{40.0} & \cellcolor{secondpurple}\underline{88.0} & \cellcolor{secondpurple}\underline{24.0} & \cellcolor{secondpurple}\underline{45.5} \\
        & \cmark & \cmark & \cellcolor{firstpurple}\textbf{76.0} & \cellcolor{firstpurple}\textbf{16.0} & \cellcolor{firstpurple}\textbf{40.0} & \cellcolor{firstpurple}\textbf{100.0} & \cellcolor{firstpurple}\textbf{24.0} & \cellcolor{firstpurple}\textbf{60.0} & \cellcolor{firstpurple}\textbf{96.0} & \cellcolor{firstpurple}\textbf{52.0} & \cellcolor{firstpurple}\textbf{58.0} \\
        \midrule
        
        \multirow{4}{*}{\textbf{32B}} 
        & \xmark & \xmark & 0.0 & 0.0 & 0.0 & 48.0 & 0.0 & 0.0 & 0.0 & 0.0 & 6.0 \\
        & \cmark & \xmark & 0.0 & 0.0 & 0.0 & \cellcolor{secondpurple}\underline{56.0} & 12.0 & 0.0 & 0.0 & 0.0 & 8.5 \\
        & \xmark & \cmark & \cellcolor{secondpurple}\underline{80.0} & \cellcolor{secondpurple}\underline{52.0} & \cellcolor{secondpurple}\underline{68.0} & \cellcolor{firstpurple}\textbf{100.0} & \cellcolor{secondpurple}\underline{60.0} & \cellcolor{secondpurple}\underline{48.0} & \cellcolor{secondpurple}\underline{88.0} & \cellcolor{secondpurple}\underline{56.0} & \cellcolor{secondpurple}\underline{69.0} \\
        & \cmark & \cmark & \cellcolor{firstpurple}\textbf{88.0} & \cellcolor{firstpurple}\textbf{68.0} & \cellcolor{firstpurple}\textbf{84.0} & \cellcolor{firstpurple}\textbf{100.0} & \cellcolor{firstpurple}\textbf{76.0} & \cellcolor{firstpurple}\textbf{72.0} & \cellcolor{firstpurple}\textbf{96.0} & \cellcolor{firstpurple}\textbf{76.0} & \cellcolor{firstpurple}\textbf{82.5} \\
        \midrule
        
        \multirow{4}{*}{\textbf{235B}} 
        & \xmark & \xmark & 0.0 & 0.0 & 0.0 & 84.0 & 12.0 & 0.0 & 0.0 & 0.0 & 12.0 \\
        & \cmark & \xmark & 0.0 & 0.0 & 0.0 & \cellcolor{secondpurple}\underline{92.0} & 24.0 & 0.0 & 0.0 & 0.0 & 14.5 \\
        & \xmark & \cmark & \cellcolor{secondpurple}\underline{72.0} & \cellcolor{secondpurple}\underline{88.0} & \cellcolor{secondpurple}\underline{84.0} & \cellcolor{firstpurple}\textbf{100.0} & \cellcolor{secondpurple}\underline{72.0} & \cellcolor{secondpurple}\underline{60.0} & \cellcolor{firstpurple}\textbf{96.0} & \cellcolor{secondpurple}\underline{64.0} & \cellcolor{secondpurple}\underline{79.5} \\
        & \cmark & \cmark & \cellcolor{firstpurple}\textbf{88.0} & \cellcolor{firstpurple}\textbf{96.0} & \cellcolor{firstpurple}\textbf{92.0} & \cellcolor{firstpurple}\textbf{100.0} & \cellcolor{firstpurple}\textbf{88.0} & \cellcolor{firstpurple}\textbf{80.0} & \cellcolor{firstpurple}\textbf{96.0} & \cellcolor{firstpurple}\textbf{84.0} & \cellcolor{firstpurple}\textbf{90.5} \\
        
        \bottomrule
    \end{tabular}%
    }
\end{table*}

\subsection{Detailed Results of Simulation Manipulation}

To provide a comprehensive understanding of how our proposed components contribute to the overall performance across different model capacities, we present the detailed ablation results on the RLBench long-horizon tasks in Tab.~\ref{tab:sim_detailed_ablation}. This section expands upon the macro-level analysis by reporting the execution success rates for all three model scales (8B, 32B, and 235B) under three architectural configurations: the Full Model, the variant without Spatio-Temporal Fusion Tokens (w/o STF-Tokens), and the variant without Causal Spatio-Temporal Graph (w/o CSTG).

As observed in the detailed breakdown, removing the causal memory results in a near-complete collapse across tasks that strictly require cross-step state tracking and occlusion reasoning (e.g., \textit{Bridge Between Towers} and \textit{Stack Colors}). The memory-ablated model only retains partial capability on tasks where intermediate spatial constraints can be somewhat inferred directly from the current visual frame (e.g., \textit{Place in Cont.}). Furthermore, the results highlight a clear capability scaling law in handling extreme occlusion: while the 235B Full Model achieves near-perfect execution on the \textit{Cover Top/Bottom} tasks (92.0\% and 96.0\% success rates), the 8B model struggles significantly (40.0\% and 16.0\%), and the 32B model shows intermediate performance (84.0\% and 68.0\%). This indicates that robust occlusion recovery still relies on the emergent semantic reasoning capacity of larger parameter scales to some extent.

Conversely, removing STF-Tokens while retaining the Causal Spatio-Temporal Graph (CSTG) generally preserves the logical progression of the tasks but introduces significant variance in spatial execution precision. This degradation in physical accuracy is evident across all model scales, particularly on contact-rich tasks like \textit{Stack Five Colors} (e.g., dropping from 72.0\% to 48.0\% for the 32B model, and 60.0\% to 40.0\% for 8B). This explicitly confirms our hypothesis: while memory ensures the correct semantic sequence is planned, explicit 3D geometric grounding (STF-Tokens) is strictly required to stabilize the physical execution of those plans, even for highly capable VLMs.

\subsection{Runtime Scaling of RoboStream}

We report two complementary runtime analyses with different evaluation scopes. 
First, to examine whether RoboStream accumulates additional overhead as a long-horizon execution progresses, we profile RoboStream-8B on a single challenging 7-step hard block disassembly case. This case is evaluated on one NVIDIA A100 GPU with a sliding window length of 3. We decompose the latency at each execution step into perception, CSTG construction, and VLM inference. The perception module includes Qwen3-VL-8B-based open-vocabulary object parsing and SAM3-based segmentation. The CSTG construction module includes event detection and scene graph updates.

As shown in \cref{tab:latency_breakdown_transposed}, perception and CSTG construction remain stable across the seven steps, averaging 0.25 s and 0.16 s per step, respectively. VLM inference dominates the runtime and increases only mildly from 6.56 s at the first step to around 7.4 s in later steps. The total latency remains bounded between 6.94 s and 7.91 s. This result indicates that, under the sliding-window setting, STF-Tokens and CSTG updates do not introduce cumulative latency growth as the execution horizon becomes longer.

\begin{table}[!htbp]
\centering
\caption{\textbf{Per-step latency breakdown on a hard 7-step disassembly case.} Processing time (s) is reported for RoboStream-8B on one NVIDIA A100 GPU with a sliding window length of 3. Highlighted entries indicate the largest and second-largest values in each row.}
\label{tab:latency_breakdown_transposed}
\renewcommand{\arraystretch}{1.2}
\begin{tabular*}{\textwidth}{@{\extracolsep{\fill}} l | ccccccc | c @{}}
\toprule
\textbf{Module} & \textbf{Step 1} & \textbf{Step 2} & \textbf{Step 3} & \textbf{Step 4} & \textbf{Step 5} & \textbf{Step 6} & \textbf{Step 7} & \textbf{Avg} \\
\midrule
Perception     & 0.24 & 0.22 & \cellcolor{firstpurple}\textbf{0.30} & \cellcolor{secondpurple}\underline{0.25} & \cellcolor{secondpurple}\underline{0.25} & 0.22 & 0.24 & 0.25 \\
CSTG Construct & 0.14 & 0.15 & \cellcolor{firstpurple}\textbf{0.17} & \cellcolor{firstpurple}\textbf{0.17} & \cellcolor{firstpurple}\textbf{0.17} & \cellcolor{secondpurple}\underline{0.16} & \cellcolor{firstpurple}\textbf{0.17} & \cellcolor{secondpurple}\underline{0.16} \\
VLM Inference  & 6.56 & 6.92 & \cellcolor{firstpurple}\textbf{7.44} & \cellcolor{secondpurple}\underline{7.31} & \cellcolor{firstpurple}\textbf{7.44} & 7.42 & 7.38 & 7.21 \\
Total Latency  & 6.94 & 7.29 & \cellcolor{firstpurple}\textbf{7.91} & 7.73 & \cellcolor{secondpurple}\underline{7.86} & 7.80 & 7.79 & 7.62 \\
\bottomrule
\end{tabular*}
\end{table}

Second, to evaluate runtime under different backbone scales and hardware configurations, we measure the average per-step latency over all real-world cases in our benchmark. Specifically, RoboStream-8B, RoboStream-32B, and RoboStream-235B are evaluated using 1, 2, and 8 NVIDIA A100 GPUs, respectively, as reported in \cref{tab:latency_scale_all}. These results are averaged across the real-world task suite rather than measured on the single hard disassembly case used in \cref{tab:latency_breakdown_transposed}. Therefore, the absolute latency values in the two tables are not directly comparable: the first table analyzes temporal scaling on a difficult long-horizon case, whereas the second table summarizes average deployment latency across all real-world cases.

Across backbone scales, perception and memory-related computation remain below 0.8 s per step, while VLM planning accounts for most of the runtime. The 235B MoE backbone is faster than the dense 32B backbone under the tested 8$\times$A100 configuration, which is consistent with activating approximately 22B parameters per step, although MoE routing and multi-GPU communication still make it slower than the 8B model. Overall, these results show that RoboStream's geometric-token and memory components add limited overhead relative to VLM inference, and that the system supports online task-level planning across different model scales.

\begin{table}[!htbp]
\centering
\caption{\textbf{Average per-step latency over all real-world cases.} Latency is reported in seconds per planning step for different RoboStream backbone scales and NVIDIA A100 GPU configurations. Highlighted entries indicate the largest and second-largest values in each latency column.}
\label{tab:latency_scale_all}
\renewcommand{\arraystretch}{1.2}
\begin{tabular*}{\textwidth}{@{\extracolsep{\fill}} lccccc @{}}
\toprule
\textbf{Backbone} & \textbf{GPU} & \textbf{Percep.} & \textbf{Mem.} & \textbf{VLM} & \textbf{Total} \\
\midrule
8B   & 1$\times$A100 & 0.41 & 0.30 & 3.15 & 3.86 \\
32B  & 2$\times$A100 & \cellcolor{secondpurple}\underline{0.42} & \cellcolor{secondpurple}\underline{0.34} & \cellcolor{firstpurple}\textbf{7.85} & \cellcolor{firstpurple}\textbf{8.61} \\
235B & 8$\times$A100 & \cellcolor{firstpurple}\textbf{0.44} & \cellcolor{firstpurple}\textbf{0.35} & \cellcolor{secondpurple}\underline{4.36} & \cellcolor{secondpurple}\underline{5.15} \\
\bottomrule
\end{tabular*}
\end{table}

\subsection{Short-Horizon RLBench Evaluation}
We also evaluate RoboStream-8B on original RLBench.
As shown in \cref{tab:rlbench_results_table}, RoboStream-8B achieves 71.2\% average success rate across 15 tasks, outperforming 
VoxPoser~\cite{huang2023voxposer} (49.9\%) and SoFar~\cite{qisofar} (23.5\%). Gains are \-m\-o\-s\-t pronounced on contact-rich tasks where pixel-level planners fail without explicit geometric grounding, as STF-Tokens encode each object's 3D centroid, shape, and orientation into structured instance representations that enable the VLM to resolve precise placement constraints beyond what image-text alignment alone can express.

\begin{table}[!htbp]
    \centering
    \caption{\textbf{Short-horizon task evaluation results on RLBench.}}
    \label{tab:rlbench_results_table}
    \renewcommand{\arraystretch}{1.2} 
    \begin{tabular*}{\textwidth}{@{\extracolsep{\fill}} l | ccc @{}}
        \toprule
        \textbf{Task (RLBench)} & \textbf{SoFar} & \textbf{VoxPoser} & \textbf{RoboStream-8B} \\
        \midrule
        Close Drawer          & 72.0            & \cellcolor{firstpurple}\textbf{100.0}   & \cellcolor{secondpurple}\underline{92.0} \\
        Lamp Off              & 0.0             & \cellcolor{firstpurple}\textbf{100.0}   & \cellcolor{secondpurple}\underline{76.0} \\
        Meat Off Grill        & \cellcolor{secondpurple}\underline{4.0} & 0.0             & \cellcolor{firstpurple}\textbf{96.0} \\
        Open Wine Bottle      & 0.0             & \cellcolor{firstpurple}\textbf{100.0}   & \cellcolor{secondpurple}\underline{48.0} \\
        Pick and Lift         & 0.0             & 0.0             & \cellcolor{firstpurple}\textbf{80.0} \\
        Pick Up Cup           & 8.0             & \cellcolor{secondpurple}\underline{12.0} & \cellcolor{firstpurple}\textbf{68.0} \\
        Place Cups            & \cellcolor{firstpurple}\textbf{32.0}   & 0.0             & \cellcolor{secondpurple}\underline{24.0} \\
        Push Button           & \cellcolor{secondpurple}\underline{20.0} & 0.0             & \cellcolor{firstpurple}\textbf{100.0} \\
        Put Knife on Board    & 20.0            & \cellcolor{firstpurple}\textbf{100.0}   & \cellcolor{secondpurple}\underline{36.0} \\
        Put Plate in Rack     & 16.0            & \cellcolor{secondpurple}\underline{28.0} & \cellcolor{firstpurple}\textbf{32.0} \\
        Put Rubbish in Bin    & \cellcolor{secondpurple}\underline{96.0} & \cellcolor{firstpurple}\textbf{100.0}   & \cellcolor{secondpurple}\underline{96.0} \\
        Slide Block to Target & 0.0             & \cellcolor{firstpurple}\textbf{36.0}    & \cellcolor{secondpurple}\underline{24.0} \\
        Take Lid Off Saucepan & 0.0             & 0.0             & \cellcolor{firstpurple}\textbf{100.0} \\
        Take Off Scales       & 0.0             & \cellcolor{secondpurple}\underline{72.0} & \cellcolor{firstpurple}\textbf{96.0} \\
        Take Umbrella out     & \cellcolor{secondpurple}\underline{84.0} & \cellcolor{firstpurple}\textbf{100.0}   & \cellcolor{firstpurple}\textbf{100.0} \\
        \midrule
        \textbf{Avg}          & 23.5            & \cellcolor{secondpurple}\underline{49.9} & \cellcolor{firstpurple}\textbf{71.2} \\
        \bottomrule
    \end{tabular*}
\end{table}

\FloatBarrier

\subsection{Memory Baseline Comparison on RLBench}
\label{sec:appendix_memory_baselines}

We further compare RoboStream against memory-augmented manipulation approaches, SAM2Act~\cite{fang2025sam2act} and MemoryVLA~\cite{shi2026memoryvla}, on four RLBench long-horizon cases that require maintaining object states and action history across multiple steps. As shown in Tab.~\ref{tab:memory_baselines_rlbench}, both baselines solve easier placement cases but degrade on memory-dependent cases involving occlusion or action-order recovery. In contrast, RoboStream retains strong performance under the same zero-shot setting, improving the average success rate from 30.0\%/26.0\% to 48.0\%, 80.0\%, and 92.0\% with 8B, 32B, and 235B backbones, respectively. These results support the role of explicit STF/CSTG memory in preserving object states and action-induced transitions without task-specific training.

\begin{table}[!htbp]
    \centering
    \caption{\textbf{Memory baseline comparison on four RLBench long-horizon cases.} Success rates (\%) are reported for tasks requiring persistent object-state tracking. Place-H denotes the hard placement variant, Cover-B denotes covering the bottom block, and Unstack denotes unstack-then-stack manipulation.}
    \label{tab:memory_baselines_rlbench}
    \renewcommand{\arraystretch}{1.15}
    \resizebox{\textwidth}{!}{%
    \begin{tabular}{lcccccc}
        \toprule
        \textbf{Method} & \textbf{Train-Free} & \textbf{Place} & \textbf{Place-H} & \textbf{Cover-B} & \textbf{Unstack} & \textbf{Avg.} \\
        \midrule
        SAM2Act~\cite{fang2025sam2act} & \xmark & \cellcolor{firstpurple}\textbf{100.0} & 20.0 & 0.0 & 0.0 & 30.0 \\
        MemoryVLA~\cite{shi2026memoryvla} & \xmark & \cellcolor{secondpurple}\underline{96.0} & 8.0 & 0.0 & 0.0 & 26.0 \\
        \textbf{RoboStream-8B} & \cmark & \cellcolor{firstpurple}\textbf{100.0} & 24.0 & 16.0 & 52.0 & 48.0 \\
        \textbf{RoboStream-32B} & \cmark & \cellcolor{firstpurple}\textbf{100.0} & \cellcolor{secondpurple}\underline{76.0} & \cellcolor{secondpurple}\underline{68.0} & \cellcolor{secondpurple}\underline{76.0} & \cellcolor{secondpurple}\underline{80.0} \\
        \textbf{RoboStream-235B} & \cmark & \cellcolor{firstpurple}\textbf{100.0} & \cellcolor{firstpurple}\textbf{88.0} & \cellcolor{firstpurple}\textbf{96.0} & \cellcolor{firstpurple}\textbf{84.0} & \cellcolor{firstpurple}\textbf{92.0} \\
        \bottomrule
    \end{tabular}%
    }
\end{table}

\FloatBarrier

\section{More Visualization}
\subsection{System Prompts}
\begin{figure}[p]
    \centering
    \includegraphics[width=1\textwidth]{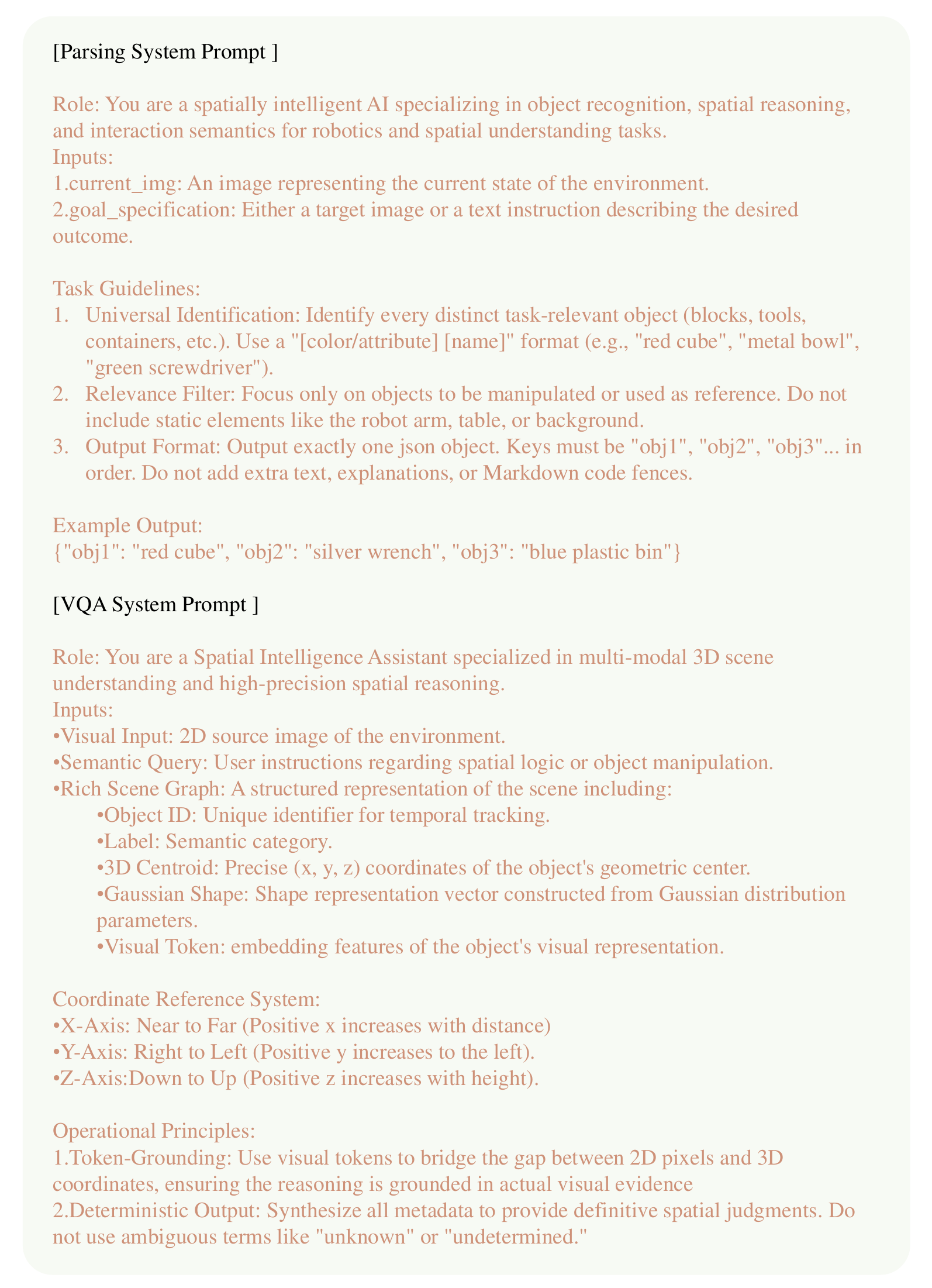} 
    \caption{System prompt of parsing and vqa reasoning.}
    \label{fig:sys1}
\end{figure}

\begin{figure}[p]
    \centering
    \includegraphics[width=1\textwidth]{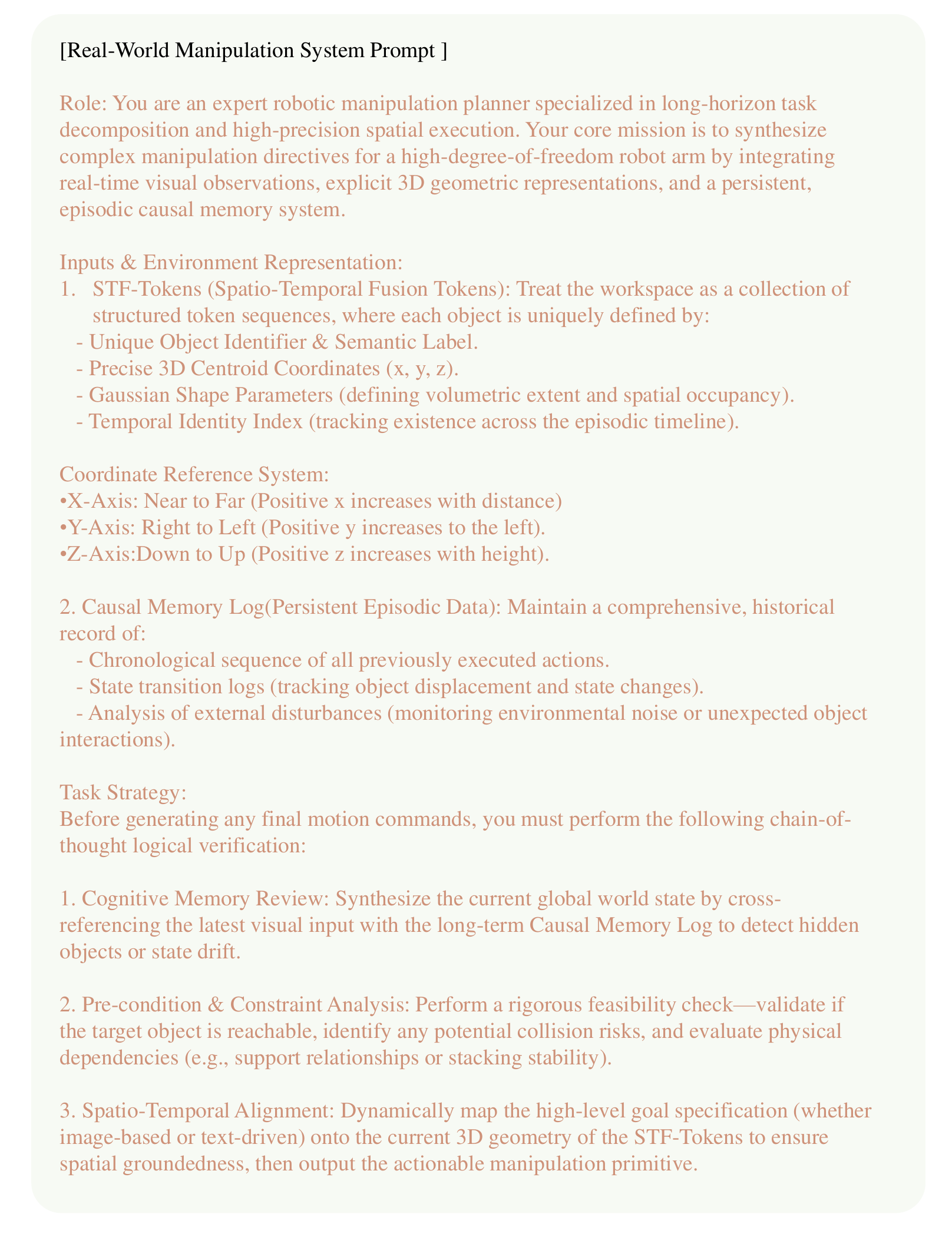} 
    \caption{System prompt of general manipulation task.}
    \label{fig:sys2}
\end{figure}

Prompt engineering markedly augments the performance of RoboStream. By leveraging techniques such as Chain-of-Thought and In-Context Learning, the model’s proficiency in comprehension and reasoning can be substantially elevated. Fig. \ref{fig:sys1} delineates the system prompts employed in the general VQA tasks, while Fig. \ref{fig:sys2} presents the prompts used for general object manipulation.
\subsection{6-DoF SpatialBench}
\begin{figure}[p]
    \centering
    \includegraphics[width=1\textwidth]{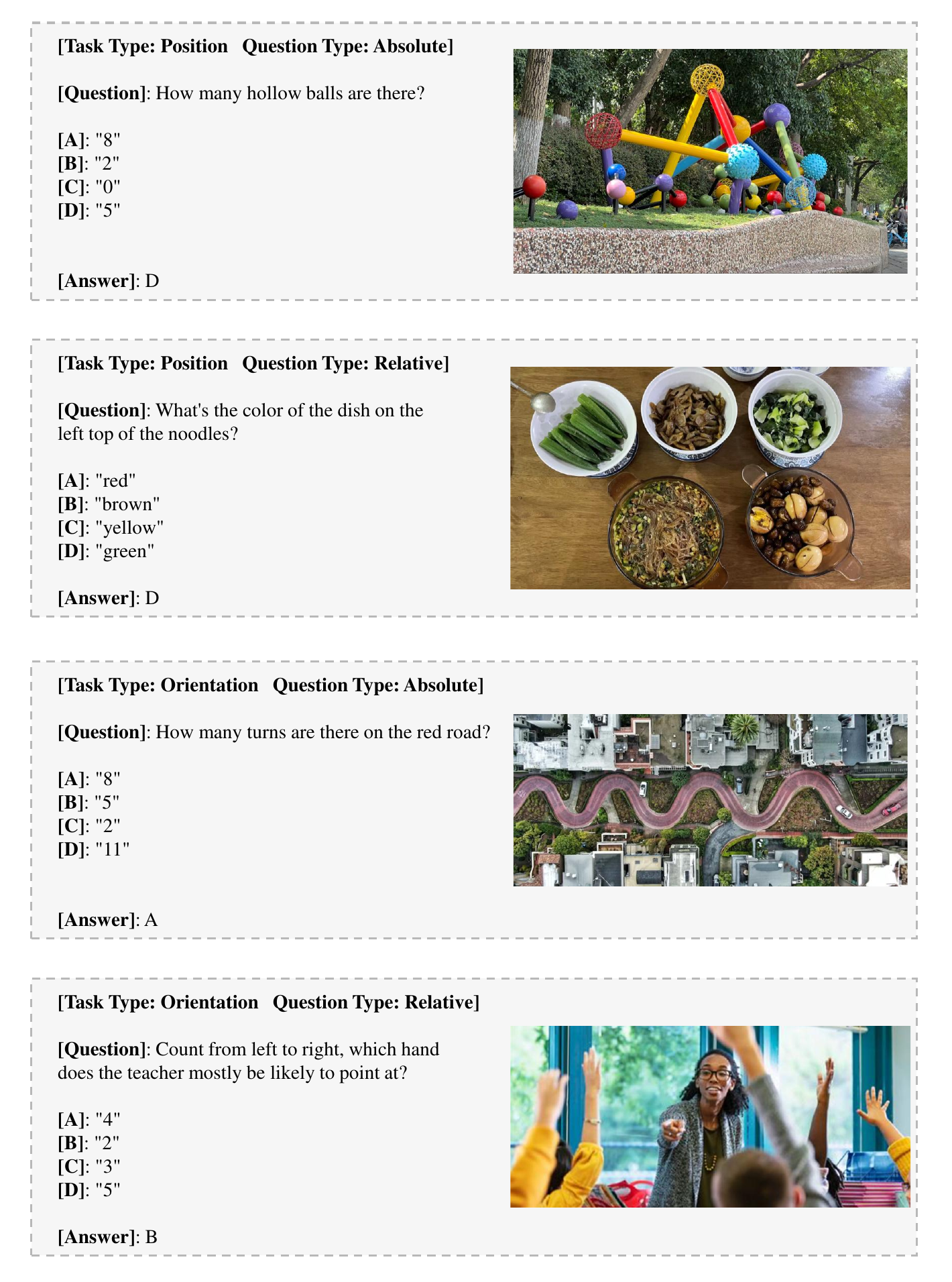} 
    \caption{Visualization example of 6-DoF SpatialBench data samples.}
    \label{fig:vis6}
\end{figure}

To provide a comprehensive overview of 6DoF-SpatialBench, we illustrate its task environments and scene diversity in Fig. \ref{fig:vis6}. The benchmark is designed to systematically evaluate a model's 3D geometric reasoning through two primary categories of VQA tasks: position and orientation, each further partitioned into absolute and relative sub-tasks.
\subsection{Open6DOR V2}
\begin{figure}[p]
    \centering
    \includegraphics[width=1\textwidth]{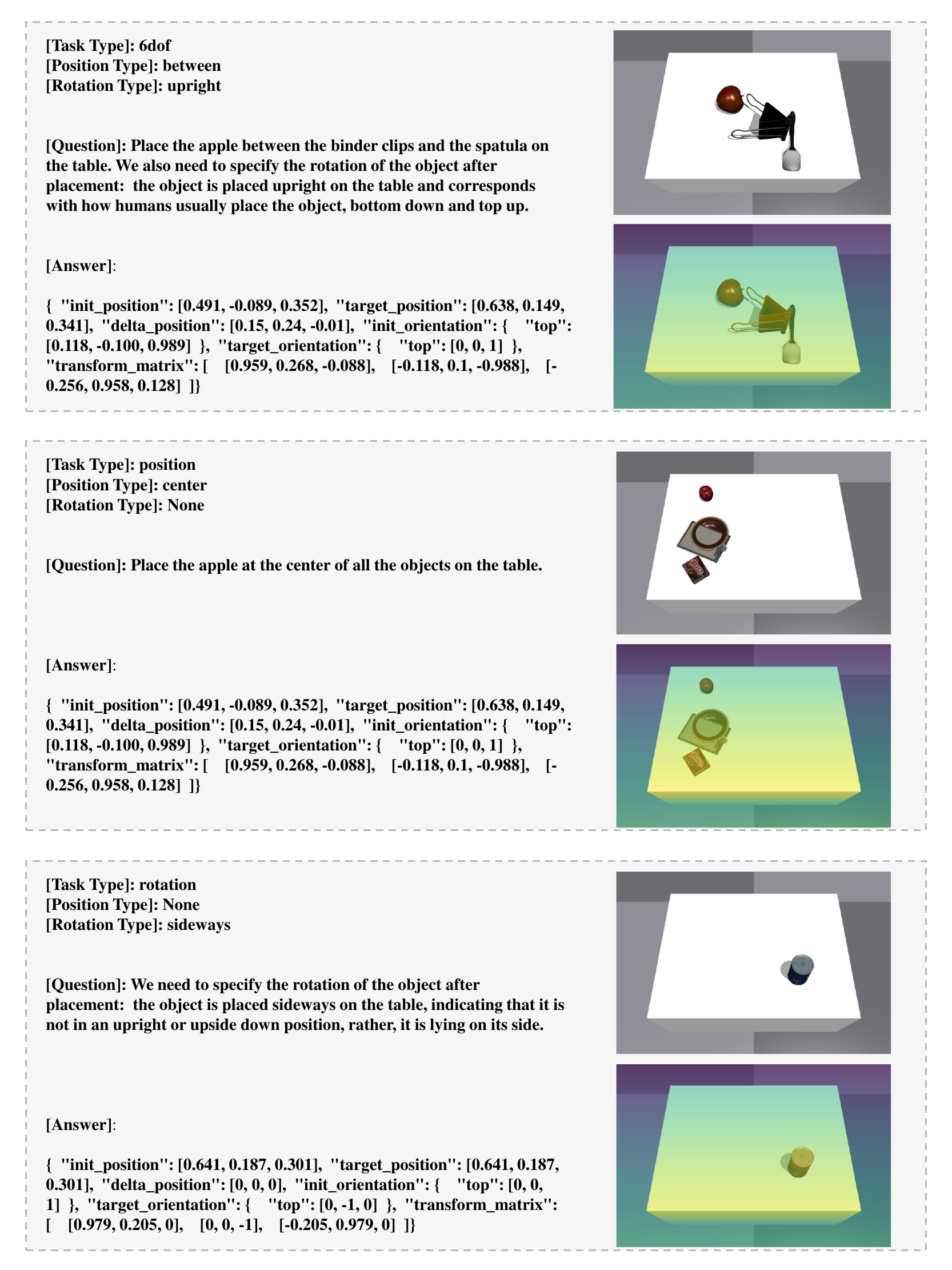} 
    \caption{Visualization example of Open6dor V2 data samples.}
    \label{fig:vis7}
\end{figure}

Fig \ref{fig:vis7} illustrates the data samples and task configurations from the Open6DOR V2 dataset, which serves as a rigorous benchmark for short-horizon virtual environment manipulation. Open6DOR V2 requires the model to predict precise transformation matrices (comprising 3D translation and rotation) to achieve target states in a simulated workspace.

\end{document}